\RequirePackage{fix-cm}

\documentclass[twocolumn]{svjour3}
\newcommand{\colorme}[1]{#1\xspace}
\newcommand{\colormeall}{}

\usepackage{array}
\usepackage{hhline}
\usepackage{multirow}
\usepackage[dvipsnames]{xcolor}
\usepackage{amsmath}
\usepackage{amsfonts}
\usepackage{amssymb}
\usepackage{graphics}
\graphicspath{{Figures_1/},{Figures_2/}}
\usepackage{graphicx}
\usepackage{epsfig}
\usepackage{xspace} 
\usepackage{soul}
\usepackage{breakurl}

\usepackage[T1]{fontenc}

\usepackage{enumitem}
\usepackage[pagebackref=true,breaklinks=false,letterpaper=true,colorlinks,bookmarks=false]{hyperref}

\usepackage{import}
\usepackage{bm}

\newcommand{\subparagraph}{}
\usepackage{titlesec}
\usepackage{etoolbox}

\makeatletter
\patchcmd{\ttlh@hang}{\parindent\z@}{\parindent\z@\leavevmode}{}{}
\patchcmd{\ttlh@hang}{\noindent}{}{}{}
\makeatother

\titleformat{\paragraph}[runin]{\itshape}{\theparagraph}{1em}{}[~~]

\newcommand{\equationspersection}{\numberwithin{equation}{section}}

\def\*#1{\mathbf{#1}}

\newcommand{\btrho}{\tilde{\bm \rho}}
\newcommand{\btz}{\tilde{\bm z}}

\newcommand{\off}[1]{}

\newcommand\inlineeqno{\stepcounter{equation}\ (\theequation)}

\newcommand{\iali}[1]{\begin{align}#1\end{align}}
\newcommand{\ieqn}[1]{\begin{equation}#1\end{equation}}
\newcommand{\imtx}[1]{\begin{bmatrix}#1\end{bmatrix}}
\newcommand{\iprf}[1]{\begin{proof}#1\end{proof}}

\newcommand{\bR}{\mathbb{R}}
\newcommand{\sr}{\mathrm{sr}}
\newcommand{\bv}{\mathbf{v}}

\usepackage[noend]{algorithmic}
\newtheorem{algorithm}{Algorithm}

\title{\colorme{LED-based Photometric Stereo: Modeling, Calibration and Numerical Solution}}
\titlerunning{LED-based Photometric Stereo: Modeling, Calibration and Numerical Solution}

\author{Yvain \textsc{Qu{\'e}au} \and Bastien \textsc{Durix} \and \\
Tao \textsc{Wu} \and Daniel \textsc{Cremers}
\and \\
Fran\c cois \textsc{Lauze}  \and Jean-Denis \textsc{Durou} 
}

\institute{Yvain \textsc{Qu{\'e}au} \at
	Department of Computer Science \\
	Technical University of Munich, Germany \\
		\email{yvain.queau@tum.de} \and
	Bastien \textsc{Durix} \at
	IRIT, UMR CNRS 5505 \\
		Université de Toulouse, France \\
		\email{bastien.durix@enseeiht.fr} \and
	Tao \textsc{Wu} \at
		Department of Computer Science \\
		Technical University of Munich, Germany \\
		\email{tao.wu@tum.de} \and
	Daniel \textsc{Cremers} \at
		Department of Computer Science \\
		Technical University of Munich, Germany \\
		\email{cremers@tum.de} \and
	Fran\c cois \textsc{Lauze} \at
		Department of Computer Science \\
		University of Copenhagen, Denmark \\
		\email{francois@di.ku.dk} \and		
	Jean-Denis \textsc{Durou} \at
		IRIT, UMR CNRS 5505 \\
		Université de Toulouse, France \\
		\email{durou@irit.fr} 
}

\begin{document}
\equationspersection
\maketitle

\begin{abstract}
\colorme{We conduct a thorough study of photometric stereo under nearby point light source illumination, from modeling to numerical solution, through calibration.}  
In the classical formulation of \colorme{photometric stereo}, the luminous fluxes are assumed to be directional, which is very difficult to achieve in practice. 
Rather, we use light-emitting diodes (LEDs) to illuminate the scene to be reconstructed. 
\colorme{Such point light sources are very convenient to use, yet they yield a more complex photometric stereo model which is arduous to solve. We first derive in a physically sound manner this model, and show how to calibrate its parameters. Then, we discuss two state-of-the-art numerical solutions. The first one alternatingly estimates the albedo and the normals, and then integrates the normals into a depth map. It is shown empirically to be independent from the initialization, but convergence of this sequential approach is not established. The second one directly recovers the depth, by formulating photometric stereo as a system of nonlinear partial differential equations (PDEs), which are linearized using image ratios. Although the sequential approach is avoided, initialization matters a lot and convergence is not established either. Therefore, we introduce a provably convergent alternating reweighted least-squares scheme for solving the original system of nonlinear PDEs. 
} 
Finally, we extend this study to the case of RGB images.
\end{abstract}



\section{Introduction}

3D-reconstruction is one of the most important goals of computer vision. Among the many techniques which can be used to accomplish this task, shape-from-shading \cite{Horn1989a} and photometric stereo~\cite{Woodham1980a} are \emph{photometric} techniques, as they use the relationship between the gray or color levels of the image, the shape of the scene, supposedly opaque, its reflectance and the luminous flux that illuminates it.

\begin{figure}[!htpb]
  \centering
	  \def\svgwidth{0.9\linewidth}
		  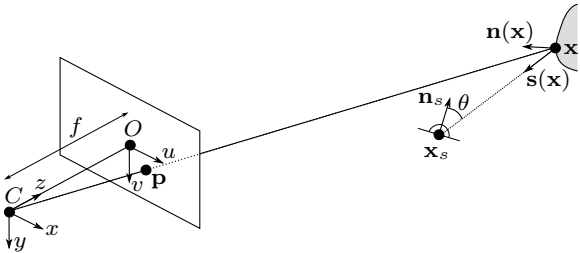
  \caption{Schematic representation of the geometric setup. A point $\mathbf{x} = [x,y,z]^\top \in \mathbb{R}^3$ on the scene surface and a pixel $\mathbf{p} = [u,v]^\top \in \mathbb{R}^2$ in the image plane are conjugated according to Eq.~\eqref{eq:1}. Eq.~\eqref{eq:model} states that, when the scene is illuminated by a LED located in $\mathbf{x}_s \in \mathbb{R}^3$, the gray level $I(\mathbf{p})$ of the pixel $\mathbf{p}$ conjugated to $\mathbf{x}$ is a function of the angle between the lighting vector $\mathbf{s}(\mathbf{x})$ and the normal $\mathbf{n}(\mathbf{x})$ to the surface in $\mathbf{x}$ (illuminance), of the angle $\theta$ between the principal direction $\mathbf{n}_s$ of the LED and $\mathbf{s}(\mathbf{x})$ (anisotropy), of the distance $\|\mathbf{x}-\mathbf{x}_s\|$ between the surface point and the light source \colorme{location} (inverse-of-square falloff), and of the albedo in $\mathbf{x}$ (Lambertian reflectance).}
  \label{fig:1}
  \end{figure}

Let us first introduce some notations that will be used throughout this paper. We describe a point $\mathbf{x}$ on the scene surface by its coordinates $[x, y, z]^\top$ in a frame originating from the optical center $C$ of the camera, such that the plane $Cxy$ is parallel to the image plane and the $Cz$ axis coincides with the optical axis and faces the scene (cf. Fig. \ref{fig:1}). The coordinates $[u, v]^\top$ of a point $\mathbf {p}$ in the image (pixel) are relative to a frame $Ouv$ whose origin is the principal point $O$, and whose axes $Ou$ and $Ov$ are parallel to $Cx$ and $Cy$, respectively. If $f$ refers to the focal length, the conjugation relationship between $\mathbf {x}$ and $\mathbf {p}$ is written, in perspective projection:
\begin{equation}
	\displaystyle\begin{cases}
		x = \dfrac{z}{f} \, u, \\[1em]
		y = \dfrac{z}{f} \, v.
	\end{cases}
\label{eq:1}
\end{equation}

The 3D-reconstruction problem consists in estimating, in each pixel $\mathbf{p}$ of a part $\Omega$ of the image domain, its conjugate point $\mathbf{x}$ in 3D-space. Eq. \eqref{eq:1} shows that it suffices to find the depth $z$ to determine $\mathbf{x} = \left[x,y,z\right]^\top$ from $\mathbf{p} = \left[u,v\right]^\top$. The only unknown of the problem is thus the \emph{depth map} $z$, which is defined as follows:
\begin{equation}
	\begin{array}{rccl}
		z\,: & \Omega \subset \mathbb{R}^2 & \to & \mathbb{R}^+ \\
		~ & \mathbf{p} = [u,v]^\top & \mapsto & z(\mathbf{p}).
	\end{array}
\label{eq:2}
\end{equation}

We are interested in this article in 3D-reconstruction of \emph{Lambertian surfaces} by photometric stereo. The \emph{reflectance} in a point of such a surface is completely characterized by a coefficient $\rho$, called \emph{albedo}, which is 0 if the point is black and~1 if it is white. Photometric stereo is nothing else than an extension of shape-from-shading: instead of a single image, the former uses $m \geqslant$ 3 shots $I^i,\, i \in \{1,\dots,m\}$, taken from the same angle, but under varying lighting. \colorme{Considering multiple images allows to circumvent the difficulties of shape-from-shading: photometric stereo techniques are able to \emph{unambiguously} estimate the 3D-shape as well as the albedo i.e., without resorting to any prior.}

A parallel and uniform illumination can be characterized by a vector $\mathbf{s} \in \mathbb {R}^3$ oriented towards the light source, whose norm is equal to the luminous flux density. We call $\mathbf{s}$ the \emph{lighting vector}. For a Lambertian surface, the classical modeling of photometric stereo is written, in each pixel $\mathbf{p} \in \Omega$, as the following system\footnote {The equalities \eqref{eq:3} are in fact proportionality relationships: see the expression \eqref{eq:17} of $I(\mathbf {p})$.}:
\begin{equation}
	I^i(\mathbf{p}) = \rho(\mathbf{x}) \,\, \mathbf{s}^i \cdot \mathbf{n}(\mathbf{x}),\qquad i\in \{1,\dots,m\},
\label{eq:3}
\end{equation}
where $I^i(\mathbf{p})$ denotes the gray level of $\mathbf{p}$ under a parallel and uniform illumination characterized by the lighting vector $\mathbf{s}^i$, $\rho(\mathbf{x})$ denotes the albedo in the point $\mathbf{x}$ conjugate to $\mathbf{p}$, and ${\mathbf{n}}(\mathbf{x})$ denotes the unit-length outgoing normal to the surface in this point. Since there is a one-to-one correspondence between the points $\mathbf{x}$ and the pixels $\mathbf{p}$, we write for convenience $\rho(\mathbf{p})$ and $\mathbf{n}(\mathbf{p})$, in lieu of $\rho(\mathbf{x})$ and $\mathbf{n}(\mathbf{x})$. Introducing the notation $\mathbf{m}(\mathbf{p}) = \rho(\mathbf {p}) \, \mathbf{n}(\mathbf{p})$, System~\eqref{eq:3} can be rewritten in matrix form:
\begin{equation}
	\mathbf{I}(\mathbf{p}) = \mathbf{S} \, \mathbf{m}(\mathbf{p}),
\label{eq:4}
\end{equation}
where vector $\mathbf{I}(\mathbf{p}) \in \mathbb{R}^m$ and matrix $\mathbf{S} \in \mathbb{R}^{m\times3}$ are defined as follows:
\begin{equation}
	\mathbf{I}(\mathbf{p}) =
		\begin{bmatrix}
			I^1(\mathbf{p}) \\
			\vdots \\
			I^m(\mathbf{p})
		\end{bmatrix}
		\qquad \text{and} \qquad
	\mathbf{S} =
		\begin{bmatrix}
			\mathbf{s}^{1 \top} \\
			\vdots \\
			\mathbf{s}^{m \top}
		\end{bmatrix}.
\label{eq:5}
\end{equation}
As soon as $m \geqslant 3$ \emph{non-coplanar} lighting vectors are used, matrix $\mathbf{S}$ has rank 3. The (unique) least-squares solution of System \eqref{eq:4} is then given by
\begin{equation}
	\mathbf{m}(\mathbf{p}) = \mathbf{S}^\dagger \, \mathbf{I}(\mathbf{p}),
\label{eq:6}
\end{equation}
where $\mathbf{S}^\dagger$ is the \emph{pseudo-inverse} of $\mathbf{S}$. From this solution, we easily deduce the albedo and the normal:
\begin{equation}
	\rho(\mathbf{p}) = \| \mathbf{m}(\mathbf{p}) \|
	\qquad \text{and} \qquad
	\mathbf{n}(\mathbf{p}) = \frac{\mathbf{m}(\mathbf{p})}{\|\mathbf{m}(\mathbf{p})\|}.
\label{eq:7}
\end{equation}
The normal field estimated in such a way must eventually be \emph{integrated} so as to obtain the depth map, knowing that the boundary conditions, the shape of domain~$\Omega$ as well as depth discontinuities significantly complicate this task \cite{Durou2016}.

To ensure lighting directionality, as is required by Model~\eqref{eq:3}, it is necessary to achieve a complex optical setup~\cite{Moreno2006}. It is much easier to use light-emitting diodes (LEDs) as light sources, but with this type of light sources, we should expect significant changes in the modeling, and therefore in the numerical \colorme{solution}. The aim of our work is to conduct a comprehensive and detailed study of photometric stereo under point light source illumination such as LEDs.

\paragraph{Related works.}

\colorme{Modeling the luminous flux emitted by a LED is a well-studied problem, see for instance~\cite{Moreno2008}. One model which is frequently considered in computer vision is that of nearby point light source. This model involves an inverse-of-square law for describing the attenuation of lighting intensity with respect to distance, which has long been identified as a key feature for solving shape-from-shading~\cite{Iwahori1990} and photometric stereo~\cite{Clark1992}. Attenuation with respect to the deviation from the principal direction of the source (anisotropy) has also been considered~\cite{Bennahmias2007}. }

\colorme{If the surface to reconstruct lies in the vicinity of a plane, it is possible to capture a map of this attenuation using a white planar reference object. Conventional photometric stereo~\cite{Woodham1980a} can then be applied to the images compensated by the attenuation maps~\cite{Angelopoulou2014a,McGunnigle2003,Sun2013}. Otherwise, it is necessary to include the attenuation coefficients in the photometric stereo model, which yields a nonlinear inverse problem to be solved.}

\colorme{This is easier to achieve if the parameters of the illumination model have been calibrated beforehand. Lots of methods exist for estimating a source location~\cite{Ackermann2013,Aoto2012,Ciortan2016,Giachetti2015,Hara2005,Powell2001,Shen2011,Takai2009}. Such methods triangulate this location during a calibration procedure, by resorting to specular spheres. This can also be achieved online, by introducing spheres in the scene to reconstruct~\cite{Liao2016}. Calibrating anisotropy is a more challenging problem, which was tackled recently in~\cite{Nie2016,Xie2015b} by using images of a planar surface. Some photometric stereo methods also circumvent calibration by (partly or completely) automatically inferring lighting during the 3D-reconstruction process~\cite{Koppal2007,Liao2016,Logothetis2017,Migita2008,Papadhimitri2014b,SSVM2017}.} 

\colorme{Still, even in the calibrated case, designing numerical schemes for solving photometric stereo under nearby point light sources remains difficult. When only two images are considered, the photometric stereo model can be simplified using image ratios. This yields a quasilinear PDE~\cite{Mecca2015,Mecca2014b} which can be solved by provably convergent front propagation techniques, provided that a boundary condition is known. To improve robustness, this  strategy has been adapted to the multi-images case in~\cite{Logothetis2017,Logothetis_2016,Mecca2016,CVPR2016}, using variational methods. However, convergence guarantees are lost. Instead of considering such a differential approach, another class of methods~\cite{Ahmad2014,Bony2013,Collins2012,Huang2015,Kolagani1992,Nie2016b,Papadhimitri2014b,Yeh2016} rather modify the classical photometric stereo framework~\cite{Woodham1980a}, by alternatingly estimating the normals and the albedo, integrating the normals into a depth map, and updating the lighting based on the current depth. Yet,  no convergence guarantee does exist. A method based on mesh deformation has also been proposed in~\cite{Xie2015}, but convergence is not established either.}

\paragraph{Contributions.} 

\colorme{In contrast to existing works which focus either on modeling, calibrating or solving photometric stereo with near point light sources such as LEDs, the objective of this article is to propose a comprehensive study of all these aspects of the problem. Building upon our previous conference papers~\cite{CVPR2016,SSVM2017,CVPR2017}, we introduce the following innovations:}
\begin{itemize}
	\item We present in Section~\ref{sec:2} an accurate model for photometric stereo under point light source illumination. As in recent works~\cite{Logothetis2017,Logothetis_2016,Mecca2015,Mecca2014b,Mecca2016,Nie2016b,Nie2016,Xie2015b}, this model takes into account the nonlinearities due to distance and to the anisotropy of the LEDs. Yet, it also clarifies the notions of albedo and of source intensity, which are shown to be relative to a reference albedo and to several parameters of the camera, respectively. This section also introduces a practical calibration procedure for the \colorme{location}, the orientation and the relative intensity of the LEDs. \\
	\item Section \ref{sec:3} \colorme{reviews and improves two state-of-the-art} numerical \colorme{solutions} in several manners. We first \colorme{modify} the \colorme{alternating} method~\cite{Ahmad2014,Bony2013,Collins2012,Huang2015,Kolagani1992,Nie2016b,Papadhimitri2014b,Yeh2016}  \colorme{by introducing an estimation of the shape scale, in order to recover the} \textit{absolute} depth without \colorme{any prior}. We then \colorme{study the PDE-based approach which employs image ratios for eliminating the nonlinearities~\cite{Logothetis2017,Logothetis_2016,Mecca2016,CVPR2016}, and empirically show that local minima can be avoided by employing an augmented Lagrangian strategy. Nevertheless,  neither of these state-of-the-art methods is provably convergent.} \\
	\item \colorme{Therefore, we introduce in Section~\ref{sec:4} a new, provably convergent method, inspired by the one recently proposed in~\cite{SSVM2017}. It is based on a tailored alternating reweighted least-squares scheme for approximately solving the non-linearized system of PDEs. Following~\cite{CVPR2017}, we further show that this method is easily extended in order to address shadows and specularities.} \\
	\item In Section \ref{sec:5}, we \colorme{build upon the analysis conducted in~\cite{CVPR2016} in order to tackle the case of RGB-valued images}, before concluding and suggesting several future research directions in Section \ref{sec:6}.
	\end{itemize}


\section{Photometric Stereo under Point Light Source Illumination}
\label{sec:2}

\colorme{Conventional}
photometric stereo~\cite{Woodham1980a} assumes that the primary luminous \colorme{fluxes are} parallel and uniform, which is difficult to guarantee. It is much easier to illuminate a scene with LEDs. 

\begin{figure}[!htpb]
\begin{center}
	\begin{tabular}{c}
		\includegraphics[width = 0.95\linewidth]{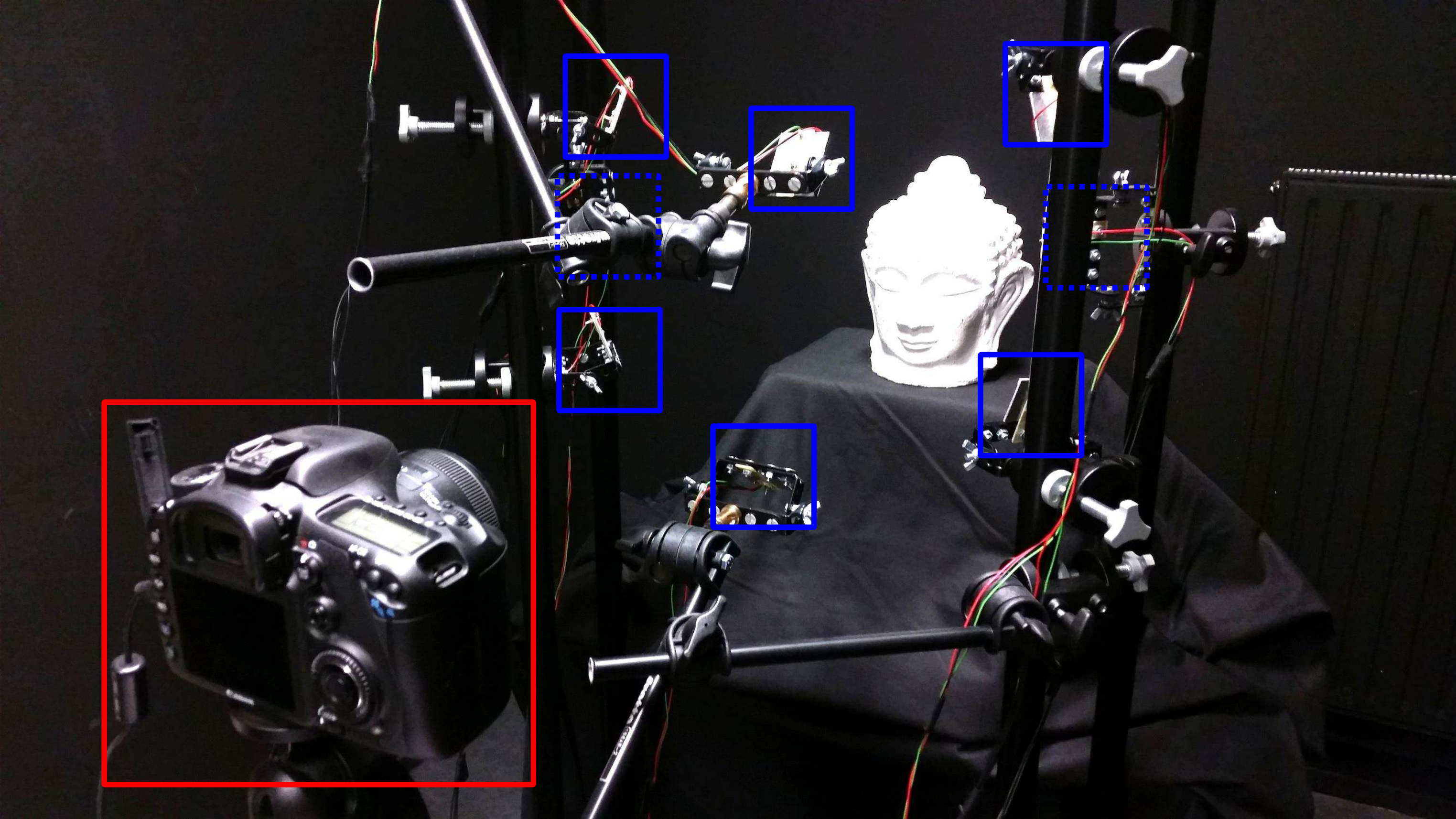} \\
		\small{(a)} \\
		~
	\end{tabular}
	\begin{tabular}{ccc}
		\includegraphics[width = 0.25\linewidth]{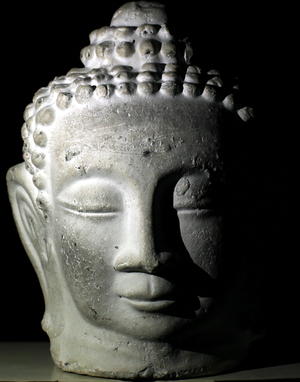} & \quad
		\includegraphics[width = 0.25\linewidth]{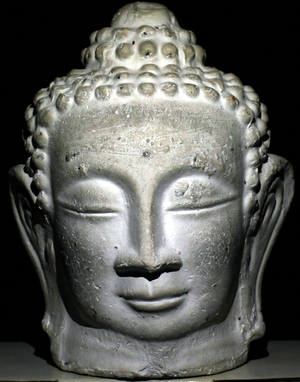} & \quad
		\includegraphics[width = 0.25\linewidth]{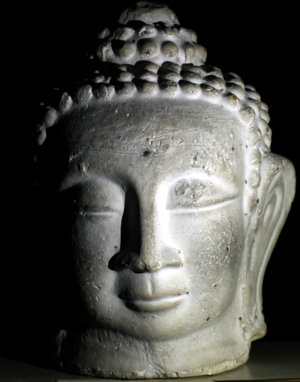} \\
		\small{(b)} & \quad \small{(c)} & \quad \small{(d)}
	\end{tabular}
\end{center}
\caption{(a) Our photometric stereo-based experimental \colorme{setup} for 3D-reconstruction of faces using {a Canon EOS 7D camera ({highlighted} in red) and $m = 8$ LEDs ({highlighted} in blue)}. {The walls are} painted in black in order to avoid {the} reflections between the scene and the environment. (b-c-d) Three out of the $m = 8$ images obtained by this \colorme{setup}.}
\label{fig:2}
\end{figure}

Keeping this in mind, we have developed a photometric stereo-based \colorme{setup} for 3D-reconstruction of faces, which includes $m = 8$ LEDs\footnote{We use white LUXEON Rebel LEDs: \url{http://www.luxeonstar.com/luxeon-rebel-leds}.} located at about $30~cm$ from the scene surface (see Fig.\ \ref{fig:2}-a). The face is photographed by a Canon EOS 7D camera with focal length $f = 35~mm$. Triggering the shutter in burst mode, while {synchronically} lighting the LEDs, provides us with $m = 8$ images such as those of Figs.\ \ref{fig:2}-b, \ref{fig:2}-c and \ref{fig:2}-d. In this section, we aim at modeling the formation of such images, by establishing the following result:~\\

\begin{result}
If the $m$ LEDs are modeled as anisotropic (imperfect Lambertian) point light sources, if the surface is Lambertian and if all the automatic settings of the camera are deactivated, then the formation of the $m$ images can be modeled as follows, for $i \in \{1,\dots,m\}$:
\begin{equation}
  I^i(\mathbf{p}) = \Psi^i \, \overline{\rho}(\mathbf{p}) 
  \left[  \frac{  \mathbf{n}^i_s \! \cdot \! \left( \mathbf{x}\!-\!\mathbf{x}^i_s \right) } {\|\mathbf{x}\!-\!\mathbf{x}^i_s\|} \right]^{\mu^i}
		\!\!\frac{\left\{(\mathbf{x}^i_s\!-\!\mathbf{x}) \cdot \mathbf{n}(\mathbf{p}) \right\}_+}{\|\mathbf{x}^i_s\!-\!\mathbf{x}\|^3},
		\label{eq:model}
\end{equation}
where:
\begin{itemize}
  \item $I^i(\mathbf{p})$ is the ``corrected gray level'' at pixel $\mathbf{p}$ conjugate to a point $\mathbf{x}$ located on the surface (cf. Eq.~\eqref{eq:17});
  \item $\Psi^i$ is the intensity of the $i$-th source multiplied by an unknown factor, which is common to all the sources and depends on several camera parameters and on the albedo $\rho_0$ of a Lambertian planar calibration pattern (cf. Eq.~\eqref{eq:19});
  \item $ \overline{\rho}(\mathbf{p})$ is the albedo of the surface point $\mathbf{x}$ conjugate to pixel $\mathbf{p}$, relatively to $\rho_0$ (cf. Eq.~\eqref{eq:28}); 
  \item $\mathbf{n}^i_s \in \mathbb{S}^2 \subset \mathbb{R}^3$ is the (unit-length) principal direction of the $i$-th source, $\mathbf{x}^i_s \in \mathbb{R}^3$ its location  (cf. Fig.~\ref{fig:2}), and $\mu^i \geq 0$ its anisotropy parameter (cf. Fig.~\ref{fig:3} and Eq.~\eqref{eq:10}); 
  \item $\{\cdot\}_+$ is the positive part operator, which accounts for self-shadows:
  \begin{equation}
    \{x\}_+ = \max\{x,0\}. 
  \end{equation} 
\end{itemize}
\end{result}

In Eq.~\eqref{eq:model}, the anisotropy parameters $\mu^i$ are (indirectly) provided by the manufacturer (cf. Eq.~\eqref{eq:11}), and the other LEDs parameters $\Psi^i$, $\mathbf{n}^i_s$ and $\mathbf{x}^i_s$ can be calibrated thanks to the procedure described in Section~\ref{sec:2.2}. The only unknowns in System~\eqref{eq:model} are thus the depth $z$ of the 3D-point $\mathbf{x}$ conjugate to $\mathbf{p}$, its (relative) albedo $\overline{\rho}(\mathbf{p})$ and its normal $\mathbf{n}(\mathbf{p})$. The estimation of these unknowns will be discussed in Sections~\ref{sec:3} and~\ref{sec:4}. Before that, let us show step-by-step how to derive Eq.~\eqref{eq:model}. 

\subsection{Modeling the Luminous Flux Emitted by a LED}
\label{sec:2.1}

For the LEDs we use, the characteristic illuminating volume is of the order of \colorme{one cubic} millimeter. Therefore, in comparison with the scale of a face, each LED can be seen as a point light source located at $\mathbf{x}_s \in \mathbb{R}^3$. At any point $\mathbf{x} \in \mathbb{R}^3$, the {lighting vector $\mathbf{s}(\mathbf{x})$} is necessarily radial \colorme{i.e.,} collinear with the unit\colorme{-length} vector $\textbf{u}_r = \frac{\mathbf{x} - \mathbf{x}_s} {\| \mathbf{x} - \mathbf{x}_s \|}$. Using spherical coordinates $(r, \theta, \phi)$ of $\mathbf {x}$ in a frame having $\mathbf{x}_s$ as origin, {it} is written
\begin{equation}
	\mathbf{s}(\mathbf{x}) = -\frac{\Phi(\theta,\phi)} {r^2} \, \textbf{u}_r,
\label{eq:8}
\end{equation}
where $\Phi(\theta,\phi)\geqslant0$ denotes the {\emph{intensity} of the source\footnote{\colorme{The intensity is expressed in lumen per steradian ($lm \cdot sr^{-1}$) i.e., in candela ($cd$).}}}, and the $1/r^2$ attenuation is a consequence of the conservation of luminous energy in a non-absorbing medium. Vector $\mathbf{s}(\mathbf{x})$ is purposely oriented in the opposite direction from that of the light, in order to simplify the writing of the Lambertian model.

{Model \eqref{eq:8} is very general. We could project the intensity $\Phi(\theta,\phi)$ on the \emph{spherical harmonics} basis, which allowed Basri et al. to model the luminous flux in the case of uncalibrated photometric stereo \cite{Basri2003}. We could also sample $\Phi(\theta,\phi)$} in the vicinity of a plane, using a plane with known reflectance~\cite {Angelopoulou2014a,McGunnigle2003,Sun2013}.

{Using} the specific characteristics of LEDs may lead to a more accurate model. Indeed, most of the LEDs emit a luminuous flux which is invariant by rotation around a \emph{{principal} direction} indicated by a unit\colorme{-length} vector~$\mathbf{n}_s$~\cite{Moreno2008}. If $\theta$ is defined relatively to $\mathbf{n}_s$, this means that $\Phi(\theta,\phi)$ is independent from $\phi$. The {lighting vector in $\mathbf{x}$ induced by} a LED located in $\mathbf{x}_s$ is thus written
\begin{equation}
	\mathbf{s}(\mathbf{x}) = \frac{\Phi(\theta)} {\|\mathbf{x}_s-\mathbf{x}\|^2} \,
		\frac{\mathbf{x}_s-\mathbf{x}}{\|\mathbf{x}_s-\mathbf{x}\|}.
\label{eq:9}
\end{equation}

\begin{figure}[!htbp]
\begin{center}
	\begin{tabular}{cc}
		\includegraphics[width = 0.4\linewidth]{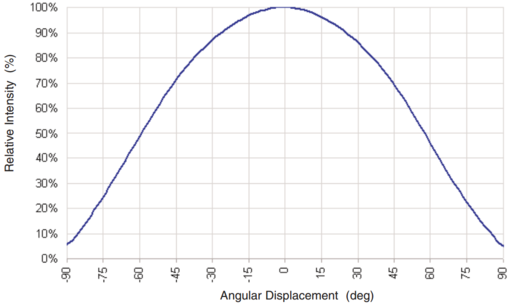} &
		\includegraphics[width = 0.5\linewidth]{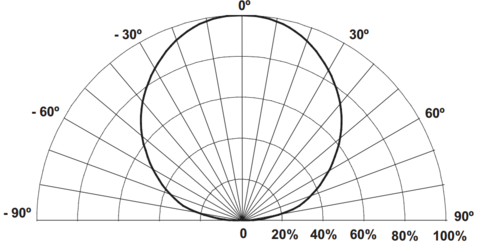}  \\
		\small{(a)} & \small{(b)}
	\end{tabular}
\end{center}
\caption{{{Intensity} patterns of the LEDs used (source: \url{http://www.lumileds.com/uploads/28/DS64-pdf}). \colorme{(a)} Anisotropy function $\Phi(\theta) / \Phi_0$ as a function of $\theta$. \colorme{(b)} Polar representation. {These diagrams show us that} $\theta_{1/2} = \pi /3$, which corresponds to $\mu = 1$ according to Eq.~\eqref{eq:11} (Lambertian source).}}
\label{fig:3}
\end{figure}

The dependency on $\theta$ of the intensity $\Phi$ characterizes the \textit{anisotropy} of the LED. The function $\Phi (\theta)$ is generally decreasing {over} $[0, \pi / 2]$ {(cf. Fig.\ \ref{fig:3})}.

An anisotropy model satisfying this constraint is that of ``imperfect Lambertian {source}'':
\begin{equation}
	\Phi(\theta) = \Phi_0 \, \cos^\mu \theta,
\label{eq:10}
\end{equation}
which contains two parameters $\Phi_0=\Phi(0)$ and $\mu\geqslant0$, and models both isotropic sources ($\mu=0$) and Lambertian sources ($\mu=1$). Model \eqref{eq:10} is empirical{, and more elaborate models are sometimes considered~\cite{Moreno2008}}, yet it has already been used in photometric stereo \cite{Logothetis2017,Logothetis_2016,Mecca2016,Mecca2015,Nie2016b,Nie2016,SSVM2017,Xie2015b}, including \colorme{the case where} all the LEDs are arranged on a plane parallel to the image plane, in such a way that $\mathbf{n}_s = [0,0,1]^\top$ \cite{Mecca2014b}. Model~\eqref{eq:10} has proven itself and, moreover, {LEDs} manufacturers provide the angle~$\theta_{1/2}$ such that $\Phi(\theta_{1/2}) = \Phi_0/2$, from which we deduce{, using \eqref{eq:10}, the value of $\mu$}:
\begin{equation}
	\mu = -\frac{\log(2)}{\log(\cos \theta_{1/2})}.
\label{eq:11}
\end{equation}
{As shown in Fig.\ \ref{fig:3}, the angle $\theta_{1/2}$ is $\pi/3$ for the LEDs we use.} From Eq. \eqref{eq:11}, we deduce that $\mu=1$, {which means that these LEDs} are Lambertian. Plugging the expression \eqref{eq:10} of $\Phi(\theta)$ into \eqref{eq:9}, {we obtain}
\begin{equation}
	\mathbf{s}(\mathbf{x}) = \Phi_0 \, \cos^\mu \theta \,
		\frac{\mathbf{x}_s-\mathbf{x}}{\|\mathbf{x}_s-\mathbf{x}\|^3},
\label{eq:12}
\end{equation}
where we explicitly keep $\mu$ to address the most general case. Model~\eqref{eq:12} thus includes seven parameters: three for the coordinates of $\mathbf{x}_s$, two for the unit vector $\mathbf{n}_s$, plus $\Phi_0$ and $\mu$. Note that $\mathbf{n}_s$ appears in this model through the angle $\theta$.

In its uncalibrated version, photometric stereo allows {the} 3D-reconstruction of a scene surface without knowing the lighting. \colorme{Uncalibrated photometric stereo} has been widely studied, {including the case} of nearby point light sources \cite{Huang2015,Koppal2007,Migita2008,Papadhimitri2014b,Yeh2016}, but if {this is possible}, we should rather calibrate the lighting\footnote{It is also necessary to calibrate the camera, since the 3D-frame is attached to it. We assume that this has been made {beforehand}.}.


\subsection{Calibrating the Luminous Flux Emitted by a LED}
\label{sec:2.2}

Most calibration methods of {a point light source} \cite{Ackermann2013,Aoto2012,Ciortan2016,Giachetti2015,Hara2005,Powell2001,Shen2011,Takai2009} do not take into account the attenuation of the {luminous flux density} as a function of the distance to the source, nor the possible anisotropy of the source, which may lead to relatively imprecise results. To our knowledge, there are few calibration procedures taking into account these phenomena. In \cite{Xie2015b}, Xie et al. use a single {pattern}, {which is} partially specular and partially Lambertian, to calibrate a LED. We intend to improve this procedure using two patterns, one specular and {the other} Lambertian. The specular one will be used to determine the {location} of the LEDs by triangulation, and the Lambertian one to determine {some} other parameters by minimizing the reprojection error, as recently proposed by Pintus et al. in \cite{Pintus2016}.

\paragraph{Specular Spherical Calibration Pattern.}
{The location} $\mathbf{x}_s$ of a LED can be determined by triangulation. In \cite{Powell2001}, Powell et al. advocate the use of a spherical mirror. To estimate the {locations} of the $m = 8$ LEDs for our \colorme{setup}, we use a billiard ball. Under perspective projection, the edge of the silhouette of a sphere is an ellipse, which we detect using {a dedicated algorithm~\cite{Patraucean2012}}. It is then easy to determine the {3D-coordinates of any point on the surface, {as well as} its normal}, since the radius of the billiard ball is known. For each {pose} of the billiard ball, detecting the reflection of the LED allows us to determine, by reflecting the line of sight on the spherical mirror, a line in 3D-space passing through $\mathbf{x}_s$. In theory, two poses of the billiard ball are enough to estimate $\mathbf{x}_s$, {even if two lines in 3D-space do not necessarily intersect}, but the use of ten poses improves the robustness of the estimation.

\vspace*{-0.5em}

\paragraph{Lambertian Model.}
To estimate the {principal} direction $\mathbf{n}_s$ and the intensity $\Phi_0$ in Model \eqref{eq:12}, we use a Lambertian calibration pattern. A surface is Lambertian if the apparent clarity of any point $\mathbf{x}$ {located on it} is independent from the {viewing angle}. The {\emph{luminance}} $L(\mathbf{x})$, which is equal to the luminous flux emitted per unit of solid angle and per unit of apparent surface, is independent from the direction of emission. However, the {luminance} is not characteristic of the surface, as it depends on the {\emph{illuminance}} $E(\mathbf{x})$ {(denoted $E$ from French ``éclairement'')}, that is to say on the luminous flux {per unit area received by the surface in $\mathbf{x}$}. The relationship between {luminance and illuminance}\footnote{{A luminance is expressed in $lm\cdot m^{-2} \cdot sr^{-1}$ (or $cd \cdot m^{-2}$), an illuminance in $lm \cdot m^{-2}$, \colorme{or lux}~($lx$).}} is written, for a Lambertian surface:
\begin{equation}
	L(\mathbf{x}) = \frac{\rho(\mathbf{x})}{\pi}\,E(\mathbf{x}),
\label{eq:13}
\end{equation}
where the albedo {$\rho(\mathbf{x})\in [0,1]$ is defined as the proportion of luminous energy \colorme{which is reemitted} i.e., $\rho(\mathbf{x}) = 1$ if $\mathbf{x}$ is white, and $\rho(\mathbf{x}) = 0$ if it is black.}

The parameter $\rho(\mathbf{x})$ {is enough to characterize} the reflectance\footnote{The reflectance is generally referred to as the bidirectional reflectance distribution function, or BRDF.} of a Lambertian surface. In addition, the {illuminance} at a point $\mathbf{x}$ of {a (\colorme{not} necessarily Lambertian) surface with normal $\mathbf{n}(\mathbf{x})$, lit by the lighting vector} $\mathbf{s}(\mathbf{x})$, is written\footnote{Negative values in the right hand side of Eq.~\eqref{eq:14} are clamped to zero in order to account for self-shadows.}
\begin{equation}
	E(\mathbf{x}) = \left\{\mathbf{s}(\mathbf{x}) \cdot \mathbf{n}(\mathbf{x})\right\}_+.
\label{eq:14}
\end{equation}

{Focusing the camera on a point $\mathbf{x}$ of the scene surface}, the {illuminance} $\epsilon(\mathbf{p})$ of the image plane, at pixel $\mathbf{p}$ conjugate to $\mathbf{x}$, is related to the {luminance} $L(\mathbf{x})$ by the following ``almost linear'' relationship \cite{Horn1986}:
\begin{equation}
	\epsilon(\mathbf{p}) = \beta \, \cos^4\alpha(\mathbf{p}) \, L(\mathbf{x}),
\label{eq:15}
\end{equation}
where $\beta$ is a proportionality coefficient characterizing the clarity of the image, which depends on several factors such as the lens aperture, the magnification, etc. Regarding {the} factor $\cos^4 \alpha(\mathbf {p})$, where $\alpha(\mathbf{p})$ is the angle between the line of sight and the optical axis, it is responsible for darkening at the periphery of the image\colorme{. This effect should not be confused with vignetting, since it occurs even with ideal lenses~\cite{Gardner1947}.}

With current photosensitive receptors, the \emph{gray level} $J(\mathbf{p})$ at pixel $\mathbf{p}$ is almost proportional\footnote{{Provided that the RAW image format is used.}} to its {illuminance} $\epsilon(\mathbf{p})$, except of course in case of saturation. Denoting~$\gamma$ this coefficient of quasi-proportionality, and combining equalities \eqref{eq:13}, \eqref{eq:14} and \eqref{eq:15}, we get the following expression of the gray level in a pixel $\mathbf{p}$ conjugate to a point $\mathbf{x}$ located on a Lambertian surface:
\begin{equation}
	J(\mathbf{p}) = \gamma \, \beta \, \cos^4\alpha(\mathbf{p}) \, \frac{\rho(\mathbf{x})}{\pi} \, \left\{\mathbf{s}(\mathbf{x}) \cdot \mathbf{n}(\mathbf{x})\right\}_+.
\label{eq:16}
\end{equation}
We have already mentioned that there {is} a one-to-one correspondence between a point $\mathbf{x}$ and its conjugate pixel $\mathbf{p}$, which allows us to denote $\rho(\mathbf{p})$ and $\mathbf{n}(\mathbf{p})$ instead of $\rho(\mathbf{x})$ and $\mathbf{n}(\mathbf{x})$. As the factor $\cos^4 \alpha(\mathbf{p})$ is easy to calculate in each pixel $\mathbf{p}$ of the photosensitive receptor, {since $\cos \alpha(\mathbf{p}) = \frac{f}{\sqrt{\|\mathbf{p}\|^2+f^2}}$}, we can very easily compensate for this source of darkening and will manipulate from now on the ``corrected gray level'':
\begin{equation}
	I(\mathbf{p}) = \frac{J(\mathbf{p})}{\cos^4\alpha(\mathbf{p})} = \gamma \, \beta \, \frac{\rho(\mathbf{p})}{\pi} \, \left\{ \mathbf{s}(\mathbf{x}) \cdot \mathbf{n}(\mathbf{p}) \right\}_+.
\label{eq:17}
\end{equation}

\begin{figure}[!htpb]
\centering
\begin{tabular}{cc}
	\includegraphics[width = 0.45\linewidth]{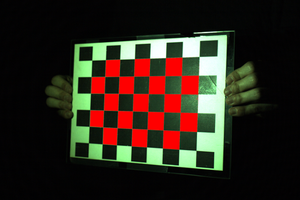} \quad &
	\includegraphics[width = 0.45\linewidth]{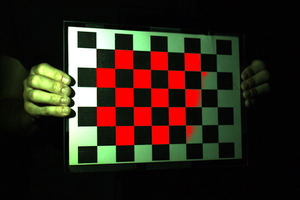} \\
	(a)  \quad & (b)  
\end{tabular}
\caption{Two out of the $q$ poses of the  Lambertian planar calibration pattern used for {the} photometric calibration of the LEDs. The parts of the white cells which are used for estimating the LEDs {principal directions} and intensities are \colorme{highlighted} in red.}
\label{fig:4}
\end{figure}

\paragraph{Lambertian Planar Calibration Pattern.}
To estimate the parameters $\mathbf{n}_s$ and $\Phi_0$ in Model \eqref{eq:12} i.e., to achieve \emph{photometric calibration}, we use a second calibration pattern consisting of a checkerboard printed on a white paper sheet, which is itself stuck on a plane (cf. Fig.~\ref{fig:4}), with the hope that the unavoidable outliers to the Lambertian model will not influence the accuracy of the estimates too much.

The use of a convex calibration pattern (planar, in this case) has a significant advantage: the {lighting vector $\mathbf{s}(\mathbf{x})$ at any point $\mathbf{x}$ of the surface is purely \emph{primary} \colorme{i.e.}, it is only due to} the light source, without ``bouncing'' on other parts of the surface of the target, {provided that} the walls and surrounding objects are covered in black (see Fig.\ \ref{fig:2}-a). {Thanks to} this observation, we can replace the {lighting vector $\mathbf{s}(\mathbf{x})$ in Eq.\ \eqref{eq:17} by the expression \eqref{eq:12} which models the luminous flux emitted by a LED}. From \eqref{eq:12} and \eqref{eq:17}, we deduce the gray level $I(\mathbf{p})$ of the image of a point $\mathbf{x}$ located on this calibration pattern, illuminated by a LED:
\begin{equation}
	I(\mathbf{p}) = \gamma \, \beta \, \frac{\rho(\mathbf{p})}{\pi} \, \Phi_0 \cos^\mu \! \theta 
		\frac{ \left\{ (\mathbf{x}_s-\mathbf{x}) \!\cdot\! \mathbf{n}(\mathbf{p}) \right\}_+}{\|\mathbf{x}_s-\mathbf{x}\|^3}.
\label{eq:18}
\end{equation}

If $q \geqslant 3$ poses of the checkerboard are used, numerous algorithms exist for unambiguously estimating the coordinates of the points $\mathbf{x}^j$ of the pattern, for the different poses $j \in \{1,\dots,q\}$. These algorithms also allow the estimation of the $q$ normals $\mathbf{n}^j$ (we omit the dependency {in} $\mathbf{p}$ of {$\mathbf{n}^j$}, since the pattern is {planar}), and the intrinsic parameters of the camera\footnote{To perform these operations, we {use} the \emph{Computer Vision} toolbox from Matlab.}. As for the albedo, if the use of white paper does not guarantee that $\rho(\mathbf{p}) \equiv 1$, it still seems reasonable to assume $\rho(\mathbf{p}) \equiv \rho_0$ \colorme{i.e.}, {to assume a uniform albedo} in the white cells. We can then group all the {multiplicative} coefficients of the right hand side of Eq.~\eqref{eq:18} into one coefficient
\begin{equation}
	\Psi = \gamma \, \beta \, \frac{\rho_0}{\pi} \, \Phi_0.
\label{eq:19}
\end{equation}
With this definition, and knowing that $\theta$ is the angle between vectors $\mathbf{n}_s$ and $\mathbf{x}-\mathbf{x}_s$, Eq.\ \eqref{eq:18} can be rewritten, in a pixel $\mathbf{p}$ of the set $\Omega^j$ containing the white pixels of the checkerboard in the $j\textsuperscript{th}$ pose (these pixels are \colorme{highlighted} in red in the images of Fig.\ \ref{fig:4}):
\begin{equation}
	I^j(\mathbf{p})
	= \Psi \left[  \displaystyle \frac{\mathbf{n}_s \cdot \left(\mathbf{x}^j-\mathbf{x}_s\right) } {\|\mathbf{x}^j-\mathbf{x}_s\|} \right]^\mu
		\displaystyle \frac{\left\{(\mathbf{x}_s-\mathbf{x}^j) \cdot \mathbf{n}^j \right\}_+}{\|\mathbf{x}_s-\mathbf{x}^j\|^3}.
\label{eq:20}
\end{equation}
To be sure that in Eq.~\eqref{eq:20}, $\Psi$ is independent from the pose $j$, we must deactivate all automatic settings of the camera, in order to make $\beta$ and $\gamma$ constant.

Since $\mathbf{x}_s$ {is already estimated, and the value of $\mu$ is known, the only unknowns {in} Eq. \eqref{eq:20} are $\mathbf{n}_s$ and $\Psi$. Two cases may occur:
\begin{itemize}
	\item If the LED to calibrate is isotropic \colorme{i.e.,} if $\mu=0$, then it is useless to estimate $\mathbf{n}_s$, and $\Psi$ can be estimated in a least-squares sense, by solving
	\begin{equation}
		\underset{\Psi}{\operatorname{\min}~} \sum_{j=1}^{q}
			\sum_{\mathbf{p} \in \Omega^j} \left[ I^j(\mathbf{p}) -
			\Psi \, \frac{\left\{(\mathbf{x}_s-\mathbf{x}^j) \cdot \mathbf{n}^j \right\}_+}{\|\mathbf{x}_s-\mathbf{x}^j\|^3} \right]^2,
	\label{eq:21}
	\end{equation}
	whose solution is given by
	\begin{equation}
		\Psi = \displaystyle\frac{\displaystyle\sum_{j=1}^{q} \sum_{\mathbf{p} \in \Omega^j} I^j(\mathbf{p}) \,
			\displaystyle\frac{\left\{ (\mathbf{x}_s-\mathbf{x}^j) \cdot \mathbf{n}^j \right\}_+}{\|\mathbf{x}_s-\mathbf{x}^j\|^3}}
			{\displaystyle\sum_{j=1}^{q} \sum_{\mathbf{p} \in \Omega^j}
			\left[ \displaystyle\frac{\left\{(\mathbf{x}_s-\mathbf{x}^j) \cdot \mathbf{n}^j \right\}_+}{\|\mathbf{x}_s-\mathbf{x}^j\|^3} \right]^2}.
	\label{eq:22}
	\end{equation}
	
	\item Otherwise (if $\mu >0$), Eq.\ \eqref{eq:20} can be rewritten
	\begin{equation}
		\underbrace{\Psi^{\frac{1}{\mu}} \, \mathbf{n}_s}_{\mathbf{m}_s} \cdot \,(\mathbf{x}^j\!-\!\mathbf{x}_s)
			\!=\! \left[\! \displaystyle I^j(\mathbf{p}) \,
			\frac{\|\mathbf{x}_s-\mathbf{x}^j\|^{3+\mu}}{\left\{(\mathbf{x}_s\!-\!\mathbf{x}^j) \!\cdot\! \mathbf{n}^j \! \right\}_+}  \right]^{\frac{1}{\mu}}\!\!.
	\label{eq:23}
	\end{equation}
	The least-squares estimation of vector $\mathbf{m}_s$ {defined in \eqref{eq:23}} is thus written
	\begin{equation}
		\underset{\mathbf{m}_s}{\operatorname{\min}~} \sum_{j=1}^{q}
			\sum_{\mathbf{p} \in \Omega^j}
			\!\left[\! \mathbf{m}_s \cdot (\mathbf{x}^j\!-\!\mathbf{x}_s)
			\!-\! \left[\! \displaystyle I^j(\mathbf{p}) \,
			\frac{\|\mathbf{x}_s\!-\!\mathbf{x}^j\|^{3+\mu}}{ \left\{(\mathbf{x}_s\!-\!\mathbf{x}^j) \cdot \mathbf{n}^j\right\}_+} \right]^{\frac{1}{\mu}} \right]^2.
	\label{eq:24}
	\end{equation}
	This linear least-squares problem can be solved using the pseudo-inverse. From this estimate, we easily deduce those of parameters {$\mathbf{n}_s$ and~$\Psi$}:
	\begin{equation}
		\mathbf{n}_s = \frac{\mathbf{m}_s}{\|\mathbf{m}_s\|}
		\qquad \text{and} \qquad
		\Psi = \|\mathbf{m}_s\|^\mu.
	\label{eq:25}
	\end{equation}
\end{itemize}

In both cases, it is impossible to deduce from the estimate of $\Psi$ that of $\Phi_0$, because in the definition \eqref{eq:19} of $\Psi$, the product $\gamma \, \beta \, \frac{\rho_0}{\pi}$ is unknown. However, since this product is the same for all LEDs ({deactivating} all automatic settings of the camera makes $\beta$ and $\gamma$ constant), all the intensities $\Phi_0^i$, $i \in \{1,\dots,m\}$, are estimated up to a common factor.

Fig.\ \ref{fig:5} shows a schematic representation of the experimental \colorme{setup} of Fig.\ \ref{fig:2}-a, where the LEDs parameters were estimated {using our calibration procedure}.

\begin{figure}[!htpb]
\centering
\begin{tabular}{c}
  \includegraphics[width = 0.9\linewidth]{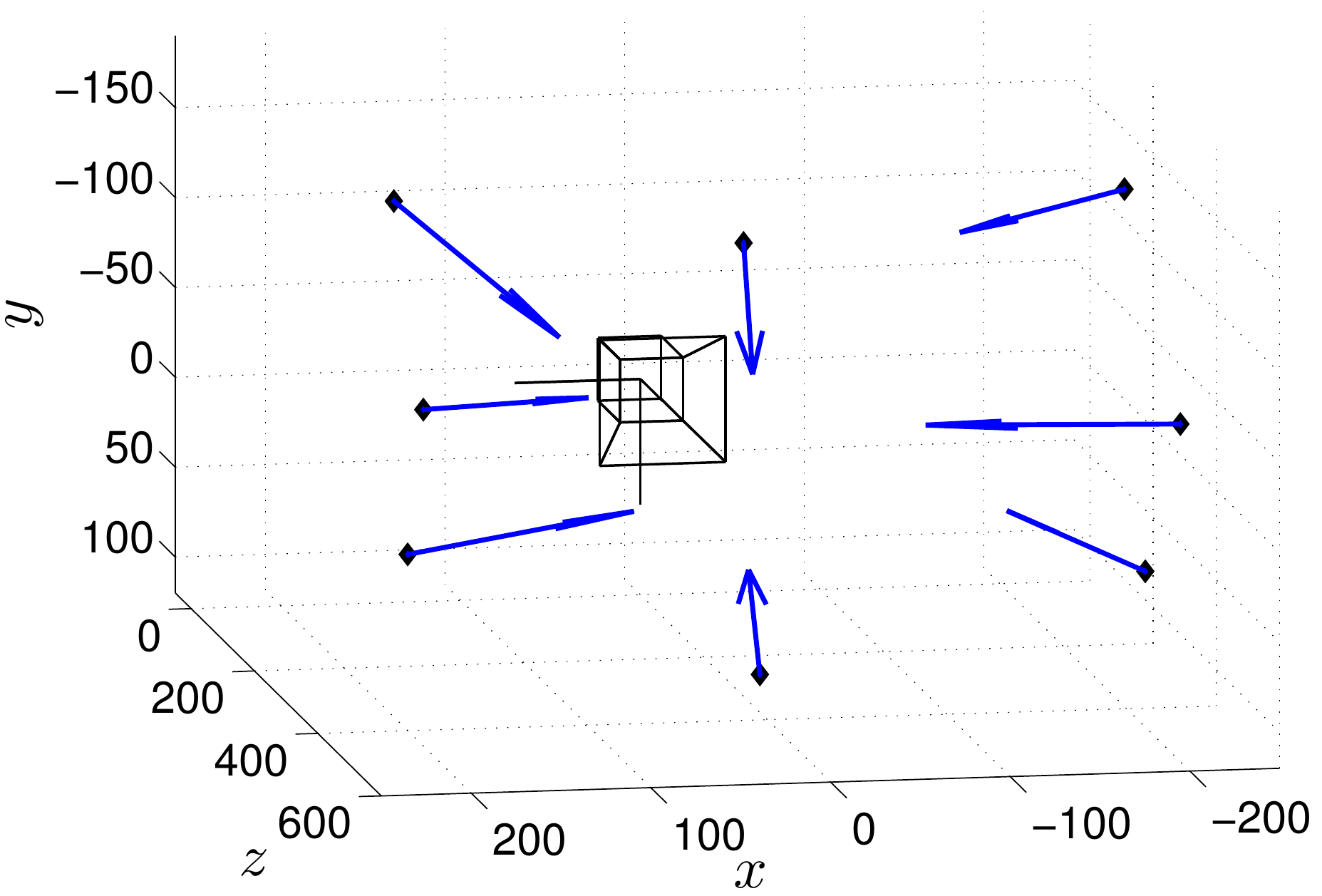} \\
	(a) \\
	\includegraphics[width = 0.9\linewidth]{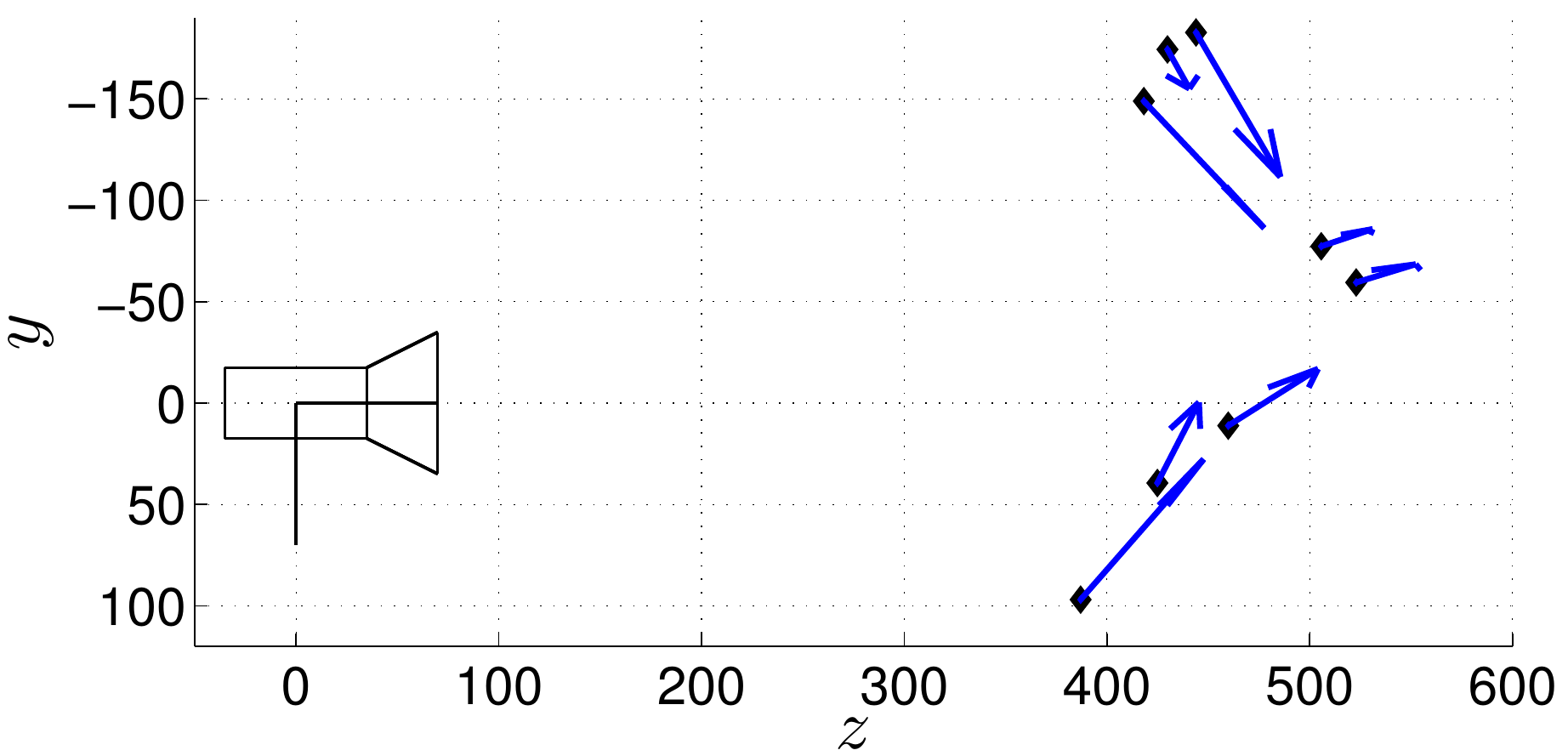} \\
	(b)
\end{tabular}
\caption{Two views of a schematic representation of the experimental \colorme{setup} of Fig.\ \ref{fig:2}-a. {The camera center is located in $(0,0,0)$. A black marker characterizes the location $\mathbf{x}_s$ of {each} LED (unit {$mm$}), the {orientation} of a blue arrow its {principal} direction $\mathbf{n}_s$, and the length of this arrow its intensity $\Phi_0$ (up to a {common} factor).}}
\label{fig:5}
\end{figure}


\subsection{Modeling Photometric Stereo with Point Light Sources}
\label{sec:2.3}

If the luminous flux emitted by a LED is described by {Model} \eqref{eq:12}, then we obtain from \eqref{eq:18} and \eqref{eq:19} the following equation for the gray level at pixel $\mathbf{p}$:
\begin{equation}
	I(\mathbf{p}) = \Psi \, \frac{\rho(\mathbf{p})}{\rho_0}
		\!\left[  \frac{\mathbf{n}_s \!\cdot\! \left(\!\mathbf{x}\!-\!\mathbf{x}_s \!\right)} {\|\mathbf{x}\!-\!\mathbf{x}_s\|} \! \right]^\mu
		\frac{\left\{(\mathbf{x}_s\!-\!\mathbf{x}) \!\cdot\! \mathbf{n}(\mathbf{p})\right\}_+}{\|\mathbf{x}_s\!-\!\mathbf{x}\|^3}.
\label{eq:26}
\end{equation}
Let us introduce a new definition of the albedo {relative} to the albedo $\rho_0$ of the Lambertian planar calibration pattern:
\begin{equation}
	\overline{\rho}(\mathbf{p}) = \frac{\rho(\mathbf{p})}{\rho_0}.
\label{eq:28}
\end{equation}
By writing Eq.~\eqref{eq:26} with respect to each LED, and by using Eq.~\eqref{eq:28}, we obtain, in each pixel $\mathbf{p}\in\Omega$, the system of equations~\eqref{eq:model}, for $i\in \{1,\dots,m\}$.

~\\


To solve this system, the introduction of the auxiliary variable $\overline{\mathbf{m}}(\mathbf{p}) = \overline{\rho}(\mathbf{p})\, \mathbf{n}(\mathbf{p})$ may seem relevant, since this vector is not constrained to have unit-length, but we will see that this trick loses part of its interest. Defining the following $m$ vectors, $i\in \{1,\dots,m\}$:
\begin{equation}
	\mathbf{t}^i(\mathbf{x}) = \Psi^i
		\left[ \frac{\mathbf{n}^i_s \cdot  \left(\mathbf{x}-\mathbf{x}^i_s\right)} {\|\mathbf{x}-\mathbf{x}^i_s\|} \right]^{\mu^i}
		\frac{\mathbf{x}^i_s-\mathbf{x}}{\|\mathbf{x}^i_s-\mathbf{x}\|^3},
\label{eq:29}
\end{equation}
and neglecting self-shadows ($\{x\}_+ = x$), then System~\eqref{eq:model} is rewritten
in matrix form:
\begin{equation}
	\mathbf{I}(\mathbf{p}) = \mathbf{T}(\mathbf{x}) \, \overline{\mathbf{m}}(\mathbf{p}),
\label{eq:31}
\end{equation}
where $\mathbf{I}(\mathbf{p})\in\mathbb{R}^m$ has been defined in \eqref{eq:5} and $\mathbf{T}(\mathbf{x})\in\mathbb{R}^{m\times3}$ is defined as follows:
\begin{equation}
	\mathbf{T}(\mathbf{x}) =
		\begin{bmatrix}
			\mathbf{t}^{1}(\mathbf{x})^\top \\
			\vdots \\
			\mathbf{t}^{m}(\mathbf{x})^\top
		\end{bmatrix}.
\label{eq:32}
\end{equation}
Eq.\ \eqref{eq:31} is similar to \eqref{eq:4}. {Knowing} the matrix field $\mathbf{T}(\mathbf{x})$ would allow us to estimate its field of pseudo-inverses in order to solve \eqref{eq:31}, just as calculating the pseudo-inverse of $\mathbf{S}$ allows us to solve \eqref{eq:4}. However, the matrix field $\mathbf{T}(\mathbf{x})$ depends on $\mathbf{x}$, and thus on the \emph{unknown depth}. This {simple} difference induces major changes when it comes to the {numerical \colorme{solution}}, as discussed in the next two sections. 



\section{A Review of Two Variational Approaches for Solving Photometric Stereo under Point Light Source Illumination, with New Insights}
\label{sec:3}


In this section, we study two variational approaches from the literature for solving photometric stereo under point light source illumination.

The first one inverts the nonlinear image formation model by recasting it as a sequence of simpler subproblems~\cite{Ahmad2014,Bony2013,Collins2012,Huang2015,Kolagani1992,Nie2016b,Papadhimitri2014b,Yeh2016}. It consists in estimating the normals and the albedo, assuming that the depth map is fixed, then integrating the normals into a new depth map, and to iterate. We show in Section~\ref{sec:3.1} how to improve this standard method in order to estimate \emph{absolute} depth, without resorting to any prior. 

The second approach first linearizes the image formation model by resorting to image ratios, then directly estimates the depth by solving the resulting system of PDEs in an approximate manner~\cite{Logothetis2017,Logothetis_2016,Mecca2016,CVPR2016}. We show \colorme{in Section~\ref{sec:3.2}} that state-of-the-art solutions, which resort to fixed point iterations, may be trapped in local minima. This shortcoming can be avoided by rather using an augmented Lagrangian algorithm.

As in these state-of-the-art methods, self-shadows will be neglected throughougt this section i.e., we abusively assume $\{x\}_+ = x$. To enforce robustness, we simply follow the approach advocated in~\cite{Bringier2012}, which systematically eliminates, in each pixel, the highest gray level, which may come from a specular highlight, as well as {the two lowest ones}, which may correspond to shadows. More elaborate methods for ensuring robustness will be discussed in Section~\ref{sec:4}.

Apart from robustness issues, we will see that the state-of-the-art methods studied in this section remain unsatisfactory, because their convergence is not established.


\subsection{Scheme Inspired by the Classical \colorme{Numerical Solution of Photometric Stereo}}
\label{sec:3.1}

{For solving Problem \eqref{eq:31}}, it seems quite natural to \colorme{adapt} the {solution} \eqref{eq:6} of the linear model~\eqref{eq:4}. To linearize \eqref{eq:31}, \colorme{we have} to assume that matrix $\mathbf{T}(\mathbf{x})$ is known. If we proceed iteratively, this can be made possible by replacing, at iteration $(k+1)$, $\mathbf{T}(\mathbf{x})$ by $\mathbf{T}(\mathbf{x}^{(k)})$. This very simple idea has led to several {numerical solutions} \cite{Ahmad2014,Bony2013,Collins2012,Huang2015,Kolagani1992,Nie2016b,Papadhimitri2014b,Yeh2016}, which all require some kind of a priori knowledge on the depth. On the contrary, the scheme we propose here requires none, which constitutes a significant improvement. 
This new scheme consists in the following algorithm:
~\\

\begin{algorithm}[alternating approach] {\ }
\label{alg:stota_1}
\begin{algorithmic}[1] 
\STATE Initialize $\mathbf{x}^{(0)}$. Set $k:=0$.

\LOOP

\STATE
Solve Problem \eqref{eq:31} in the least-squares sense {in} each $\mathbf{p}\in\Omega$, replacing $\mathbf{T}(\mathbf{x})$ by $\mathbf{T}(\mathbf{x}^{(k)})$, which provides a new estimation of $\overline{\mathbf{m}}(\mathbf{p})$:
	\begin{equation}
		\overline{\mathbf{m}}^{(k+1)}(\mathbf{p}) = \mathbf{T}(\mathbf{x}^{(k)})^\dagger \, \mathbf{I}(\mathbf{p}).
	\label{eq:33}
	\end{equation}

\STATE 
Deduce a new estimation of the normal $\mathbf{n}(\mathbf{p})$: 
	\begin{equation}
		\mathbf{n}^{(k+1)}(\mathbf{p}) = \frac{\overline{\mathbf{m}}^{(k+1)}(\mathbf{p})}{\| \overline{\mathbf{m}}^{(k+1)}(\mathbf{p}) \|}. 
	\label{eq:34} 
	\end{equation}

\STATE
Integrate the new normal field $\mathbf{n}^{(k+1)}$ into an updated {3D-shape} $\mathbf{x}^{(k+1)}$, up to a scale factor.

\STATE 
Estimate this scale factor by nonlinear optimization.

\STATE
Set $k:=k+1$ as long as $k < k_{\text{max}}$.

\ENDLOOP

\STATE $
  \overline{\rho}(\mathbf{p}) = \left\| \overline{\mathbf{m}}^{(k_{\text{max}})}(\mathbf{p}) \right\|. \hfill
 \inlineeqno$

\end{algorithmic}
\end{algorithm}

%
%
%
%
%
%

For this scheme to be \colorme{completely specified}, we need to set the initial {3D-shape} $\mathbf{x}^{(0)}$. {We use as initial guess a fronto-parallel plane at distance $z_0$ from the camera, $z_0$ being a rough estimate of the \colorme{mean} distance from the camera to the scene surface. }

\paragraph{Integration of Normals.}
\colorme{Stages 3 and 4 of the scheme above are trivial and can be achieved pixelwise, but Stages 5 and 6 are trickier.}
From the equalities in~\eqref{eq:1}, and by denoting $\nabla z(\mathbf{p}) = \left[\partial_u z(\mathbf{p}),\partial_v z(\mathbf{p})\right]^\top$ the gradient of $z$ {in} $\mathbf{p}$, it is easy to deduce that the (non-unit-length) vector
\begin{equation}
	\overline{\mathbf{n}}(\mathbf{p}) =
		\begin{bmatrix}
			f \, \partial_u z(\mathbf{p}) \\
			f \, \partial_v z(\mathbf{p}) \\
			-z(\mathbf{p}) - \mathbf{p} \cdot \nabla z(\mathbf{p})
		\end{bmatrix}
\label{eq:35}
\end{equation}
is normal to the surface. Expression \eqref{eq:35} shows that integrating the (unit-length) normal field $\mathbf{n}$ allows to estimate the depth $z$ only up to a scale factor $\kappa \in \mathbb{R}$, since:
\begin{equation}
	\mathbf{n}(\mathbf{p}) \propto
		\begin{bmatrix}
			f \, \partial_u z(\mathbf{p}) \\
			f \, \partial_v z(\mathbf{p}) \\
			-z(\mathbf{p}) \!-\! \mathbf{p} \!\cdot\! \nabla z(\mathbf{p})
		\end{bmatrix} \propto
		\begin{bmatrix}
			f \, \partial_u (\kappa \, z)(\mathbf{p}) \\
			f \, \partial_v (\kappa \, z)(\mathbf{p}) \\
			- (\kappa \, z)(\mathbf{p}) \!-\! \mathbf{p} \!\cdot\! \nabla (\kappa \, z)(\mathbf{p})
		\end{bmatrix}.
\label{eq:36}
\end{equation}
The collinearity of $\overline{\mathbf{n}}(\mathbf{p})$ and $\mathbf{n}(\mathbf{p}) = [n_1(\mathbf{p}),n_2(\mathbf{p}),n_3(\mathbf{p})]^\top$ leads to the system
\begin{equation}
	\begin{cases}
		n_3(\mathbf{p}) \, f \, \partial_u z(\mathbf{p}) \!+\! n_1(\mathbf{p})  \left[
		z(\mathbf{p}) \!+\! \mathbf{p} \cdot \! \nabla z(\mathbf{p}) \right] = 0, \\
		n_3(\mathbf{p}) \, f \, \partial_v z(\mathbf{p}) \!+\! n_2(\mathbf{p})  \left[ z(\mathbf{p}) \!+\! \mathbf{p}
		\cdot \! \nabla z(\mathbf{p}) \right] = 0,
	\end{cases}
\label{eq:37}
\end{equation}
which is homogeneous in $z(\mathbf{p})$. Introducing the \colorme{change of} variable $\tilde{z} = \log (z)$, \colorme{which is valid since $z > 0$,~\eqref{eq:37} is rewritten}
\begin{equation}
  \begin{cases}
    \left[f \, n_3(\mathbf{p}) + u \, n_1(\mathbf{p}) \right] \partial_u \tilde{z}(\mathbf{p}) + v \, n_1(\mathbf{p}) \partial_v \tilde{z}(\mathbf{p}) = - n_1(\mathbf{p}), \\
   u \, n_2(\mathbf{p}) \partial_u \tilde{z}(\mathbf{p}) +   \left[f \, n_3(\mathbf{p}) + v \, n_2(\mathbf{p}) \right] \partial_v \tilde{z}(\mathbf{p})   = - n_2(\mathbf{p}).
  \end{cases}
  \label{eq:38}
\end{equation}
\colorme{The determinant of this system is equal to 
\begin{equation}
  f \, n_3(\mathbf{p}) \left[u \, n_1(\mathbf{p}) \!+\! v \, n_2(\mathbf{p}) \!+\! f \, n_3(\mathbf{p})\right] 
  =
 f \, n_3(\mathbf{p}) \left[ \overline{\mathbf{p}} \cdot \mathbf{n}(\mathbf{p})\right],
\end{equation}
if we denote} 
\begin{equation}
	\overline{\mathbf{p}} = [u,v,f]^\top.
\label{eq:39}
\end{equation}
It is \colorme{then} easy to deduce \colorme{the solution of~\eqref{eq:38}:}
\begin{equation}
	\nabla \tilde{z}(\mathbf{p}) =
		- \frac{1}{\overline{\mathbf{p}} \cdot \mathbf{n}(\mathbf{p})}
		\begin{bmatrix}
			n_1(\mathbf{p}) \\
			n_2(\mathbf{p})
		\end{bmatrix}.
\label{eq:40}
\end{equation}

Let us now come back to Stages 5 and 6 of Algorithm~\ref{alg:stota_1}. The \colorme{new} normal field is $\mathbf{n}^{(k+1)}(\mathbf{p})$, from which we {can} deduce the gradient $\nabla \tilde{z}^{(k+1)}(\mathbf{p})$ thanks to Eq. \eqref{eq:40}. By integrating this gradient between a pixel $\mathbf{p}_0$, chosen arbitrarily inside $\Omega$, and {any pixel $\mathbf{p}\in\Omega$}, and knowing that $z = \exp\{\tilde{z}\}$, we obtain:
\begin{equation}
	z^{(k+1)}(\mathbf{p}) \!=\! z^{(k+1)}(\mathbf{p}_0) \, \exp\!\left\{ \! \int_{\mathbf{p}_0}^{\mathbf{p}} \!\!\!\! \nabla \tilde{z}^{(k+1)}(\mathbf{q}) \cdot \mathrm{d}\mathbf{q}\!\right\}.
\label{eq:41}
\end{equation}
This integral can be calculated along one single path \colorme{inside $\Omega$} going from $\mathbf{p}_0$ to $\mathbf{p}$, but since the gradient field $\nabla \tilde{z}^{(k+1)}(\mathbf{p})$ is never rigorously integrable {in practice}, this calculus usually depends on the choice of the path \cite{Wu1988}. The most common \colorme{parry} to this well-known problem consists in {resorting to} a variational approach{, see for instance \cite{Durou2016} for some discussion}.

Expression \eqref{eq:41} confirms that the depth can {only be calculated, from $\mathbf{n}^{(k+1)}(\mathbf{p})$, up to} a scale factor equal to $z^{(k+1)}(\mathbf{p}_0)$. Let us determine this scale factor by minimization of the reprojection error of Model \eqref{eq:31} over the entire domain $\Omega$. Knowing that, from \eqref{eq:1} and~\eqref{eq:39}, we get $\mathbf{x} = \frac{z}{f}\, \overline{\mathbf{p}}$, this comes down to solving the following nonlinear least-squares problem:
\begin{align}
	& z^{(k+1)}(\mathbf{p}_0) =  \underset{w \, \in \, \mathbb{R}^+}{\arg\min~} \mathcal{E}_{\mathrm{alt}}(w):=
	\sum_{\mathbf{p} \in \Omega}
	\Big\|\mathbf{I}(\mathbf{p}) \nonumber \\
	& \quad -\! \mathbf{T} \Big(\frac{w}{f} \exp\left\{\int_{\mathbf{p}_0}^{\mathbf{p}} \nabla \tilde{z}^{(k+1)}(\mathbf{q})
	\cdot \mathrm{d}\mathbf{q}\right\} \overline{\mathbf{p}} \Big) \,
	\overline{\mathbf{m}}^{(k+1)}(\mathbf{p}) \Big\|^2,
\label{eq:42}
\end{align}
which allows us to eventually write the 3D-shape update (Stages 5 and 6):
\begin{equation}
	\mathbf{x}^{(k+1)} \!=\! \frac{z^{(k+1)}(\mathbf{p}_0)}{f} \, \exp\!\left\{\!\int_{\mathbf{p}_0}^{\mathbf{p}}\!\!\!\!
	\nabla \tilde{z}^{(k+1)}(\mathbf{q}) \cdot \mathrm{d}\mathbf{q}\!\right\} \overline{\mathbf{p}}.
\label{eq:43}
\end{equation}

\paragraph{Experimental Validation.}
\colorme{Despite the lack of theoretical guarantee, convergence of this scheme is empirically observed, provided that the initial 3D-shape $\mathbf{x}^{(0)}$ is not too distant from the scene surface.}
For the curves in Fig.~\ref{fig:6}, several fronto-parallel planes with equation $z\equiv z_0$ were \colorme{tested} as initial guess. The mean distance {from the camera} to the scene being approximately $700~mm$, it is not surprising that the fastest convergence is observed for this value of $z_0$. \colorme{Besides, this graph also shows that under-estimating the initial scale quite a lot is not a problem, whereas over-estimating it severely slows down the process.}

\begin{figure}[!htpb]
\begin{center}
	\includegraphics[width = 0.7\linewidth]{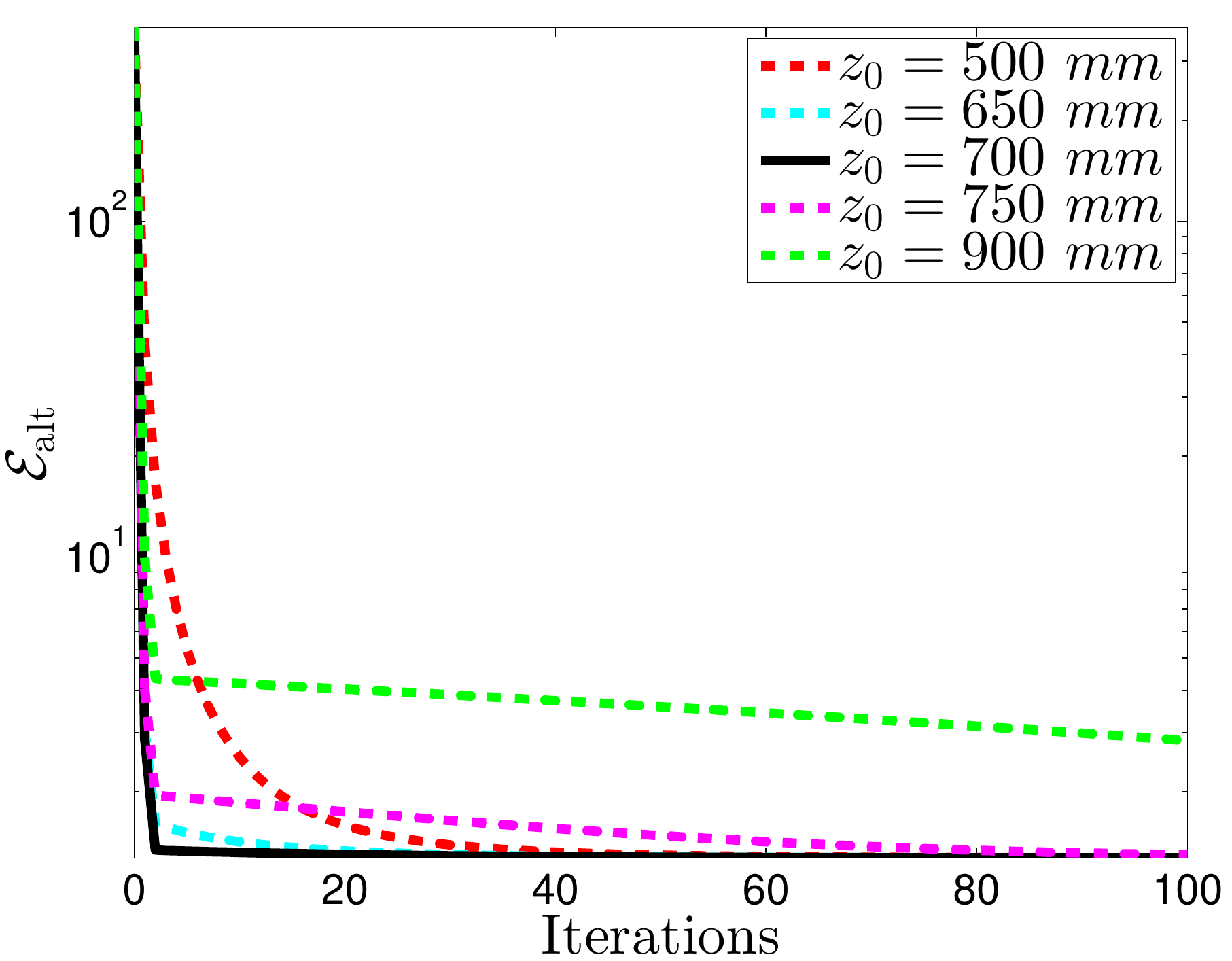}
\end{center}
\caption{Evolution of the energy $\mathcal{E}_{\mathrm{alt}}$ \colorme{of the alternating approach}, defined in \eqref{eq:42}, in function of the iterations, when the initial 3D-shape is a fronto-parallel plane with equation {$z\equiv z_0$}. The used data are the $m=8$ images of the plaster statuette of Fig.\ \ref{fig:2}. {The proposed scheme {consists} in alternating normal estimation, normal integration and scale estimation (cf. Algorithm~\ref{alg:stota_1})}. {It} converges towards the same solution (at different speeds), for the five tested values of $z_0$.}
\label{fig:6}
\end{figure}

Fig.\ \ref{fig:7} allows to compare the 3D-shape obtained by photometric stereo, from sub-images of size $920\times1178$ in full resolution (bounding box of the statuette), which contain $773 794$ pixels inside $\Omega$, with the ground truth obtained by laser scanning, which contains $1 753 010$ points. The points density is thus almost the same on the front of the statuette, since we {did} not reconstruct its back. However, our result is achieved in less than ten seconds (five iterations of a Matlab code on a recent i7 processor), instead of several hours for the ground truth, while we also estimate the albedo.

\begin{figure}[!htpb]
\begin{center}
	\begin{tabular}{ccc}
		\includegraphics[width = 0.235\linewidth]{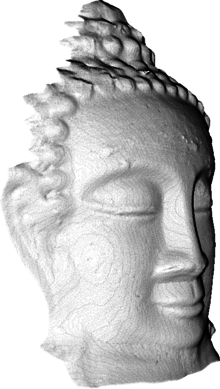} & 
		\includegraphics[width = 0.33\linewidth]{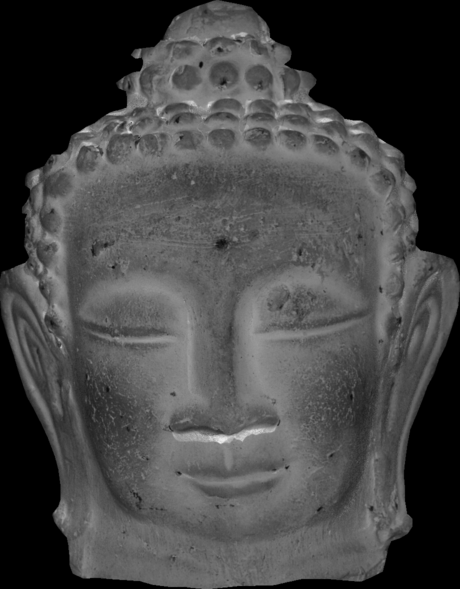} & 
		\includegraphics[width = 0.285\linewidth]{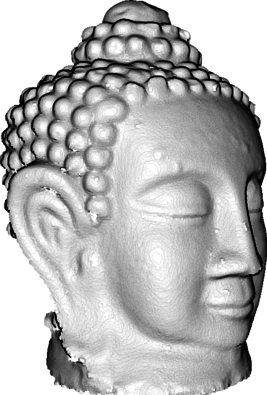} \\
		\small{(a)} &  \small{(b)} &  \small{(c)}
	\end{tabular}
\end{center}
\caption{(a) {3D-reconstruction} and {(b)} albedo obtained with Algorithm~\ref{alg:stota_1}. (c) Ground truth 3D-shape obtained by laser scanning. {Photometric stereo not only provides a {3D-shape} qualitatively similar to the laser scan, but also provides the albedo.}}
\label{fig:7}
\end{figure}

\begin{figure}[!htpb]
\begin{center}
	\begin{tabular}{c}
		\def\svgwidth{0.85\linewidth}
		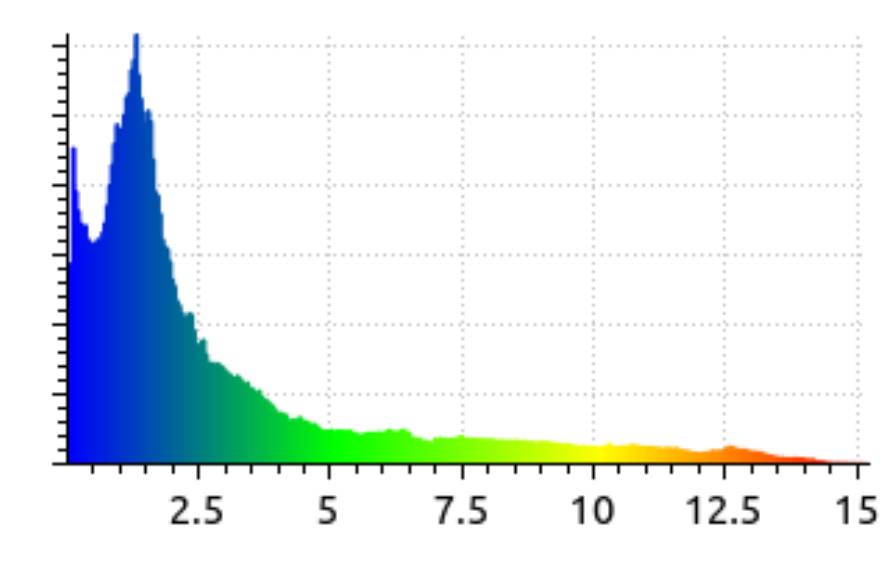 \\
		\small{(a)} \\
		\includegraphics[height = 0.45\linewidth]{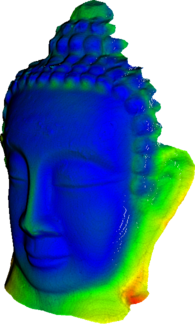} \\
		\small{(b)}
	\end{tabular}
\end{center}
\caption{(a) Histogram of point-to-point distances between the \colorme{alternating} 3D-reconstruction and the ground truth (cf.~Fig.~\ref{fig:7}). The median value is $1.3~mm$. (b) Spatial distribution of these distances. {The histogram peak is not located in zero. As we will see in Section~\ref{sec:3.2}, this bias can be avoided by resorting to a differential approach based on PDEs.}}
\label{fig:8}
\end{figure}

Fig.\ \ref{fig:8}-a shows the histogram of point-to-point distances between our result (Fig.\ \ref{fig:7}-a) and the ground truth (Fig.\ \ref{fig:7}-c). The median value is $1.3~mm$. The spatial distribution of these distances (Fig.\ \ref{fig:8}-b), shows that the largest distances are observed on the highest slopes of the surface. This clearly comes from the facts that, even for a diffuse material such as plaster, the Lambertian model is not valid under skimming lighting, and that self-shadows were neglected.

More realistic reflectance models, such as the one proposed by Oren and Nayar in \cite{Oren1995}, would perhaps improve accuracy of the 3D-reconstruction in such points, and we will see in Section~\ref{sec:4} how to handle self-shadows. But, as we shall see now, bias also comes from normal integration. In the next section, we describe a different formulation of photometric stereo which permits to avoid integration, by solving a system of PDEs in $z$.


\subsection{\colorme{Direct Depth Estimation using Image Ratios}}
\label{sec:3.2}

The scheme proposed in Section \ref{sec:3.1} suffers from several defects. It requires to integrate the gradient $\nabla \tilde{z}^{(k+1)}(\mathbf{p})$ at each iteration. This {is} not \colorme{achieved} by the naive formulation \eqref{eq:42}, {but using} more sophisticated methods which allow to overcome the problem of non-integrability \cite{Durou2009}. \colorme{Still, bias due to inaccurate normal estimation should not have to be corrected during integration. }
\colorme{Instead, it seems more justified to directly estimate the depth map, without resorting to intermediate normal estimation. This can be achieved by recasting photometric stereo as a system of quasilinear PDEs.} 

\paragraph{Differential Reformulation of Problem \eqref{eq:31}.}
Let us recall (cf.~Eq.~\eqref{eq:1}) that the {coordinates of the 3D-point} $\mathbf{x}$ conjugate to a pixel $\mathbf{p}$ {are} completely characterized by the depth $z(\mathbf{p})$:
\begin{equation}
	\mathbf{x} = \frac{z(\mathbf{p})}{f} \,
		\begin{bmatrix}
			\mathbf{p} \\
			f
		\end{bmatrix}.
\label{eq:44}
\end{equation}
The vectors $\mathbf{t}^i(\mathbf{x})$ defined in \eqref{eq:29} thus depend on the unknown depth values $z(\mathbf{p})$. Using once again the change of variable $\tilde{z} = \log(z)$\footnote{Without this change of variable, one would obtain a system of homogeneous PDEs in lieu of~\eqref{eq:54}, which would need regularization to be solved, see~\cite{CVPR2016}.}, we consider from now on each~$\mathbf{t}^i$, $i \in \{1,\dots,m\}$, as a vector field depending on the unknown map $\tilde{z}$:
\begin{equation}
\begin{array}{rccl}
\mathbf{t}^i(\tilde{z}):& \Omega & \to & \mathbb{R}^3 \\[-0.5em]
 & \mathbf{p} & \mapsto & \mathbf{t}^i(\tilde{z})(\mathbf{p}) =  \Psi^i \left[ - \frac{ \mathbf{n}_s^i \cdot  \mathbf{v}^i(\tilde{z})(\mathbf{p})}{\|\mathbf{v}^i(\tilde{z})(\mathbf{p})\|}\right]^{\mu^i}
 \frac{\mathbf{v}^i(\tilde{z})(\mathbf{p})}{\|\mathbf{v}^i(\tilde{z})(\mathbf{p})\|^3},
\end{array}
\label{eq:t_i}
\end{equation}
 where each field $\mathbf{t}^i(\tilde{z})$ depends in a nonlinear way on the unknown (log-) depth map $\tilde{z}$, through the following vector field:
\begin{equation}
\begin{array}{rccl}
\mathbf{v}^i(\tilde{z}):& \Omega & \to & \mathbb{R}^3 \\[-0.5em]
 & \mathbf{p} & \mapsto & \mathbf{v}^i(\tilde{z})(\mathbf{p}) =  \mathbf{x}^i_s
-
\frac{\exp\left(\tilde{z}(\mathbf{p})\right)}{f} \,
		\begin{bmatrix}
			\mathbf{p} \\
			f
		\end{bmatrix}.
\end{array}
\end{equation}
Knowing that the (non-unit-length) vector $\overline{\mathbf{n}}(\mathbf{p})$ defined in \eqref{eq:35}, divided by $z(\mathbf{p})$, is normal to the surface, and still neglecting self-shadows, we can rewrite System~\eqref{eq:model}{, in each pixel $\mathbf{p}\in\Omega$:}
\begin{align}
&I^i(\mathbf{p}) = \frac{\overline{\rho}(\mathbf{p})}{d(\tilde{z})(\mathbf{p})} \,
 \mathbf{t}^i(\tilde{z})(\mathbf{p})  \cdot
 \begin{bmatrix}
		f \nabla \tilde{z}(\mathbf{p}) \\
		-1 - \mathbf{p} \cdotp \nabla \tilde{z}(\mathbf{p})
	\end{bmatrix}, \nonumber \\
& \qquad\qquad\qquad\qquad\qquad\qquad\qquad	~ i\in \{1,\dots,m\},
	\label{eq:3.16}
\end{align}
with
\begin{equation}
    d(\tilde{z})(\mathbf{p}) =  \sqrt{f^2 \left\| \nabla \tilde{z}(\mathbf{p}) \right\|^2 + \left(-1 - \mathbf{p} \cdotp \nabla \tilde{z}(\mathbf{p}) \right)^2  }.   \label{eq:50} 
\end{equation}
}

\paragraph{Partial Linearization of \eqref{eq:3.16} using Image Ratios.}

In comparison with Eqs.\ \eqref{eq:model}, the PDEs \eqref{eq:3.16} explicitly depend on the \colorme{unknown map $\tilde{z}$, and thus remove the need for alternating normal estimation and integration. However, these equations contain} two difficulties: they are nonlinear and cannot be solved locally. We can eliminate the nonlinearity due to the {coefficient of normalization} $d(\tilde{z})(\mathbf{p})$. Indeed, neither the relative albedo $\overline{\rho}(\mathbf{p})$, nor this coefficient, depend on the index~$i$ of the LED. We \colorme{deduce from any pair $\{i,j\} \in \{1,\dots,m\}^2$, $i \neq j$, of equations from~\eqref{eq:3.16}, the following equalities:}
\begin{align}
	\frac{\overline{\rho}(\mathbf{p})}{d(\tilde{z})(\mathbf{p})} & = \frac{I^i(\mathbf{p})}{\mathbf{a}^i(\tilde{z})(\mathbf{p}) \cdot \nabla \tilde{z}(\mathbf{p}) - b^i(\tilde{z})(\mathbf{p}) } \nonumber \\
	& = \frac{I^j(\mathbf{p})}{\mathbf{a}^j(\tilde{z})(\mathbf{p}) \cdot \nabla \tilde{z}(\mathbf{p}) - b^j(\tilde{z})(\mathbf{p}) },
\label{eq:51}
\end{align}
{with the following definitions of $\mathbf{a}^i(\tilde{z})(\mathbf{p})$ and $b^i(\tilde{z})(\mathbf{p})$, 
denoting $\mathbf{t}^i(\tilde{z})(\mathbf{p}) = [t^i_{1}(\tilde{z})(\mathbf{p}),t^i_{2}(\tilde{z})(\mathbf{p}),t^i_{3}(\tilde{z})(\mathbf{p})]^\top$}:
\begin{align}
	\mathbf{a}^i(\tilde{z})(\mathbf{p}) & = f
		\begin{bmatrix}
			t^i_{1}(\tilde{z})(\mathbf{p}) \\
			t^i_{2}(\tilde{z})(\mathbf{p})
		\end{bmatrix}
		- t^i_{3}(\tilde{z})(\mathbf{p}) \, \mathbf{p}, \\
	b^i(\tilde{z})(\mathbf{p}) & = t^i_{3}(\tilde{z})(\mathbf{p}).		
\end{align}

From {the equalities} \eqref{eq:51}, we obtain: 
\begin{align}
	& \underbrace{\begin{bmatrix}
		I^i(\mathbf{p}) \, \mathbf{a}^j(\tilde{z})(\mathbf{p}) - I^j(\mathbf{p}) \, \mathbf{a}^i(\tilde{z})(\mathbf{p})
	\end{bmatrix}}_{\mathbf{a}^{i,j}(\tilde{z})(\mathbf{p})} \cdot \, \nabla \tilde{z}(\mathbf{p}) \nonumber \\
	& \qquad =
	\underbrace{\left[I^i(\mathbf{p}) \, b^j(\tilde{z})(\mathbf{p}) - I^j(\mathbf{p}) \,
		b^i(\tilde{z})(\mathbf{p})\right]}_{b^{i,j}(\tilde{z})(\mathbf{p})}.
\label{eq:53}
\end{align}

The fields $\mathbf{a}^{i,j}(\tilde{z})$ and $b^{i,j}(\tilde{z})$ defined in \eqref{eq:53} depend on $\tilde{z}$ but not on $\nabla \tilde{z}${: Eq.~\eqref{eq:53} is thus a \emph{quasi-linear} PDE in $z$ over $\Omega$}. It could be solved by the characteristic strips expansion method \cite{Mecca2015,Mecca2014b} if we were dealing with $m=2$ images only, but using a larger number of images is necessary in order to design a robust 3D-reconstruction method. Since we are provided with $m > 2$ images, we follow {\cite{Gotardo2015,Logothetis2017,Logothetis_2016,Mecca2016,CVPR2016,Smith2016} and write $\binom{m}{2}$ PDEs such as~\eqref{eq:53} formed by} the $\binom{m}{2}$ pairs $\{i,j\} \in \{1,\dots,m\}^2$, $i \neq j$. Forming the matrix field $\mathbf{A}(\tilde{z}):\,\Omega \to \mathbb{R}^{\binom{m}{2} \times 2}$ by concatenation of the row vectors $\mathbf{a}^{i,j}(\tilde{z})(\mathbf{p})^\top$, and the vector field $\mathbf{b}(\tilde{z}):\,\Omega \to \mathbb{R}^{\binom{m}{2}}$ by concatenation of the scalar values $b^{i,j}(\tilde{z})(\mathbf{p})$, the system of PDEs to solve is written:
\begin{equation}
	\mathbf{A}(\tilde{z}) \, \nabla \tilde{z} = \mathbf{b}(\tilde{z}) \quad \text{over~} \Omega.
\label{eq:54}
\end{equation}

This new differential formulation of photometric stereo seems simpler than the original differential formulation~\eqref{eq:3.16}, since the main source of nonlinearity, due to the denominator $d(\tilde{z})(\mathbf{p})$, has been eliminated. However, it still presents two difficulties. First, the PDEs~\eqref{eq:54} are generally incompatible and hence do not admit an exact solution. It is thus necessary to estimate an approximate one, by resorting to a variational approach. Assuming that each of the $\binom{m}{2}$ equalities in System \eqref{eq:54} is satisfied up to an additive, zero-mean, Gaussian noise\footnote{In fact, any noise assumption should be formulated on the images, and not on Model \eqref{eq:54}, which was obtained by considering ratios of gray levels: if the noise on gray levels is Gaussian, then that on ratios is Cauchy-distributed \cite{Hinkley1969}. Hence, the least-squares solution \eqref{eq:55} is the best linear unbiased estimator, but it is not the optimal solution.}, one should estimate such a solution by solving the following variational problem:
\begin{equation}
	\underset{\tilde{z}: \Omega \rightarrow\mathbb{R}}{\min~} \,\mathcal{E}_{\mathrm{rat}}(\tilde{z}) := \|\mathbf{A}(\tilde{z}) \, \nabla \tilde{z}
		- \mathbf{b}(\tilde{z}) \|_{L^2(\Omega)}^2.
\label{eq:55}
\end{equation}

 Second, the PDEs~\eqref{eq:53} do not allow to estimate the scale of the scene. Indeed, when all the depth values simultaneously tend to infinity, then both members of~\eqref{eq:53} tend to zero (because the coordinates of $\mathbf{t}^i$ do so, cf.~\eqref{eq:t_i}). Thus, a large, distant 3D-shape will always ``better'' fit these PDEs (in the sense of the criterion $\mathcal{E}_{\mathrm{rat}}$ defined in Eq.~\eqref{eq:55})  than a small, nearby one (cf. Figs.~\ref{fig:11} and~\ref{fig:10}). A ``locally optimal'' solution close to a very good initial estimate should thus be sought.  


\paragraph{Fixed Point Iterations for Solving~\eqref{eq:55}.}

It has been proposed in \cite{Logothetis2017,Logothetis_2016,Mecca2016,CVPR2016} to iteratively estimate a solution of Problem~\eqref{eq:55}, \colorme{by uncoupling} the (linear) estimation of $\tilde{z}$ from the (nonlinear) estimations of $\mathbf{A}(\tilde{z})$ and of $\mathbf{b}(\tilde{z})$. \colorme{This can be achieved} by rewriting \eqref{eq:55} as the following constrained optimization problem:
\begin{equation}
	\begin{array}{l}
		\underset{\tilde{z}: \Omega \rightarrow\mathbb{R}}{\min~} \, \|\mathbf{A} \, \nabla \tilde{z} - \mathbf{b}\|_{L^2(\Omega)}^2 \\
		\text{s.t.~}
		\begin{cases}
			\mathbf{A} &\!\!\!= \mathbf{A}(\tilde{z}), \\
			\mathbf{b} &\!\!\!= \mathbf{b}(\tilde{z}),
		\end{cases} 
	\end{array}
\end{equation}
and resorting to a fixed point iterative scheme:
\begin{align}
	\tilde{z}^{(k+1)} = &~ \underset{\tilde{z}: \Omega \rightarrow\mathbb{R}}{\arg\min~} \|\mathbf{A}^{(k)} \, \nabla \tilde{z} - \mathbf{b}^{(k)} \|_{L^2(\Omega)}^2, \label{eq:57} \\
		\mathbf{A}^{(k+1)} = &~ \mathbf{A}(\tilde{z}^{(k+1)}), \\
		\mathbf{b}^{(k+1)} = &~\mathbf{b}(\tilde{z}^{(k+1)}).
\end{align}

\begin{figure}[!htpb]
\begin{center}
	\includegraphics[width = 0.9\linewidth]{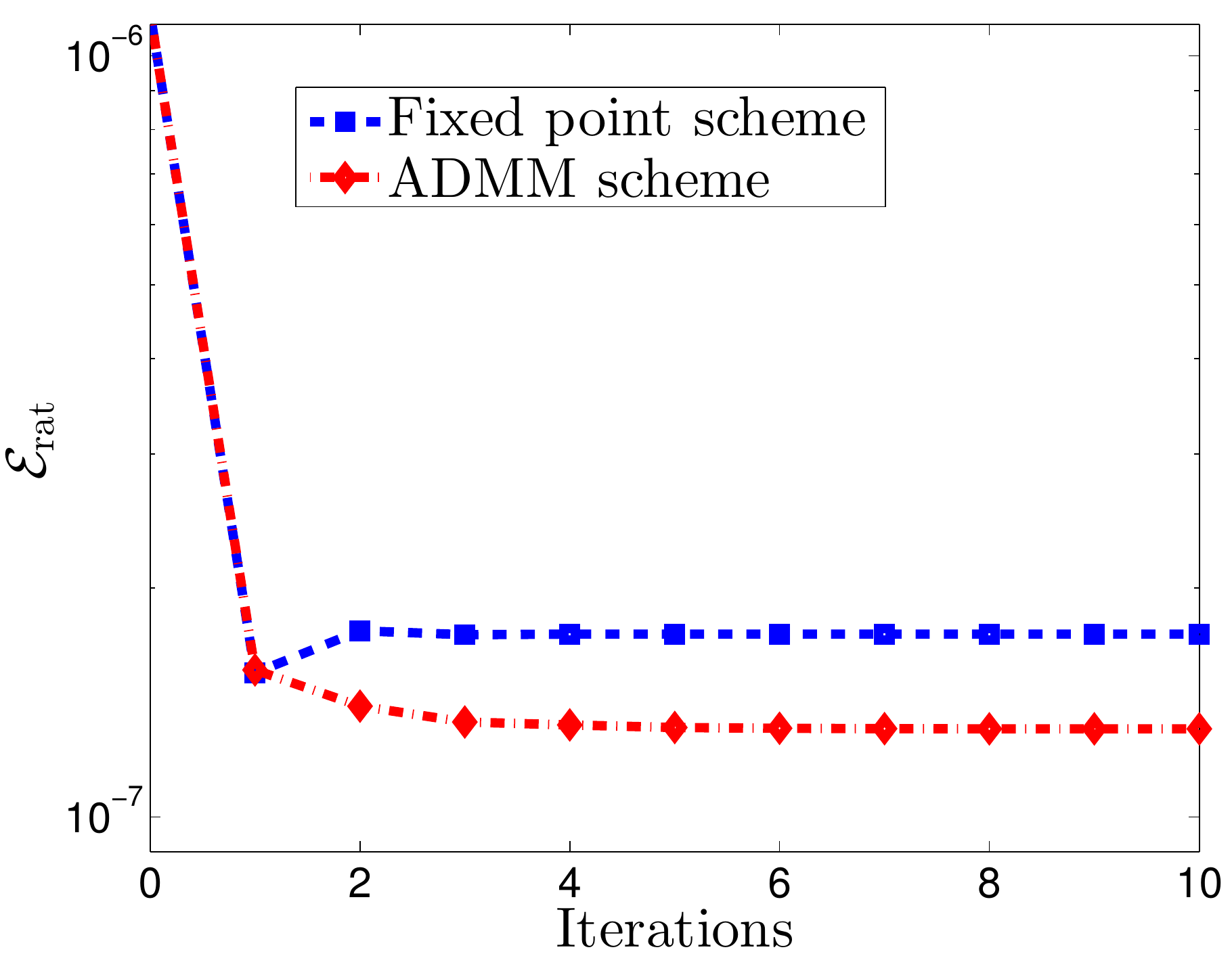}
\end{center}
\caption{Evolution of the energy \colorme{$\mathcal{E}_{\mathrm{rat}}$ of the ratio-based approach}, defined in \eqref{eq:55}, in function of the iterations, for the data of Fig.\ \ref{fig:2} (the initial 3D-shape is a fronto-parallel plane with {equation} $z \equiv 700~mm$). {With the fixed point scheme, the energy is not always decreased after each iteration,} contrarily to the ADMM scheme we are going to introduce.}
\label{fig:9}
\end{figure}

In the linear least-squares variational problem \eqref{eq:57}, the solution can be computed only up to an additive constant. Therefore, the matrix of the system arising from the normal equations associated to the discretized problem will be symmetric, positive, but rank-1 deficient, and thus only semi-definite. Fig.~\ref{fig:9} shows that this may cause \colorme{the fixed point scheme not to decrease the energy after each iteration.} This issue can be resolved by resorting to the alternating direction method of multipliers (ADMM algorithm), a standard procedure which dates back to the {70's} \cite{GabayMercier1976,Glowinski1975}, but has {been revisited recently}~\cite{Boyd2011}. 

\paragraph{\colorme{ADMM Iterations for Solving \eqref{eq:55}.}}


Instead of ``freezing'' the nonlinearities of the variational problem~\eqref{eq:55}, 
 $\tilde{z}$ \colorme{can be estimated} not only from the linearized parts, but also from the nonlinear ones. In this view, we introduce an auxiliary variable $\overline{z}$ and reformulate Problem~\eqref{eq:55} as follows:
\begin{equation}
	\begin{array}{rl}
		& \underset{\overline{z},\tilde{z}}{\min} \left\| \mathbf{A}(\overline{z}) \, \nabla \tilde{z} - \mathbf{b}(\overline{z}) \right\|_{L^2(\Omega)}^2 \\
		& \text{s.t.~} \tilde{z} = \overline{z}.
	\end{array}
\label{eq:59}
\end{equation}

In order to solve the constrained optimization problem \eqref{eq:59}, let us introduce a dual variable $h$ and a descent step $\nu$. A local solution of~\eqref{eq:59} is then obtained at convergence of the following algorithm:

\begin{figure*}[!ht]
\centering
	\begin{tabular}{ccccc}
		\includegraphics[height = 0.3\linewidth]{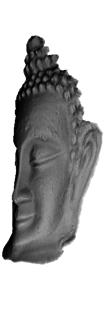} & \,\,
		\includegraphics[height = 0.3\linewidth]{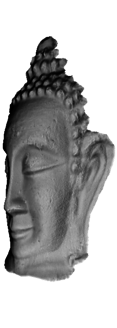} & \,\,
		\includegraphics[height = 0.3\linewidth]{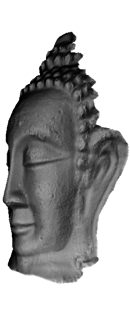} & \,\,
		\includegraphics[height = 0.3\linewidth]{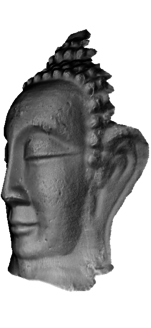} & \,\,
		\includegraphics[height = 0.3\linewidth]{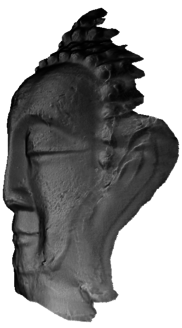} \\
		\small{(a) $z_0 = 500~mm$} & \,\,
		\small{(b) $z_0 = 650~mm$} & \,\,
		\small{(c) $z_0 = 700~mm$} & \,\,
		\small{(d) $z_0 = 750~mm$} & \,\,
		\small{(e) $z_0 = 900~mm$}
	\end{tabular}
\caption{{3D-reconstructions after 10 iterations of the ADMM scheme, taking as initial guess {different fronto-parallel planes $z \equiv z_0$}. The median of the distances to ground truth is, from left to right: $3.05~mm$, $2.88~mm$, $1.68~mm$, $2.08~mm$ and $5.86~mm$. When the initial guess is too close to the camera, the 3D-reconstruction is flattened, while the scale is overestimated when starting too far away from the camera (although this yields a lower energy, see Fig.~\ref{fig:10}).}}
\label{fig:11}
\end{figure*}

\begin{algorithm}[ratio-based ADMM approach] {\ }
\label{alg:stota_2}
\begin{algorithmic}[1] 
\STATE Initialize $\tilde{z}^{(0)} = \overline{z}^{(0)}$, $h^{(0)} \equiv 0$. Set $k:=0$.

\LOOP

\STATE
Update $\tilde{z}$ by using the linear part, ``while {keeping $\tilde{z}$ close to $\overline{z}^{(k)}$}'':
\vspace*{-0.5em}
	\begin{align}
		\tilde{z}^{(k+1)} & = \underset{\tilde{z}}{\arg\min} \left\| \mathbf{A}(\overline{z}^{(k)}) \nabla \tilde{z} - \mathbf{b}(\overline{z}^{(k)}) \right\|_{L^2(\Omega)}^2 \nonumber \\[-0.75em]
		& \qquad + \frac{1}{2 \, \nu} \left\|\tilde{z} - \overline{z}^{(k)} + h^{(k)}\right\|_{L^2(\Omega)}^2.
	\label{eq:60}
	\end{align}
\vspace*{-1em}

\STATE 
Update $\overline{z}$ by using the nonlinear part, ``while {keeping $\overline{z}$ close to $\tilde{z}^{(k+1)}$}'':
\vspace*{-0.5em}
	\begin{align}
		{\overline{z}}^{(k+1)} & = \underset{\overline{z}}{\arg\min} \left\| \mathbf{A}(\overline{z}) \, \nabla \tilde{z}^{(k+1)} - \mathbf{b}(\overline{z}) \right\|_{L^2(\Omega)}^2 \nonumber \\[-0.75em]
	&	\qquad+ \frac{1}{2 \, \nu} \left\|\tilde{z}^{(k+1)} \!\!-\! \overline{z} \!+\! h^{(k)}\right\|_{L^2(\Omega)}^2.
	\label{eq:61}
	\end{align}
\vspace*{-1em}	
\STATE
Update the dual variable $h$:
\vspace*{-0.75em}
	\begin{equation}
		h^{(k+1)} = h^{(k)} + \tilde{z}^{(k+1)} -\overline{z}^{(k+1)}.
	\end{equation}
\vspace*{-2em}
\STATE
If the stopping criterion is not satisfied, then set $k:=k+1$.

\ENDLOOP

\end{algorithmic}
\end{algorithm}

Stage \eqref{eq:60} of Algorithm~\ref{alg:stota_2} is a linear least-squares problem which can be solved using the normal equations of its discrete formulation\footnote{In our experiments, the gradient operator $\nabla$ is discretized by forward, first-order finite differences with a Neumann boundary condition.}. The presence of the regularization term {now guarantees} the positive definiteness of the matrix of the system. This matrix is however too large to be inverted directly. Therefore, we resort to the conjugate gradient algorithm.

{Thanks to} the auxiliary variable $\overline{z}$, which decouples {$\nabla \tilde{z}$ and $\tilde{z}$ in Problem \eqref{eq:59}}, Stage \eqref{eq:61} of Algorithm~\ref{alg:stota_2} is a \emph{local} nonlinear least-squares problem: in fact, {$\nabla \overline{z}$} is not involved in this problem, which can be \colorme{solved} pixelwise. Problem \eqref{eq:61} thus reduces to a nonlinear least-squares estimation problem of one real variable, which can be solved by a standard method such as the Levenberg-Marquardt algorithm.

Because of the nonlinearity of Problem \eqref{eq:61}, it is unfortunately impossible to guarantee convergence \colorme{for}  {this} ADMM scheme, which depends on the initialization and on parameter $\nu$ \cite{Boyd2011}. A reasonable initialization strategy consists in using the solution provided by Algorithm~\ref{alg:stota_1} (cf.~Section \ref{sec:3.1}). As for the descent {step $\nu$}, we iteratively calculate its optimal value according to the \emph{Penalty Varying Parameter} procedure described in \cite{Boyd2011}. Finally, the iterations stop when the relative variation of the criterion of Problem \eqref{eq:55} falls under a threshold equal to $10^{-4}$. 

Fig.~\ref{fig:9} shows that with such choices, Problem \eqref{eq:55} is solved more efficiently than with the fixed point scheme: the energy is now decreased at each iteration. Fig.~\ref{fig:10} shows that this is the case whatever the initial guess, although initialization has a strong impact on the solution, as confirmed by {Fig.\ \ref{fig:11}}. 

\begin{figure}[!htpb]
\begin{center}
	\includegraphics[width = 0.85\linewidth]{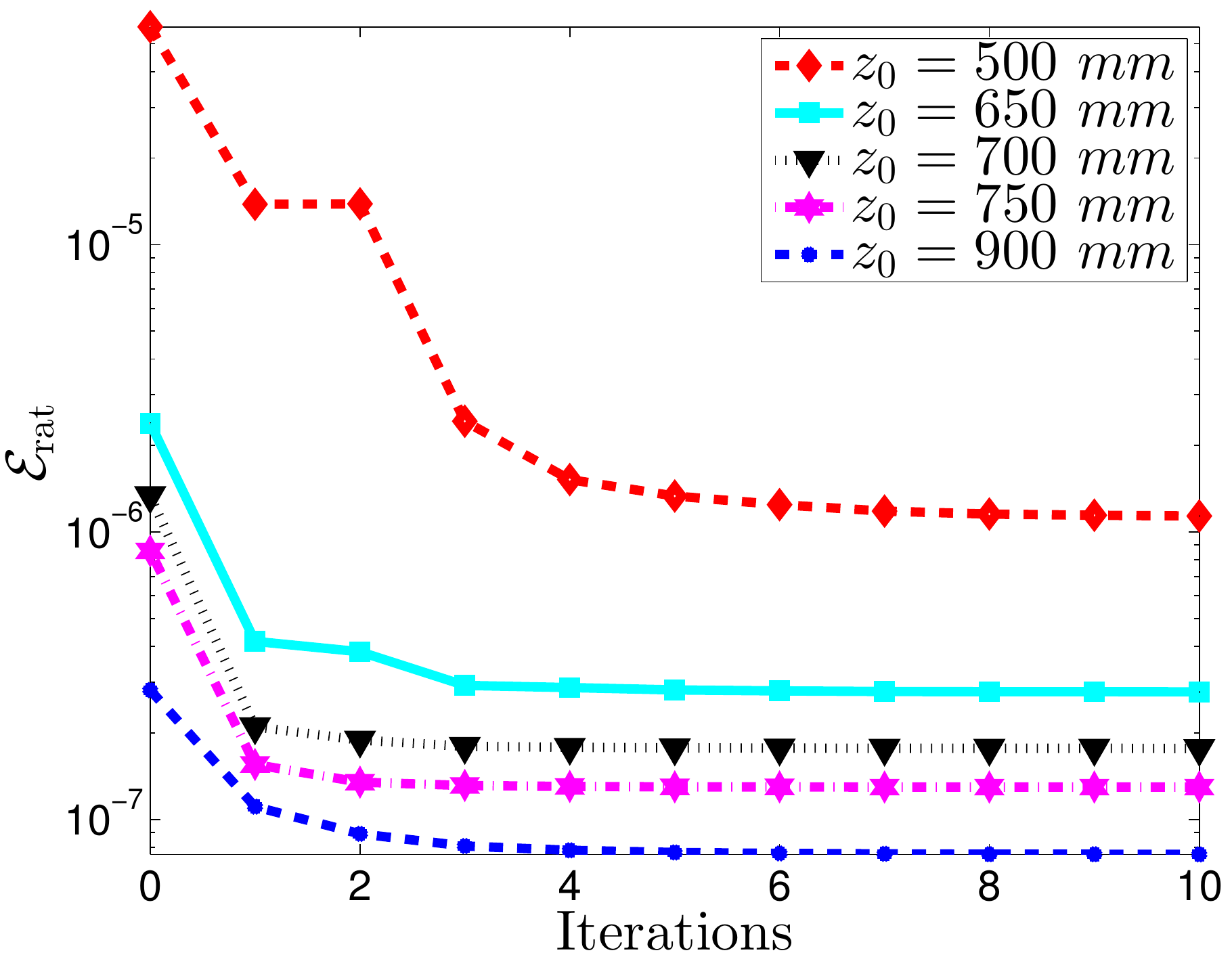} 
\end{center}
\caption{Evolution of the energy $\mathcal{E}_{\mathrm{rat}}$ defined in \eqref{eq:55}, in function of the iterations, for the data of Fig.\ \ref{fig:2}. Using as initialization $\tilde{z}^{(0)} \equiv \log(z_0)$, the ADMM scheme always converges towards a local minimum, yet this minimum strongly depends on the value of $z_0$.
{Besides, a lower final energy does not necessarily means a better 3D-reconstruction, as shown in Fig.~\ref{fig:11}. Hence, not only a careful initial guess is of primary importance, but the criterion derived from image ratios prevents automatic scale estimation.}}
\label{fig:10}
\end{figure}

\begin{figure}[!htpb]
\centering
	\begin{tabular}{c}
		\includegraphics[height = 0.5\linewidth]{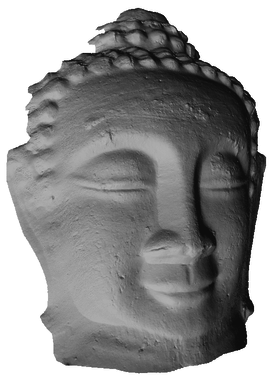} \\
		\small{(a)} \\		
  	\def\svgwidth{0.75\linewidth}
		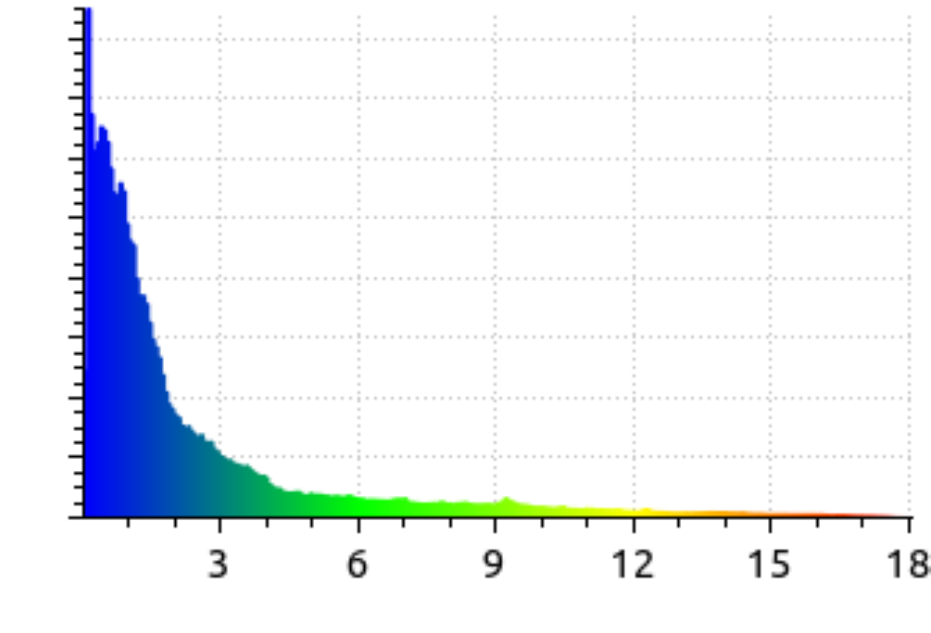 \\
    \small{(b)}
	\end{tabular}
\caption{(a) 3D-reconstruction obtained with Algorithm~\ref{alg:stota_2}, using the result from Fig.\ \ref{fig:7}-a as initial guess. (b) Histogram of point-to-point distances between this 3D-shape and the ground truth (cf. Fig.\ \ref{fig:7}-c). The median value is $1.2~mm$.}
\label{fig:12}
\end{figure}

Fig.~\ref{fig:12} shows the 3D-reconstruction obtained by refining the results of Section~\ref{sec:3.1} using Algorithm~\ref{alg:stota_2}. At first sight, the 3D-shape depicted in Fig.\ \ref{fig:12}-a seems hardly different from that of Fig.\ \ref{fig:7}-a, but the comparison of histograms in Figs.\ \ref{fig:8}-a and \ref{fig:12}-b indicates that bias has been significantly reduced. \colorme{This shows the superiority of direct depth estimation over alternating normal estimation and integration}. 

However, the lack of convergence guarantees and the strong dependency on the initialization remain limiting bottlenecks. The method discussed in the next section overcomes both these issues.


{\colormeall

\section{A New, Provably Convergent Variational Approach for Photometric Stereo under Point Light Source Illumination}
\label{sec:4}

When it comes to solving photometric stereo under point light source illumination, there are two main difficulties: the dependency of the lighting vectors on the depth map (cf. Eq.~\eqref{eq:t_i}), and the presence of the nonlinear coefficient ensuring that the normal vectors have unit-length (cf. Eq.~\eqref{eq:50}). 

The alternating strategy from Section~\ref{sec:3.1} solves the former issue by freezing the lighting vectors at each iteration, and the latter by simultaneously estimating the normal vector and the albedo. The objective function tackled in this approach, which is based on the reprojection error, seems to be the most relevant. Indeed, the final result  seems to be independent from the initialization, although convergence is not established.

On the other hand, the differential strategy from Section~\ref{sec:3.2} explicitly tackles the nonlinear dependency of lighting on the depth, and eliminates the other nonlinearity using image ratios. Directly estimating depth reduces bias, but the objective function derived from image ratios admits a global solution which is not acceptable (depth uniformly tending to $+\infty$), albedo is not estimated and convergence is not established either. 

Therefore, an ideal numerical solution should: ($i$) build upon a differential approach, in order to reduce bias, ($ii$) avoid linearization using ratios, in order to avoid the trivial solution and allow albedo estimation, and ($iii$) be provably convergent. The variational approach presented in this section, initially presented in~\cite{SSVM2017}, satisfies these three criteria. 

\subsection{Proposed Discrete Variational Framework}

The nonlinearity of the PDEs~\eqref{eq:3.16} with respect to $\nabla \tilde{z}$, due to the nonlinear dependency of $d(\tilde{z})$ (see Eq.~\eqref{eq:50}), is challenging. We could explicitly consider this nonlinear coefficient within a variational framework~\cite{Hoeltgen_2016}, but we rather take inspiration from the way conventional photometric stereo~\cite{Woodham1980a} is linearized and integrate the nonlinearity inside the albedo variable, as we proposed recently in~\cite{SSVM2017,CVPR2017}. Instead of estimating $\overline{\rho}(\mathbf{p})$ in each pixel $\mathbf{p}$, we thus rather estimate:
\begin{equation}
  \tilde{\rho}(\mathbf{p}) = \frac{\overline{\rho}(\mathbf{p})}{d(\tilde{z})(\mathbf{p})}.
  \label{eq:63}
\end{equation}


The system of PDEs~\eqref{eq:3.16} is then rewritten as
\begin{align}
& 	I^i(\mathbf{p}) = \tilde{\rho}(\mathbf{p}) \,
	\left[\mathbf{Q}(\mathbf{p}) \, \mathbf{t}^i(\tilde{z})(\mathbf{p}) \right] \cdot
	\begin{bmatrix}
		\nabla \tilde{z}(\mathbf{p}) \\
		- 1
	\end{bmatrix},\nonumber \\
&	\qquad\qquad\qquad\qquad\qquad\qquad\qquad	~ i\in \{1,\dots,m\}, 
	\label{eq:64}
\end{align} 
where we use the following notation, $\forall \mathbf{p} = \left[u,v\right]^\top \in \Omega$:
\begin{align}
  \mathbf{Q}(\mathbf{p}) =&  \begin{bmatrix}
    f & 0 & -u \\
    0 & f & -v \\
    0 & 0 & 1
  \end{bmatrix}.
  \label{eq:49}   
\end{align} 

System~\eqref{eq:64} is a system of \emph{quasilinear} PDEs in $(\tilde{\rho},\tilde{z})$, because $\mathbf{t}^i(\tilde{z})$ only depends on $\tilde{z}$, and not on $\nabla \tilde{z}$. Once $\tilde{\rho}$ and $\tilde{z}$ are estimated, it is straightforward to recover the ``real'' albedo $\overline{\rho}$ using \eqref{eq:63}.

Let us now denote $j \in \{ 1, \dots, n\}$ the indices of the pixels inside $\Omega$, $I^i_j$ the gray level of pixel $j$ in image $I^i$, $\btrho \in \mathbb{R}^n$ and $\tilde{\mathbf{z}} \in \mathbb{R}^n$ the vectors stacking the unknown values $\tilde{\rho}_j$ and $\tilde{z}_j$, $\mathbf{t}^i_j(\tilde{z}_j) \in \mathbb{R}^3$ the vector $\mathbf{t}^i(\tilde{z})$ at pixel $j$, which smoothly (though nonlinearly) depends on $\tilde{z}_j$, and $\mathbf{Q}_j$ the matrix defined in Eq.~\eqref{eq:49} at pixel $j$. Then, the discrete counterpart of System~\eqref{eq:64} is written as the following system of nonlinear equations in $(\btrho,\tilde{\mathbf{z}})$:
\begin{align}
&	I^i_j = \tilde{\rho}_j  \,
	\left[\mathbf{Q}_j  \, \mathbf{t}^i_j(\tilde{z}_j) \right] \cdot
	\begin{bmatrix}
		\left(\nabla \mathbf{\tilde{z}} \right)_j \\
		- 1
	\end{bmatrix}, \nonumber \\ 
&	\qquad\qquad\qquad\qquad  i\in \{1,\dots,m\},\, j \in \{1,\dots,n\},
	\label{eq:65}
\end{align} 
where $\left(\nabla \tilde{\mathbf{z}} \right)_j \in \mathbb{R}^2$ represents a finite differences approximation of the gradient of $\tilde{z}$ at pixel $j$\footnote{\colorme{In our experiments, we use the same discretization as in Section~\ref{sec:3.2}, for fair comparison.}}.

Our goal is to jointly estimate $\btrho\in\mathbb{R}^n$ and $\btz\in\mathbb{R}^n$ from the set of nonlinear equations~\eqref{eq:65}, as solution of  the following discrete optimization problem:
\begin{equation}
\min_{\substack{ \btrho,\btz}}
\mathcal{E}(\btrho,\btz) := \sum_{j=1}^n \sum_{i=1}^m\phi\left(r^i_j(\btrho,\btz)\right),
\label{eq:66}
\end{equation}
where the residual $r^i_j(\btrho,\btz)$ depends \emph{locally} (and linearly) on $\btrho$, but \emph{globally} (and nonlinearly) on $\btz$: 
\begin{equation}
r^i_j(\btrho,\btz) = \tilde{\rho}_j \left\{ \zeta^i_j(\btz) \right\}_+ - I^i_j,
\end{equation}
with
\begin{equation}
\zeta^i_j(\btz) = \left[ \mathbf{Q}_j \mathbf{t}^i_j(\tilde{z}_j) \right] \cdot \begin{bmatrix} (\nabla \btz)_j \\ -1 \end{bmatrix}. 
\end{equation}
An advantage of our formulation is to be generic i.e., independent from the choice of the operator $\{\cdotp\}_+$ and of the function $\phi$. For fair comparison with the algorithms in Section~\ref{sec:3}, one can use $\{x\}_+ = x$ and $\phi(x) = \phi_{\text{LS}}(x) = x^2$. To improve robustness, self-shadows can be explicitly handled by using $\{x\}_+ = \max\{x,0\}$, and the estimator $\phi$ can be chosen as any $\mathbb{R} \to \mathbb{R}^+$ function which is even, twice continuously differentiable, and monotonically increasing over $\mathbb{R}^+$ such that:
\begin{equation} \label{eq:dasEq}
\frac{\phi'(x)}{x}\geq\phi''(x),~ \forall x \in \mathbb{R}. 
\end{equation}
A typical example is Cauchy's  robust M-estimator\footnote{See~\cite{CVPR2017} for some discussion and comparison with state-of-the-art robust methods~\cite{Ikehata2014b,Mecca2016,Wu2010}.}:
\begin{equation}
  \phi_{\text{Cauchy}}(x) = \lambda^2 \log\left(1+\frac{x^2}{\lambda^2}\right),  
\end{equation}
where the parameter $\lambda$ is user-defined  (we use $\lambda = 0.1$).

\subsection{Alternating Reweighted Least-Squares for Solving~\eqref{eq:66}}
\label{sec:4.2}

Our goal is to find a local minimizer $(\btrho^*,\btz^*)$ for~\eqref{eq:66}, which must satisfy the following first-order conditions\footnote{We use the notation $\frac{\partial}{\partial}$ to avoid the confusion with the spatial derivatives denoted by $\nabla$, and neglect the fraction when the derivation variable is obvious.}:
\iali{
\frac{\partial \mathcal{E}}{\partial\btrho}(\btrho^*\!,\!\btz^*)
&\!=\!\!\sum_{j=1}^n\! \sum_{i=1}^m\!  \phi'(r^i_j(\btrho^*\!,\btz^*)) \frac{\partial r^i_j}{\partial \btrho }(\btrho^*\!,\btz^*\!)
\!=\! {\bm 0}, \label{eq:69}\\[-0.5em]
\frac{\partial \mathcal{E}}{\partial\btz}(\btrho^*\!,\!\btz^*)
&\!=\!\!\sum_{j=1}^n \!\sum_{i=1}^m\! \phi'(r^i_j(\btrho^*\!,\btz^*))  \frac{\partial r^i_j}{\partial \btz }(\btrho^*\!,\btz^*\!)
\!=\! {\bm 0}, \label{eq:70}
}
with:
\begin{align}
\frac{\partial r^i_j}{\partial \tilde{\rho}_l }(\btrho^*,\btz^*) &= \begin{cases} 
\{\zeta^i_j(\btz^*)\}_+ & \text{~if~}  l = j, \\
0  & \text{~if~}  l \neq j, \\
\end{cases} \\
\frac{\partial r^i_j}{\partial \btz }(\btrho^*,\btz^*) &=\tilde{\rho}^*_j \, \chi(\zeta^i_j(\btz^*))  \,  \partial\zeta^i_j(\btz^*). \label{eq:72} 
\end{align}
In~\eqref{eq:72}, $\chi$ is the (sub-)derivative of $\{\cdotp\}_+$, which is a constant function equal to $1$ if $\{x\}_+ = x$, and the Heaviside function if $\{x\}_+ = \max\{x,0\}$.

For this purpose, we derive an alternating reweighted least-squares (ARLS) scheme. Suggested by its name, the ARLS scheme alternates Newton-like steps over~$\btrho$ and~$\btz$, which can be interpreted as iteratively reweighted least-squares iterations. Similar to the famous iteratively reweighted least-squares~\cite{Wolke1988} (IRLS) algorithm, ARLS solves the original (possibly non-convex) problem~\eqref{eq:66} iteratively, by recasting it as a series of simpler quadratic programs. 

Given the current estimate $(\btrho^{(k)},\btz^{(k)})$ of the solution, ARLS first freezes $\btz$ and updates $\btrho$ by minimizing the following local quadratic approximation of $\mathcal{E}(\cdot,\btz^{(k)})$ around $\btrho^{(k)}$\footnote{The right hand side function in Eq.~\eqref{eq:local_approx} is a majorant of $\mathcal{E}(\cdot,\btz^{(k)})$, and it is easily verified that its value and gradient are equal to those of $\mathcal{E}(\cdot,\btz^{(k)})$ in $\btrho^{(k)}$. It is therefore suitable as approximation. }:
\begin{align}
 & \mathcal{E}(\cdot,\btz^{(k)}) \approx \sum_{j=1}^n\sum_{i=1}^m \Bigg\{ \phi\left(r^i_j(\btrho^{(k)},\btz^{(k)})\right) \nonumber \\
 & \quad  + \frac{\phi'(r^i_j(\btrho^{(k)},\btz^{(k)}))}{r^i_j(\btrho^{(k)},\btz^{(k)})} \, \frac{r^i_j(\cdot,\btz^{(k)})^2 - r^i_j(\btrho^{(k)},\btz^{(k)})^2 }{2}  \Bigg\},
 \label{eq:local_approx}
\end{align}
where we set $\frac{\phi'(r^i_j(\btrho^{(k)},\btz^{(k)}))}{r^i_j(\btrho^{(k)},\btz^{(k)})} = 0$ if $r^i_j(\btrho^{(k)},\btz^{(k)}) = 0$.


Then, $\btrho$ is freezed and $\btz$ is updated by minimizing a local quadratic approximation of $\mathcal{E}(\btrho^{(k+1)},\cdot)$ around $\btz^{(k)}$, which is in all points similar to~\eqref{eq:local_approx}. Iterating this procedure yields the following alternating sequence of reweighted least-squares problems: 
\begin{align}
\btrho^{(k+1)} &= \underset{\btrho\in\bR^n}{\arg\min}~
\mathcal{E}_{\btrho}(\btrho;\btrho^{(k)},\btz^{(k)}) := \nonumber \\
& \qquad \frac12 \sum_{j=1}^n\sum_{i=1}^m w^i_j(\btrho^{(k)},\btz^{(k)}) \, r^i_j(\btrho,\btz^{(k)})^2, \label{eq:75} \\
\btz^{(k+1)} &= \underset{\btz\in\bR^n}{\arg \min}~
\mathcal{E}_{\btz}(\btz;\btrho^{(k+1)},\btz^{(k)}) := \nonumber \\
&~ \frac12 \sum_{j=1}^n\sum_{i=1}^m  w^i_j(\btrho^{(k+1)},\btz^{(k)}) \, r^i_j(\btrho^{(k+1)},\btz)^2. \label{eq:76}
\end{align}
Here, the functions $\mathcal{E}_{\btrho}$ and $\mathcal{E}_{\btz}$ are the above local quadratic approximations minus the constants which play no role in the optimization, and the following (lagged) weight variable $w$ is used\footnote{\colorme{Since $\phi$ is supposed even and monotonically increasing over $\mathbb{R}^+$, this variable can be used as weight because, $\forall x \in \mathbb{R} \backslash \{0\}$, $\phi'(x) / x \geq 0$ and thus $w^i_j(\btrho,\btz) \geq 0$.}}: 
\begin{equation}
w^i_j(\btrho,\btz) = \begin{cases}
  \dfrac{\phi'(r^i_j(\btrho,\btz))}{r^i_j(\btrho,\btz)} &\text{~if~}r^i_j(\btrho,\btz) \neq 0,  \\
  0& \text{~otherwise}.
\end{cases}
\label{eq:74}
\end{equation}



\paragraph{Solution of the $\btrho$-subproblem. } 

Problem~\eqref{eq:75} can be rewritten as the following $n$ independent linear least-squares problems, $j \in \{1,\dots,n\}$:
\begin{equation}
\tilde{\rho}_j^{(k+1)} \!=\!  \underset{\tilde{\rho}_j\in\bR}{\arg\min}~ \frac12
\sum_{i=1}^m \! w^i_j(\btrho^{(k)}\!,\btz^{(k)})\, r^i_j(\btrho,\btz^{(k)})^2.
\label{eq:N1}
\end{equation}
Each problem~\eqref{eq:N1} almost always admits a unique solution. When it does not, we set $\tilde{\rho}_j^{(k+1)} = \tilde{\rho}_j^{(k)}$. The update thus admits the following closed-form solution:
\begin{equation}
\tilde{\rho}_j^{(k+1)} =
\begin{cases}
 \dfrac{\sum_{i=1}^m  w^i_j(\btrho^{(k)},\btz^{(k)}) \left\{ \zeta^i_j(\btz^{(k)}) \right\}_+ I^i_j  }{\sum_{i=1}^m w^i_j(\btrho^{(k)},\btz^{(k)}) \left\{ \zeta^i_j(\btz^{(k)}) \right\}_+^2 }  \\[1em]
 \qquad \text{if~}  \sum_{i=1}^m w^i_j(\btrho^{(k)},\btz^{(k)}) \left\{ \zeta^i_j(\btz^{(k)}) \right\}_+^2  > 0,\\[0.5em]
 \tilde{\rho}_j^{(k)} ~ \text{if~}  \sum_{i=1}^m w^i_j(\btrho^{(k)},\btz^{(k)}) \left\{ \zeta^i_j(\btz^{(k)}) \right\}_+^2  = 0. 
\end{cases}
\label{eq:rho_closed}
\end{equation}
The second case in~\eqref{eq:rho_closed} means that $\btrho^{(k+1)}$ is set to be the solution of~\eqref{eq:75} which has minimal (Euclidean) distance to $\btrho^{(k)}$.

The update~\eqref{eq:rho_closed} can also be obtained by remarking that, since~\eqref{eq:75} is a linear least-squares problem, the solution of the equation $\partial \mathcal{E}_{\btrho}(\btrho;\btrho^{(k)},\btz^{(k)}) = {\bm 0}$ is attained in one step of the Newton method: 
\ieqn{ 
\label{eq:NBV}
\btrho^{(k+1)}\!=\!\btrho^{(k)}\!-\!H_{\btrho}(\btrho^{(k)}\!,\btz^{(k)})^\dagger \, \partial \mathcal{E}_{\btrho}(\btrho^{(k)}\!;\btrho^{(k)}\!,\btz^{(k)}).
}
In~\eqref{eq:NBV}, the $n$-by-$n$ matrix $H_{\btrho}(\btrho^{(k)},\btz^{(k)})$ is the Hessian of $\mathcal{E}_{\btrho}(\cdot;\btrho^{(k)},\btz^{(k)})$ at $\btrho^{(k)}$\footnote{Lemma~\ref{lem:1} shows that it is a positive semi-definite approximation of the Hessian $\frac{\partial^2 \mathcal{E}}{\partial \btrho^2}(\btrho^{(k)},\btz^{(k)})$, hence the notation.}, i.e.: 
\begin{align}
&\delta\btrho^\top \!H_{\btrho}(\btrho^{(k)},\btz^{(k)})  \, \delta\btrho
=\sum_{j=1}^n\sum_{i=1}^m w^i_j(\btrho^{(k)},\btz^{(k)}) \nonumber \\[-0.5em]
& \qquad\qquad\qquad\qquad\qquad\qquad\left(\delta\tilde{\rho}_j\{\zeta^i_j(\btz^{(k)})\}_+\right)^2  
  \label{eq:78}
\end{align} 
for any $\delta\btrho = \left[ \delta\tilde{\rho}_1,\dots,\delta\tilde{\rho}_n \right]^\top \in\bR^n$. Since the $n$ problems~\eqref{eq:N1} are independent, it is a diagonal matrix with entry $(j,j)$ equal to $e_j = \displaystyle\sum_{i=1}^m w^i_j(\btrho^{(k)},\btz^{(k)}) \left\{ \zeta^i_j(\btz^{(k)}) \right\}_+^2$. This matrix is singular if one of the entries $e_j$ is equal to zero, but its pseudo-inverse always exists: it is an $n$-by-$n$ diagonal matrix whose entry $(j,j)$ is equal to $1/ e_j$ as soon as $e_j >0$, and to $0$ otherwise. The updates~\eqref{eq:rho_closed} and~\eqref{eq:NBV} are thus strictly equivalent.

\paragraph{Solution of the $\btz$-subproblem. }

The depth update~\eqref{eq:76} is a nonlinear least-squares problem, due to the nonlinearity of $r^i_j(\btrho,\btz)$ with respect to $\btz$. 
We therefore introduce an additional linearization step i.e., we follow a Gauss-Newton strategy. A first-order Taylor approximation of $r^i_j(\btrho^{(k+1)},\cdot)$ around $\btz^{(k)}$ yields, using~\eqref{eq:72}:
\begin{align}
&\mathcal{E}_{\btz}(\btz;\btrho^{(k+1)},\btz^{(k)}) \approx \overline{\mathcal{E}}_{\btz}(\btz;\btrho^{(k+1)},\btz^{(k)}):= \notag\\[-0.5em]
& ~ \frac12 \sum_{j=1}^n\sum_{i=1}^m  w^i_j(\btrho^{(k+1)},\btz^{(k)}) \Big(r^i_j(\btrho^{(k+1)},\btz^{(k)}) \notag\\[-0.5em]
& \quad +\tilde{\rho}^{(k+1)}_j
\chi(\zeta^i_j(\btz^{(k)})) \,
(\btz-\btz^{(k)})^\top  \partial\zeta^i_j(\btz^{(k)})\Big)^2. 
\label{eq:achtung}
\end{align}
Therefore, we replace the update~\eqref{eq:76} by
\begin{equation}
\btz^{(k+1)} = \underset{\btz\in\bR^n}{\arg \min}~\overline{\mathcal{E}}_{\btz}(\btz;\btrho^{(k+1)},\btz^{(k)}), 
\label{eq:ohmy}
\end{equation}
which is a linear least-squares problem whose solution is attained in one step of the Newton method\footnote{Similar to the $\btrho$-subproblem, $\btz^{(k+1)}$ is taken to be of minimal distance to $\btz^{(k)}$ whenever non-uniqueness of the solution in \eqref{eq:ohmy} is encountered. The pseudo-inverse operator in~\eqref{eq:wasisdas} takes care of such cases~\cite[Theorem 5.5.1]{Golub2012}.}: 
\ieqn{ 
\btz^{(k+1)}\!=\!\btz^{(k)}\!-H_{\btz}(\!\btrho^{(k+1)},\!\btz^{(k)}\!)^\dagger \, \partial \overline{\mathcal{E}}_{\btz}(\!\btz^{(k)};\btrho^{(k+1)},\!\btz^{(k)}\!),
\label{eq:wasisdas}
}
where the $n$-by-$n$ matrix $H_{\btz}(\btrho^{(k+1)},\btz^{(k)})$ is the Hessian of $\overline{\mathcal{E}}_{\btz}(\cdot;\btrho^{(k+1)},\btz^{(k)})$ at $\btz^{(k)}$, i.e.:
\iali{ 
& \delta\btz^\top H_{\btz}(\btrho^{(k+1)},\btz^{(k)}) \delta\btz=
\sum_{j=1}^n\sum_{i=1}^m \, w^i_j(\btrho^{(k+1)},\btz^{(k)}) \notag\\[-0.5em]
& \qquad\qquad\quad \Big(\tilde{\rho}^{(k+1)}_j 
\chi(\zeta^i_j(\btz^{(k)}))\delta\btz^\top \partial\zeta^i_j(\btz^{(k)})\Big)^2
 \label{eq:82}
}
for any $\delta\btz\in\bR^n$. 

In practice, $H_{\btz}(\btrho^{(k+1)}\!,\btz^{(k)}\!)^\dagger\,\partial \overline{\mathcal{E}}_{\btz}(\!\btz^{(k)};\btrho^{(k+1)}\!,\btz^{(k)}\!)$ in Eq. \eqref{eq:wasisdas} is computed (inexactly) by preconditioned conjugate gradient iterations up to a relative tolerance of $10^{-4}$ (less than fifty iterations in our experiments).

\paragraph{Implementation details.}
The proposed ARLS algorithm is summarized in Algorithm~\ref{alg:1}. 

\begin{algorithm}[alternating reweighted least-squares] {\ }
\label{alg:1}
\begin{algorithmic}[1] 
\STATE Initialize $\btrho^{(0)},\btz^{(0)}\in\bR^n$. Set $k:=0$.
\LOOP

\STATE
Compute $\btrho^{(k+1)}$ by using~\eqref{eq:rho_closed}.
%
%
\STATE 
Compute $\btz^{(k+1)}$ by using~\eqref{eq:wasisdas}.

\STATE
If the stopping criterion is not satisfied, then set $k:=k+1$.
\ENDLOOP
\end{algorithmic}
\end{algorithm}

In our experiments, we use constant vectors as initializations for~$\btz$ and~$\btrho$ i.e., the surface is initially approximated by a plane with uniform albedo. Iterations are stopped when the relative difference between two successive values of the energy $\mathcal{E}$ defined in~\eqref{eq:66} falls below a threshold set to $10^{-3}$. 
In our setup using $m=8$ HD images and a recent i7 processor at $3.50~GHz$ with $32$ $GB$ of RAM, each depth update (the albedo one has negligible cost) required a few seconds, and $10$ to $50$ updates were enough to reach convergence. 

%
%

\subsection{Convergence Analysis}
\label{sec:4.3}

In this subsection, we present a local convergence theory for the proposed ARLS scheme. The proofs are provided in appendix.

When we write $A\succeq B$ (resp.~$A\succ B$), this means that the difference matrix $A-B$ is positive semidefinite (resp.~positive definite). The spectral radius of a matrix is denoted by $\sr(\cdot)$. 

\paragraph{ARLS as Newton iterations.} It is easily deduced from Eqs.~\eqref{eq:69},~\eqref{eq:75} and~\eqref{eq:74} that $\partial \mathcal{E}_{\btrho}(\btrho^{(k)};\btrho^{(k)},\btz^{(k)}) = \frac{\partial \mathcal{E}}{\partial\btrho}(\btrho^{(k)},\btz^{(k)})$, and thus~\eqref{eq:NBV} also writes
\begin{equation}
  \btrho^{(k+1)} \!=\!\btrho^{(k)} \!-\! H_{\btrho}(\btrho^{(k)}\!,\btz^{(k)})^\dagger \, \frac{\partial \mathcal{E}}{\partial\btrho}(\btrho^{(k)}\!,\btz^{(k)}),
  \label{eq:Newton1}
\end{equation}
which is a quasi-Newton step with respect to the $\btrho$-subproblem in~\eqref{eq:66}, provided that $H_{\btrho}(\btrho^{(k)},\btz^{(k)})$ is a ``reasonable'' approximation of $\frac{\partial^2 \mathcal{E}}{\partial \btrho^2}(\btrho^{(k)},\btz^{(k)})$. Lemma~\ref{lem:1} will clarify what ``reasonable'' means here.

Regarding the $\btz$-update, let us remark that the Gauss-Newton step~\eqref{eq:ohmy} for~\eqref{eq:76} can also be viewed as an approximate solution of the $\btz$-subproblem in~\eqref{eq:66}, linearized around~$\btz^{(k)}$ as follows:
\iali{
&\min_{\btz\in\bR^n} \tilde{\mathcal{E}}_{\btz}(\btz;\btrho^{(k+1)},\btz^{(k)}):= \sum_{j=1}^n\sum_{i=1}^m 
  \,\phi\Big(r^i_j(\btrho^{(k+1)},\btz^{(k)}) \notag\\[-0.5em]
&\quad +\tilde{\rho}^{(k+1)}_j
 \chi(\zeta^i_j(\btz^{(k)})) \,
 (\btz-\btz^{(k)})^\top \partial\zeta^i_j(\btz^{(k)})\Big).
 \label{eq:raus}
}
Since $\partial \overline{\mathcal{E}}_{\btz}(\btz^{(k)};\btrho^{(k+1)},\btz^{(k)}) = \partial \tilde{\mathcal{E}}_{\btz}(\btz^{(k)};\btrho^{(k+1)},\btz^{(k)})$ (see Eqs.~\eqref{eq:74},~\eqref{eq:achtung} and~\eqref{eq:raus}),~\eqref{eq:wasisdas} also writes 
\ieqn{ 
\label{eq:NN}
\btz^{(k+1)}\!=\!\btz^{(k)}\!-H_{\btz}(\btrho^{(k+1)},\btz^{(k)})^\dagger \, \partial\tilde{\mathcal{E}}_{\btz}(\!\btz^{(k)};\btrho^{(k+1)},\!\btz^{(k)}\!),
}
which is a quasi-Newton step for~\eqref{eq:raus}\footnote{And thus a quasi-Newton step with respect to the $\btz$-subproblem in~\eqref{eq:66}, since $\partial\tilde{\mathcal{E}}_{\btz}(\btz^{(k)};\btrho^{(k+1)},\btz^{(k)}) = \frac{\partial \mathcal{E}}{\partial \btz}(\btrho^{(k+1)},\btz^{(k)})$.}, provided that matrix $H_{\btz}(\btrho^{(k+1)},\btz^{(k)})$ is a ``reasonable'' approximation of the Hessian $\partial^2\tilde{\mathcal{E}}_{\btz}(\cdot,\btrho^{(k+1)},\btz^{(k)})$ at $\btz^{(k)}$. Let us now explain our meaning of ``reasonable''. 

\paragraph{A majorization result.} The following lemma establishes the (local) majorization properties of $H_{\btrho}$ and $H_{\btz}$ over the Hessian matrices $\frac{\partial^2 \mathcal{E}}{\partial \btrho^2}$ and $\partial^2 \tilde{\mathcal{E}}_{\btz}$, respectively. 

\begin{lemma} \label{lem:1}
If the following condition holds at $(\btrho^*,\btz^*)$:
\begin{equation}
\label{eq:83}
\zeta^i_j(\btz^*)\neq 0, \quad \forall (i,j)\in \{1,\dots,m\}\times \{1,\dots,n\},
\end{equation}
then we have 
\begin{equation}
  \begin{cases}
H_{\btrho}(\btrho,\btz)&\!\!\!\succeq \frac{\partial^2 \mathcal{E}}{\partial \btrho^2}(\btrho,\btz),  \\
H_{\btz}(\btrho,\btz) &\!\!\!\succeq \partial^2 \tilde{\mathcal{E}}_{\btz}(\btz;\btrho,\btz),
\end{cases}
\end{equation}
whenever $(\btrho,\btz)$ lies in some small neighborhood of $(\btrho^*,\btz^*)$.
\end{lemma}

\paragraph{Convergence proof for ARLS.} The next theorem contains the main result of our local convergence analysis. 

\begin{theorem} \label{thm:1}
Assume that, for some iteration $k$, the iterate $(\btrho^{(k)},\btz^{(k)})$ generated by Algorithm \ref{alg:1} is sufficiently close to some local minimizer $(\btrho^*,\btz^*)$ where, in addition to~\eqref{eq:83}, the following conditions hold:
\begin{align}
& \frac{\partial \mathcal{E}}{\partial\btrho}(\btrho^*,\btz^*)={\bm 0}, \qquad
\frac{\partial \mathcal{E}}{\partial\btz}(\btrho^*,\btz^*)={\bm 0}, \label{eq:87} \\
& \imtx{
\dfrac{\partial^2 \mathcal{E}}{\partial\btrho^2}(\btrho^*,\btz^*) & \dfrac{\partial^2 \mathcal{E}}{\partial\btrho\partial\btz}(\btrho^*,\btz^*) \\
\dfrac{\partial^2 \mathcal{E}}{\partial\btrho\partial\btz}(\btrho^*,\btz^*) & \dfrac{\partial^2 \mathcal{E}}{\partial\btz^2}(\btrho^*,\btz^*)
}
\succ \mathbf{O},  \label{eq:88} \\
& \partial^2 \tilde{\mathcal{E}}_{\btz}(\btz^*;\btrho^*,\btz^*) \succ \mathbf{O}, \label{eq:89} \\
& \sr\!\left(\!\! \partial^2 \tilde{\mathcal{E}}_{\btz}(\btz^*;\btrho^*\!\!,\!\btz^*)^{-1}\!
\left(\!\dfrac{\partial^2 \mathcal{E}}{\partial\btz^2}(\btrho^*\!,\!\btz^*\!)\!-\!\partial^2 \tilde{\mathcal{E}}_{\btz}(\btz^*;\btrho^*\!\!,\!\btz^*\!)\!\right)\!\right)\!<\!1. \label{eq:90}
\end{align}
Then we have $\lim_{k\to\infty}(\btrho^{(k)},\btz^{(k)})=(\btrho^*,\btz^*)$.
\end{theorem}

As a remark, conditions \eqref{eq:87} and \eqref{eq:88} assumed in Theorem \ref{thm:1} are typically referred to as the first-order and the second-order sufficient optimality conditions, while conditions \eqref{eq:89} and~\eqref{eq:90} are similar to the local convergence criteria for Gauss-Newton method, see e.g. \cite[Theorem 1]{Gratton2007}. They always seem satisfied in our experiments i.e., the convergence of ARLS in form of Algorithm~\ref{alg:1} is always observed. If needed, these conditions may however be explicitly enforced by replacing $\{\cdotp\}_+$ by its (smooth) proximity operator, and incorporating a line search step into ARLS, see~\cite{SSVM2017}. 



\subsection{Experimental Validation}

For fair comparison with the methods discussed in Section~\ref{sec:3}, we first consider least-squares estimation without explicit self-shadows handling i.e., $\phi(x) = x^2$ and $\{x\}_+ = x$. 
The results in Figs.\ \ref{fig:13} and~\ref{fig:14} show that, unlike the previous least-squares differential method from Section~\ref{sec:3.2}, the new scheme always converges towards a similar solution for a wide range of initial estimates.

\begin{figure*}[!htpb]
\begin{tabular}{ccc}
	\includegraphics[width = 0.37\linewidth]{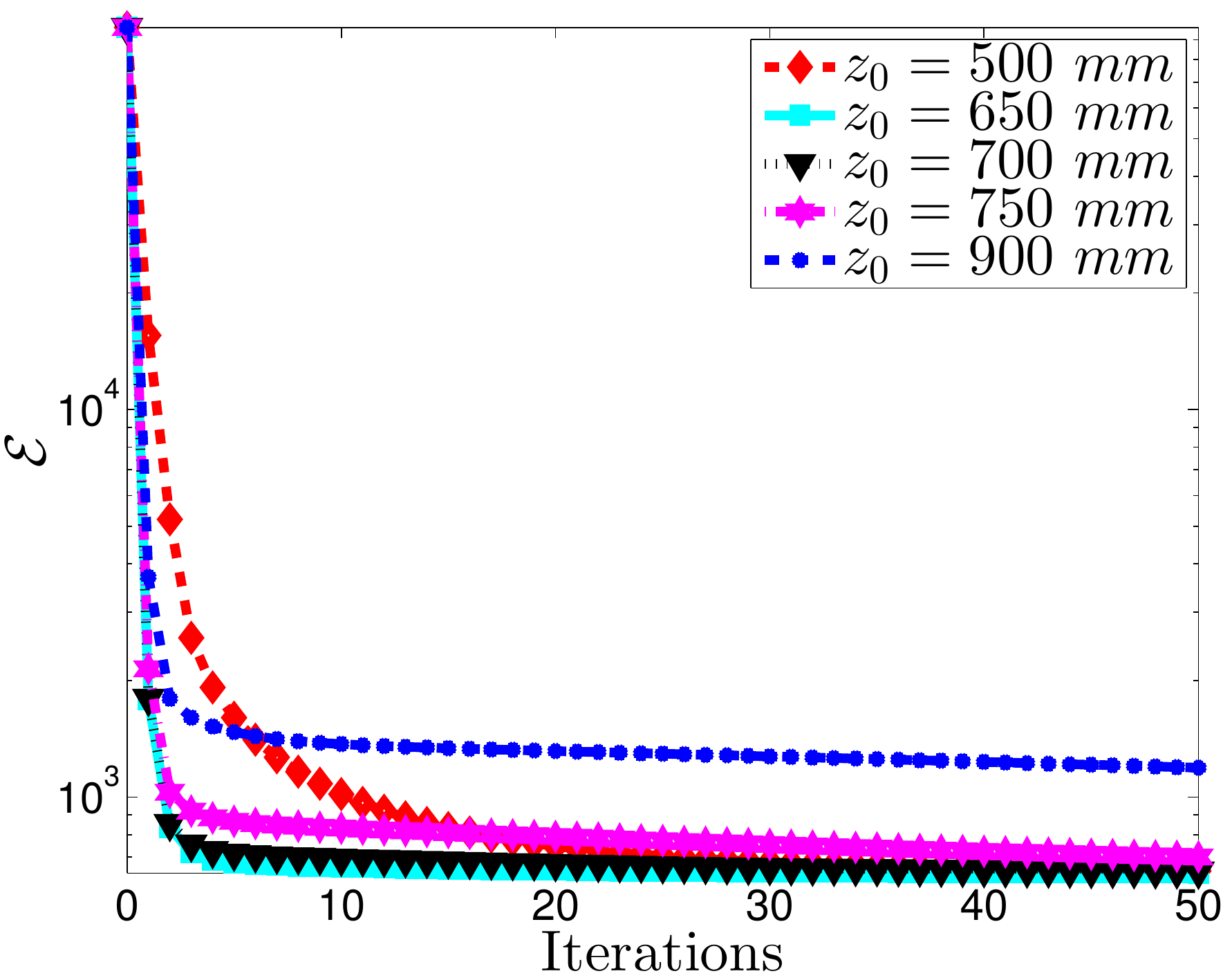} &
	\includegraphics[height = 0.3\linewidth]{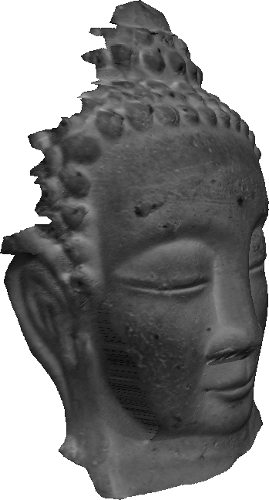} &
	\def\svgwidth{0.415\linewidth}
	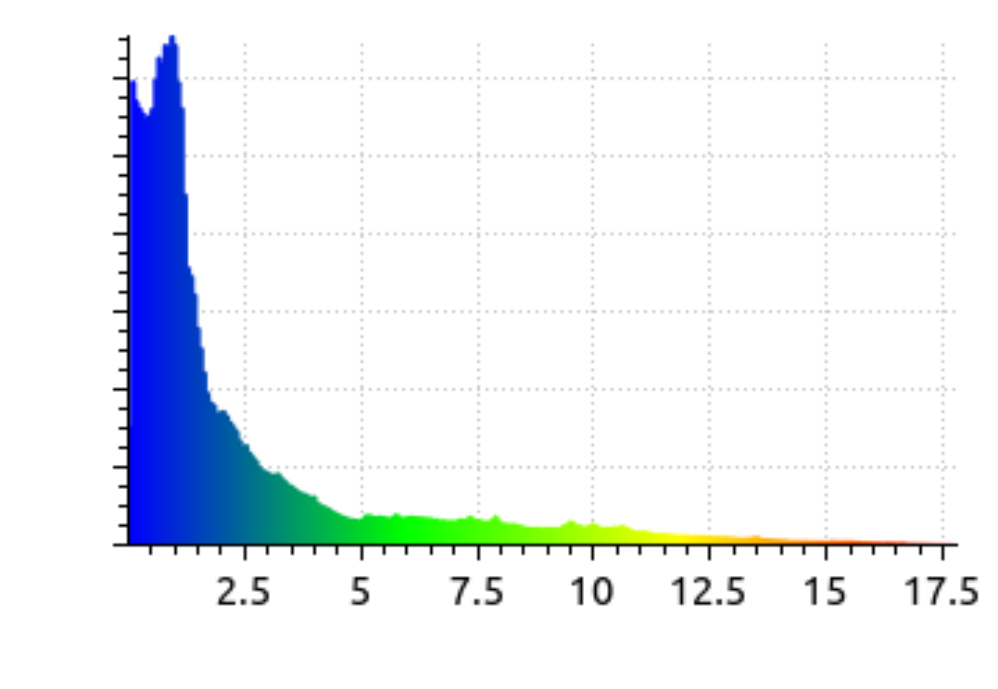   \\
	{\small (a)} & {\small (b)} & {\small (c)}
\end{tabular}
\caption{\colorme{(a) Evolution of the energy $\mathcal{E}$ of the proposed approach, defined in \eqref{eq:66}, using least-squares estimation, in function of the iterations, for the data of Fig.\ \ref{fig:2}. As long as the initial scale is not over-estimated too much, the proposed scheme converges towards similar solutions for different initial estimates (cf. Fig.~\ref{fig:14}), though with different speeds. (b) 3D-model obtained at convergence, using $z_0 = 750~mm$. (c) Histogram of point-to-point distances between (b) and the ground truth (cf. Fig.~\ref{fig:7}-c). As in the experiment of Fig.~\ref{fig:12}, the median value is $1.2~mm$, yet this result is almost independent from the initialization, and is obtained using a provably convergent algorithm. }}
\label{fig:13}
\end{figure*}

\begin{figure*}[!ht]
\centering
	\begin{tabular}{ccccc}
		\includegraphics[height = 0.3\linewidth]{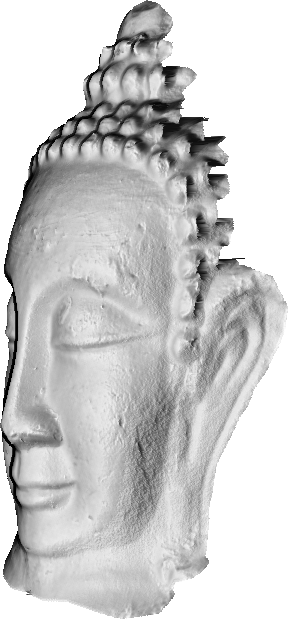} & \,\,
		\includegraphics[height = 0.3\linewidth]{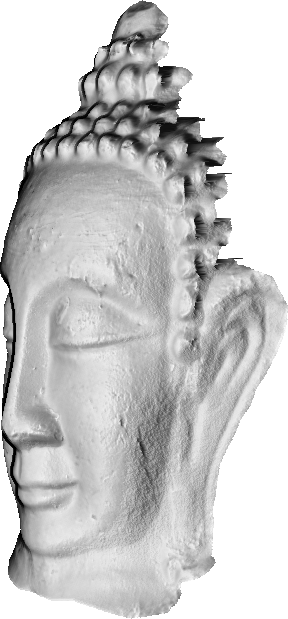} & \,\,
		\includegraphics[height = 0.3\linewidth]{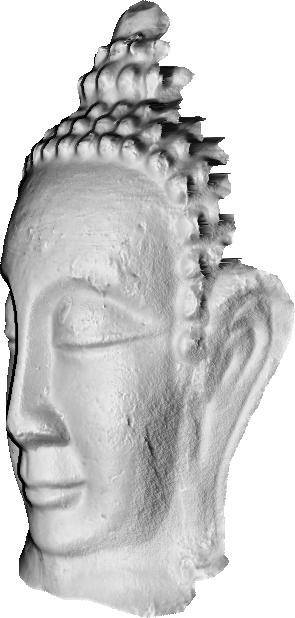} & \,\,
		\includegraphics[height = 0.3\linewidth]{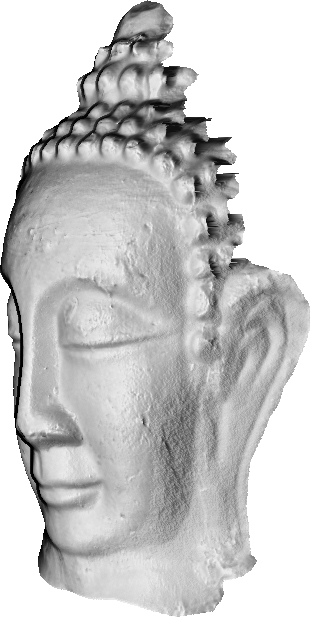} & \,\,
		\includegraphics[height = 0.3\linewidth]{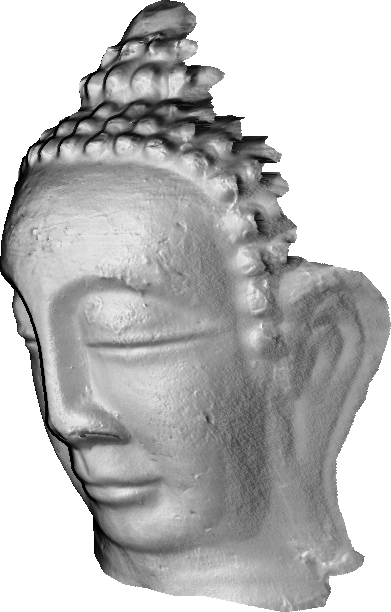} \\
		\small{(a) $z_0 = 500~mm$} & \,\,
		\small{(b) $z_0 = 650~mm$} & \,\,
		\small{(c) $z_0 = 700~mm$} & \,\,
		\small{(d) $z_0 = 750~mm$} & \,\,
		\small{(e) $z_0 = 900~mm$}
	\end{tabular}
\caption{{\colorme{3D-reconstructions after 50 iterations of the proposed scheme, taking as initial guess {different fronto-parallel planes $z \equiv z_0$} and using least-squares estimation. Similar results are obtained whatever the initialization, at least as long as the initial scale is not over-estimated too much. }}}
\label{fig:14}
\end{figure*}

Although the accuracy of the results obtained with this new scheme is not improved, 
 the influence of the initialization is much reduced and convergence is guaranteed. Besides, it is straightforward to improve robustness by simply changing the definitions of the function $\phi$ and of the operator $\{ \cdotp \}_+$, while ensuring robustness of the ratio-based approach is not an easy task~\cite{Mecca2016,Smith2016}. Fig.~\ref{fig:15} shows the result obtained using Cauchy's M-estimator $\Phi_{\text{Cauchy}}$ and explicit self-shadows handling i.e., $\{x\}_+ = \max\{x,0\}$.

\begin{figure*}[!ht]
\centering
\begin{tabular}{ccc}
	\includegraphics[width = 0.37\linewidth]{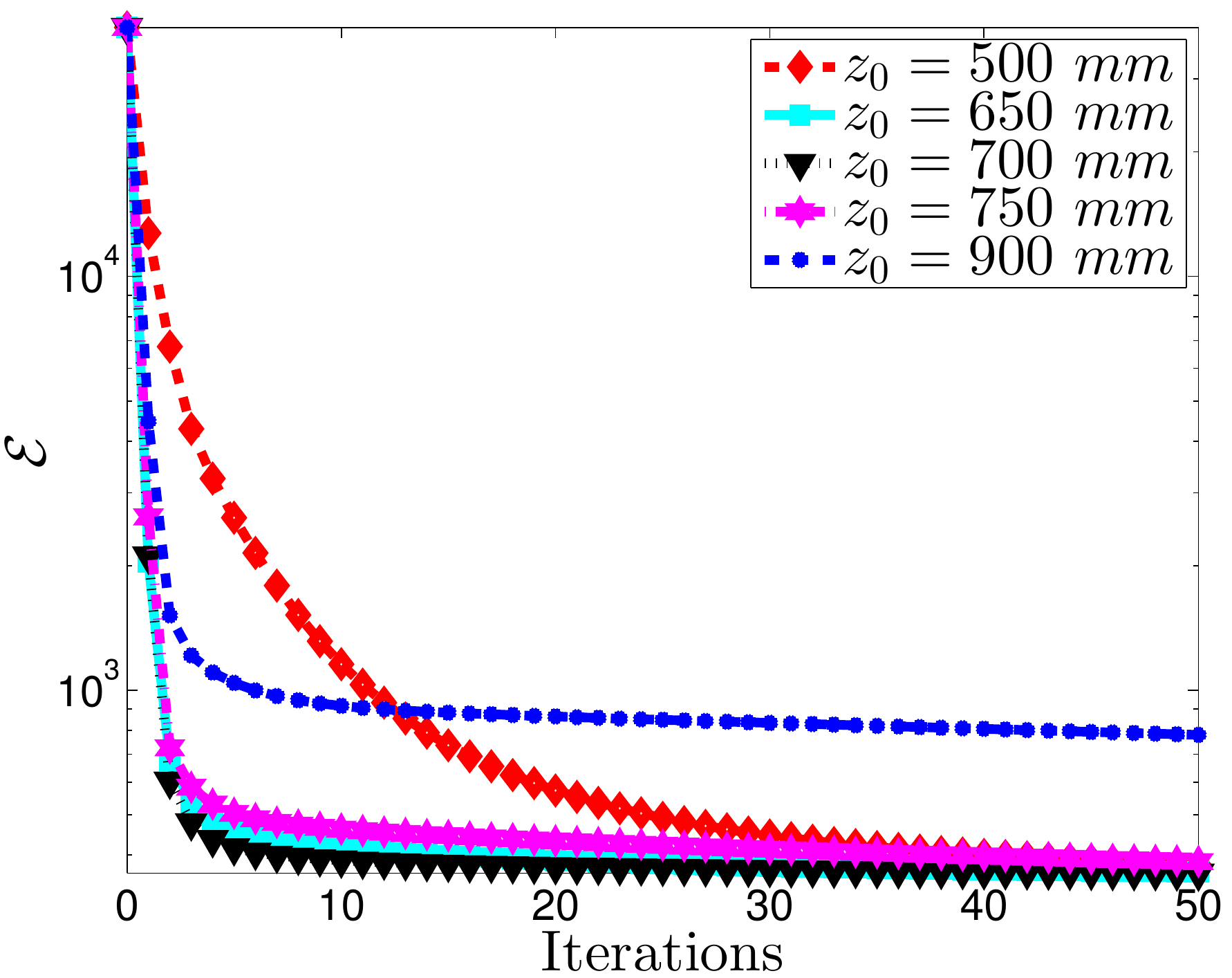} &
	\includegraphics[height = 0.31\linewidth]{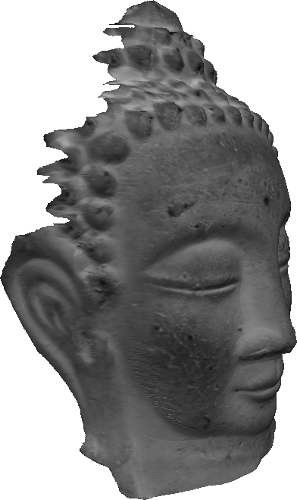} & 
	\def\svgwidth{0.4\linewidth}
	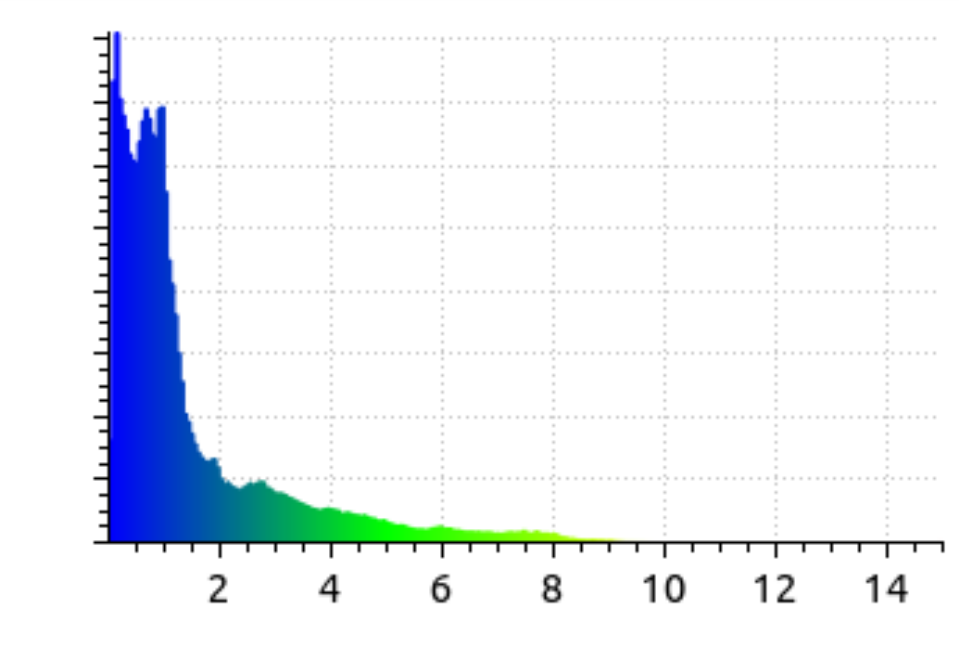   \\
	{\small (a)} & {\small (b)} & {\small (c)}
\end{tabular}
\caption{{\colorme{Same as Fig. \ref{fig:13}, but 
using Cauchy's robust M-estimator and explicit self-shadows handling. Despite the non-convexity of the estimator, convergence is similar to that obtained in the previous experiment. However, the median value of the 3D-reconstruction error is now $0.91~mm$, which is to be compared with the previous value $1.2~mm$ (cf. Fig.\ \ref{fig:13}).}}}
\label{fig:15}
\end{figure*}

}


\section{Estimating Colored 3D-models by Photometric Stereo}
\label{sec:5}

So far, we have considered only gray level images. In this section, we extend our study to RGB-valued images, in order to estimate colored 3D-models using photometric stereo. Similar to Section~\ref{sec:2}, we will first establish the image formation model and discuss calibration. Then, we will show how to modify the algorithm from Section~\ref{sec:4} in order to handle RGB images.


\subsection{Spectral Dependency of the Luminous Flux Emitted by a LED}

\begin{figure}[!htpb]
\begin{center}
	\begin{tabular}{cc}
		\!\!\!\!\!\includegraphics[width = 0.59\linewidth]{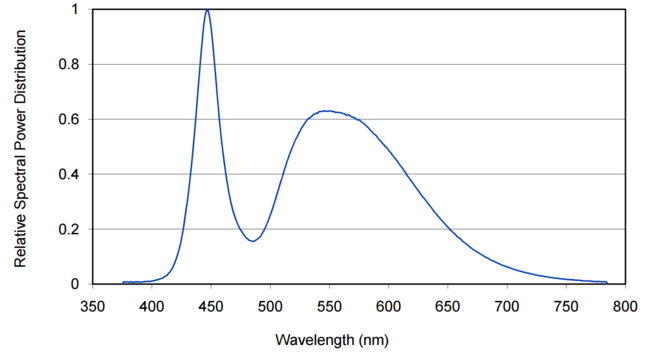} & \!\!\!\!\!\!
		\includegraphics[width = 0.4\linewidth]{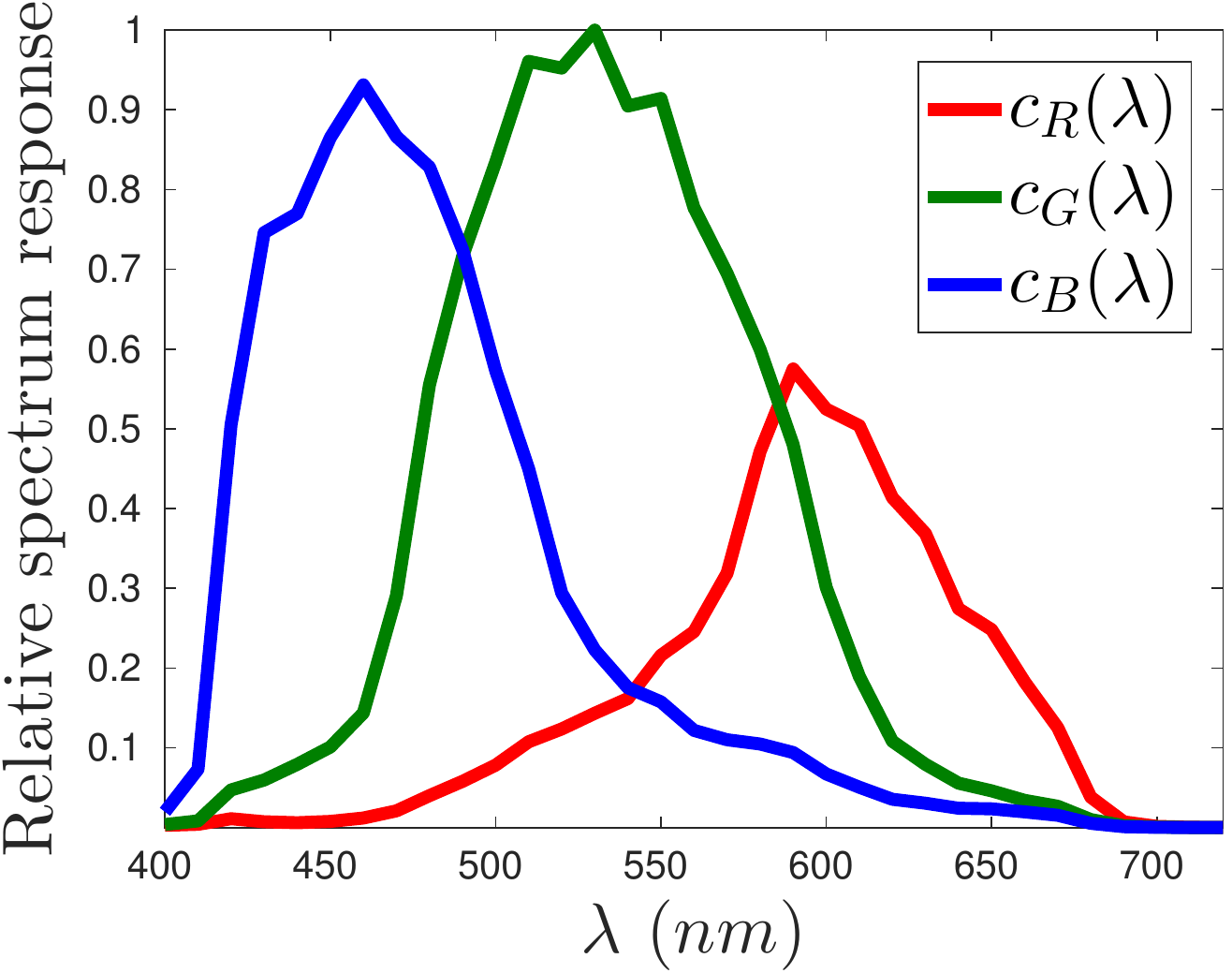} \\
		\small{(a)} & \small{(b)}
	\end{tabular}
\end{center}
\caption{{(a) Emission spectrum $\Phi(\lambda)$ of the {LEDs used} (source: \url{http://www.lumileds.com/uploads/28/DS64-pdf}). (b) Camera response functions in {the three channels $R$, $G$, $B$}, for the Canon EOS 50D camera~\cite{Jiang2013} (which is similar to the Canon EOS 7D we use). Our extension to RGB images of the calibration procedure \colorme{from} Section \ref{sec:2.2} requires nothing else than a camera and two calibration patterns. Therefore, we do not need any of these diagrams in practice.}}
\label{fig:16}
\end{figure}

{We need to introduce a spectral dependency in Model \eqref{eq:12} to extend our study to color}. It seems reasonable to limit this dependency to the intensity ($\lambda$ denotes the wavelength):
\begin{equation}
	\mathbf{s}(\mathbf{x},\lambda) = \Phi(\lambda) \, \cos^\mu \theta \,
		\frac{\mathbf{x}_s-\mathbf{x}}{\|\mathbf{x}_s-\mathbf{x}\|^3}.
\label{eq:103}
\end{equation}
Model \eqref{eq:103} is more complex than Model \eqref{eq:12}, because the intensity $\Phi_0 \in \mathbb{R}^+$ has been replaced by the \emph{emission spectrum} $\Phi(\lambda)$, which is a function {(cf.~Fig.~\ref{fig:16}-a)}. {The calibration} of $\Phi(\lambda)$ could be achieved by using a spectrometer, but {we will show how to extend} the procedure from Section \ref{sec:2.2}, which requires nothing else than a camera and {two} calibration patterns.

Given a point $\mathbf{x}$ of a Lambertian surface with albedo $\rho(\mathbf{x})$, {under the illumination described by the lighting vector} $\mathbf{s}(\mathbf{x})$, we get from \eqref{eq:13}, \eqref{eq:14} and \eqref{eq:15} the expression of the {illuminance} $\epsilon(\mathbf{p})$ of the image plane in {the} pixel $\mathbf{p}$ conjugate to~$\mathbf{x}$:
\begin{equation}
	\epsilon(\mathbf{p}) = \beta \, \cos^4\alpha(\mathbf{p}) \, \frac{\rho(\mathbf{x})}{\pi} \, \left\{\mathbf{s}(\mathbf{x}) \cdot \mathbf{n}(\mathbf{x}) \right\}_+.
\end{equation}
{This expression is easily extended to the case where {$\mathbf{s}(\mathbf{x})$ and $\rho(\mathbf{x})$} depend on $\lambda$:}
\begin{equation}
	\epsilon(\mathbf{p},\lambda) = \beta \, \cos^4\alpha(\mathbf{p}) \, \frac{\rho(\mathbf{x},\lambda)}{\pi} \, \left\{ \mathbf{s}(\mathbf{x},\lambda) \cdot \mathbf{n}(\mathbf{x}) \right\}_+.
\label{eq:105}
\end{equation}
{The one-to-one correspondence between the points $\mathbf{x}$ and the pixels $\mathbf{p}$ allows us to denote} $\rho(\mathbf{p},\lambda)$ and $\mathbf{n}(\mathbf{p})$, in lieu of $\rho(\mathbf{x},\lambda)$ and $\mathbf{n}(\mathbf{x})$. In addition, the light effectively received by each cell goes through a colored filter characterized by its \emph{transmission spectrum} $c_\star(\lambda)$, $\star \in \{R,G,B\}$, whose maximum lies, respectively, in the red, green and blue ranges {(cf. Fig.~\ref{fig:16}-b). To define the \emph{color levels} $I_\star(\mathbf{p})$, $\star \in \{R,G,B\}$, by similarity with the expression \eqref{eq:17} of the (corrected) gray level~$I(\mathbf{p})$, we must multiply \eqref{eq:105} by $c_\star(\lambda)$, and integrate} over the entire spectrum:
\begin{equation}
	I_\star(\mathbf{p}) \!=\! \frac{\gamma \, \beta}{\pi} \, \left\{ \left[\int_{\lambda=0}^{+\infty} \!\! c_\star(\lambda) \, \rho(\mathbf{p},\lambda) \,\mathbf{s}(\mathbf{x},\lambda) \, \mathrm{d}\lambda \right] \cdot \mathbf{n}(\mathbf{p}) \right\}_+.
\label{eq:106}
\end{equation}
Using a Lambertian calibration pattern which is uniformly white \colorme{i.e.,} such that $\rho(\mathbf{p},\lambda) \equiv \rho_0$, allows us to rewrite \eqref{eq:106} as follows:
\begin{equation}
	I_\star(\mathbf{p}) = \gamma \, \beta \, \frac{\rho_0}{\pi} \, \left\{ \left[\int_{\lambda=0}^{+\infty} \!\!\!\!\!\!\! c_\star(\lambda) \,\mathbf{s}(\mathbf{x},\lambda) \, \mathrm{d}\lambda \right] \cdot \mathbf{n}(\mathbf{p}) \right\}_+,
\label{eq:107}
\end{equation}
{which is indeed an extension of \eqref{eq:22} to RGB images, since \eqref{eq:107}} can be rewritten
\begin{equation}
	I_\star(\mathbf{p}) = \gamma \, \beta \, \frac{\rho_0}{\pi} \, \left\{ \mathbf{s}_\star(\mathbf{x}) \cdot \mathbf{n}(\mathbf{p}) \right\}_+,
\label{eq:108}
\end{equation}
{provided that the three \emph{colored lighting vectors} $\mathbf{s}_\star(\mathbf{x})$ are defined as follows}:
\begin{equation}
	\mathbf{s}_\star(\mathbf{x}) = \int_{\lambda=0}^{+\infty} c_\star(\lambda) \, \mathbf{s}(\mathbf{x},\lambda) \, \mathrm{d}\lambda, \quad \star \in \{R,G,B\}.
\label{eq:109}
\end{equation}
Replacing the lighting vector $\mathbf{s}(\mathbf{x},\lambda)$ in \eqref{eq:109} by its expression \eqref{eq:103}, we obtain
the following extension of Model \eqref{eq:12} to color:
\begin{equation}
	\mathbf{s}_\star(\mathbf{x}) = \Phi_\star \, \cos^\mu \theta \,
		\frac{\mathbf{x}_s-\mathbf{x}}{\|\mathbf{x}_s-\mathbf{x}\|^3}, \qquad \star \in \{R,G,B\},
\label{eq:110}
\end{equation}
{where the \emph{colored intensities} $\Phi_\star$ are defined as follows}:
\begin{equation}
	\Phi_\star =  \int_{\lambda=0}^{+\infty} c_\star(\lambda) \, \Phi(\lambda) \, \mathrm{d}\lambda, \qquad \star \in \{R,G,B\}.
\label{eq:111}
\end{equation}
{The spectral dependency of the lighting vector $\mathbf{s}(\mathbf{x},\lambda)$ expressed in \eqref{eq:103} is thus \emph{partially} described by Model \eqref{eq:110}}, which contains nine parameters: three for the coordinates of $\mathbf{x}_s$, two for the unit-length vector $\mathbf{n}_s$, plus the three colored intensities $\Phi_R$, $\Phi_G$, $\Phi_B$, and the anisotropy parameter $\mu$. {\colorme{Nonetheless}, since the definition \eqref{eq:111} of $\Phi_\star$ depends on $c_\star(\lambda)$, it follows that the parameters $\Phi_R$, $\Phi_G$ and $\Phi_B$ are not really characteristic of the LED, but of the camera-LED pair.}


\subsection{Spectral Calibration of the Luminous Flux Emitted by a LED}
\label{sec:5.2}

{We use again the Lambertian planar calibration pattern from Section \ref{sec:2.2}. Since it is convex, the incident light} comes solely from the LED. We can thus replace $\mathbf{s}_\star(\mathbf{x})$ by its definition \eqref{eq:110} in the expression \eqref{eq:108} of the color level $I_\star(\mathbf{p})$. {Assuming} that~$\mathbf{x}_s$ is estimated by triangulation and that the anisotropy parameter~$\mu$ is provided by the manufacturer, we then have to solve, in each channel $\star \in \{R,G,B\}$, the following problem, \colorme{which is an extension of Problem~\eqref{eq:24}} ($q$ is the number of poses of the \colorme{Lambertian} calibration pattern):
\begin{equation}
	\underset{\mathbf{m}_{s,\star}}{\operatorname{\min}~} \displaystyle\sum_{j=1}^{q} \!
	\sum_{\mathbf{p} \in \Omega^j} \!\!
	\left[\! \mathbf{m}_{s,\star} \!\cdot\! (\mathbf{x}^j\!-\!\mathbf{x}_s)
	\!-\! \left[\! \displaystyle I_\star^j(\mathbf{p}) \,
	\frac{\|\mathbf{x}_s\!-\!\mathbf{x}^j\|^{3+\mu}}{ \left\{(\mathbf{x}_s\!-\!\mathbf{x}^j) \cdot \mathbf{n}^j\right\}_+} \!\right]^{\!\frac{1}{\mu}} \right]^{\!2}\!\!,
\label{eq:112}
\end{equation}
where $\mathbf{m}_{s,\star}$ is defined by analogy with $\mathbf{m}_s$ {(cf. \eqref{eq:23})}:
\begin{equation}
	\mathbf{m}_{s,\star} = {\Psi_\star}^{\frac{1}{\mu}} \, \mathbf{n}_s,
\end{equation}
and $\Psi_\star$ is defined by analogy with $\Psi$ {(cf. \eqref{eq:19})}:
\begin{equation}
	\Psi_\star = \gamma \, \beta \, \frac{\rho_0}{\pi} \, \Phi_\star.
\label{eq:114}
\end{equation}
Each problem \eqref{eq:112} allows us to estimate a colored intensity $\Phi_R$, $\Phi_G$ or $\Phi_B$ (up to a {common factor}) and the {principal} direction $\mathbf{n}_s$, which is thus estimated three times. Table \ref{tab:1} groups the values obtained for {one of the LEDs} of our \colorme{setup}. The three estimates of $\mathbf{n}_s$ are consistent, but instead of arbitrarily choosing one of them, we compute the weighted mean of these estimates, \colorme{using spherical coordinates}.

\begin{table}[!ht]
{\scriptsize
\begin{center}
	{\renewcommand{\arraystretch}{1.5}
		\begin{tabular}{|c|c|c|}
		\hline
		Red channel & Green channel & Blue channel \\
		\hline
		$\widehat{\mathbf{n}}_{s,R} = \begin{bmatrix}
			0.205 \\
			-0.757 \\
			0.621
		\end{bmatrix}$ & $\widehat{\mathbf{n}}_{s,G} = \begin{bmatrix}
			0.194 \\
			-0.769 \\
			0.608
		\end{bmatrix}$ & $\widehat{\mathbf{n}}_{s,B} = \begin{bmatrix}
			0.188 \\
			-0.844 \\
			0.503
		\end{bmatrix}$ \\
		\hline $\widehat{\Psi}_R=3.10\times10^7$ & $\widehat{\Psi}_G=5.49\times10^7$ &
		$\widehat{\Psi}_B=3.37\times10^7$ \\
		\hline
		\end{tabular}
	}
\end{center}
}
\caption{Parameters of one of the LEDs of our \colorme{setup}, estimated by {solving}~\eqref{eq:112} in each color channel.}
\label{tab:1}
\end{table}

{In Table \ref{tab:1}, the values of $\widehat{\Psi}_R$, $\widehat{\Psi}_G$ and $\widehat{\Psi}_B$ are given without unit because, from the definition \eqref{eq:114} of $\Psi_\star$}, only their relative values are meaningful. As it happens, the value of $\widehat{\Psi}_G$ is roughly twice as much as {those} of $\widehat{\Psi}_R$ and $\widehat{\Psi}_B$, but this does not mean that $\Phi(\lambda)$ is twice \colorme{higher} in the green range than in the red or in the blue ranges, since the definition \eqref{eq:111} of {a given} colored intensity $\Phi_\star$ also depends on the transmission spectrum $c_\star(\lambda)$ in the considered channel.

Our calibration procedure relies on the assumption that the calibration pattern is uniformly white i.e., that $\rho(\mathbf{p},\lambda) \equiv \rho_0$, which \colorme{may be inexact, yet in no way does this question our rationale}. Indeed, if we assume that the color of ``white'' cells from the Lambertian checkerboard (cf.~Fig.~\ref{fig:4}) is uniform \colorme{i.e.,} $\rho(\mathbf{p},\lambda) = \rho(\lambda)$, \colorme{$\forall \mathbf{p} \in \Omega^j$}, and if we denote $\rho_0$ the maximum value of~$\rho(\lambda)$, Eq.\ \eqref{eq:107} is still valid, provided that $c_\star(\lambda)$ is replaced by the function $\overline{c}_\star(\lambda)$ defined as follows\footnote{Since each colored intensity $\Phi_\star$ depends on the transmission spectrum $c_\star(\lambda)$ by its definition \eqref{eq:111},~\eqref{eq:wunderbar} implies that $\Phi_\star$ also depends on the color of the paper upon which the checkerboard is printed.
Hence, the color of the paper will somehow influence the estimated color of the observed scene.}:
\begin{equation}
	\overline{c}_\star(\lambda) = \frac{\rho(\lambda)}{\rho_0} \, c_\star(\lambda).
	\label{eq:wunderbar}
\end{equation}


\subsection{Photometric Stereo under {Colored} Point Light Source Illumination}

If we pretend to extend {Model \eqref{eq:26} to RGB images}, then it must be possible to write the color level at $\mathbf{p}${, in each channel $\star \in \{R,G,B\}$,} in the following manner:
\begin{equation}
	\!I_\star(\mathbf{p}) \!=\! \Psi_\star \, \frac{\rho_\star(\mathbf{p})}{\rho_0}
		\!\left[\!  \frac{\mathbf{n}_s \!\cdot\! \left(\!\mathbf{x}\!-\!\mathbf{x}_s \!\right)} {\|\mathbf{x}\!-\!\mathbf{x}_s\|} \! \right]^\mu
		\frac{\left\{(\mathbf{x}_s\!-\!\mathbf{x}) \!\cdot\! \mathbf{n}(\mathbf{p})\right\}_+}{\|\mathbf{x}_s\!-\!\mathbf{x}\|^3}
\label{eq:116}
\end{equation}
where {the \emph{colored albedos} $\rho_\star(\mathbf{p})$ are some extensions of the albedo $\rho(p)$ to the RGB case. Equating both expressions of $I_\star(\mathbf{p})$ given in \eqref{eq:106} and in \eqref{eq:116}}, and using the definition~\eqref{eq:103} of $\mathbf{s}(\mathbf{x},\lambda)$, we obtain:
\begin{equation}
\Psi_\star  \frac{\rho_\star(\mathbf{p})}{\rho_0} \,   = \frac{\gamma \beta}{\pi}  \displaystyle\int_{\lambda=0}^{+\infty} \!\!\! c_\star(\lambda) \, \rho(\mathbf{p},\lambda) \, \Phi(\lambda) \, \mathrm{d}\lambda.
\label{eq:117}
\end{equation}
Using the definitions~\eqref{eq:114} and~\eqref{eq:111} of $\Psi_\star$ and $\Phi_\star$,~\eqref{eq:117} yields the following expression for the colored albedos:
\begin{equation}
	\rho_\star(\mathbf{p}) \!=\!
	\frac{\displaystyle\int_{\lambda=0}^{+\infty} \!\!\! c_\star(\lambda) \, \rho(\mathbf{p},\!\lambda) \,{\Phi}(\lambda) \, \mathrm{d}\lambda}
	{\displaystyle\int_{\lambda=0}^{+\infty} c_\star(\lambda) \, {\Phi}(\lambda) \, \mathrm{d}\lambda},\, \star \!\in\! \{\!R,\!G,\!B\},
\label{eq:118}
\end{equation}
which is the mean of $\rho(\mathbf{p},\lambda)$ over the entire spectrum, weighted by the product $c_\star(\lambda) \, {\Phi}(\lambda)$. In addition, although the transmission spectrum $c_\star(\lambda)$ depends only on the camera, the emission spectrum ${\Phi}(\lambda)$ usually varies from one LED to another. Thus, generalizing photometric stereo under point light source illumination to RGB images requires to {superscript the colored albedos by the LED index $i$. Hence,} it seems that we have to solve, in each pixel $\mathbf{p}\in\Omega$, the following problem:
\begin{align}
&	I_\star^i(\mathbf{p}) = \Psi^i_\star \, \frac{\rho^i_\star(\mathbf{p})}{\rho_0}
		\!\left[\!  \frac{\mathbf{n}^i_s \!\cdot\! \left(\!\mathbf{x}\!-\!\mathbf{x}_s^i \!\right)} {\|\mathbf{x}\!-\!\mathbf{x}_s^i\|} \! \right]^{\mu^i}
		\frac{\left\{(\mathbf{x}_s^i\!-\!\mathbf{x}) \!\cdot\! \mathbf{n}(\mathbf{p})\right\}_+}{\|\mathbf{x}_s^i\!-\!\mathbf{x}\|^3},\nonumber \\
& \qquad\qquad\qquad\quad~ i\in \{1,\dots,m\},~ \star \in \{R,G,B\}.
\label{eq:119}
\end{align}
System \eqref{eq:119} is {underdetermined}, because it contains $3m$ equations with $3m+3$ unknowns: {one} colored albedo $\rho_\star^i(\mathbf{p})$ per equation, the depth $z(\mathbf{p})$ of the 3D-point $\mathbf{x}$ conjugate to $\mathbf{p}$ \colorme{(from which we get the coordinates of $\mathbf{x}$), and the normal $\mathbf{n}(\mathbf{p})$}. Apart from this numerical difficulty, the dependency on $i$ of {the colored albedos} is puzzling: while it is clear that the albedo is a photometric characteristic of the surface, independent from the lighting, it should go the same for {the colored albedos}. This shows that the extension to RGB images of photometric stereo is potentially {intractable in the general case. However, such an extension is known to be possible} in two specific cases \cite{CVPR2016}:
\begin{itemize}
	\item For a non-colored surface \colorme{i.e.,} when $\rho(\mathbf{p},\lambda) = \rho(\mathbf{p})$, we deduce from \eqref{eq:118} that $\rho_R(\mathbf{p}) = \rho_G(\mathbf{p}) = \rho_B(\mathbf{p}) = \rho(\mathbf{p})$. Problem \eqref{eq:119} is thus written:
	\begin{align}
		& I_\star^i(\mathbf{p}) = \Psi^i_\star \, \frac{\rho(\mathbf{p})}{\rho_0}
		\!\left[\!  \frac{\mathbf{n}^i_s \!\cdot\! \left(\!\mathbf{x}\!-\!\mathbf{x}_s^i \!\right)} {\|\mathbf{x}\!-\!\mathbf{x}_s^i\|} \! \right]^{\mu^i}
		\frac{\left\{(\mathbf{x}_s^i\!-\!\mathbf{x}) \!\cdot\! \mathbf{n}(\mathbf{p})\right\}_+}{\|\mathbf{x}_s^i\!-\!\mathbf{x}\|^3},\nonumber \\[-0.5em]
		&\qquad\qquad\qquad i\in \{1,\dots,m\},~ \star \in \{R,G,B\}.
		\label{eq:madouebeniguet}
	\end{align}
	If the albedo is known, and if a channel dependency is added to the sources parameters $\mathbf{x}^i_s$, $\mathbf{n}^i_s$ and $\mu^i$, then System~\eqref{eq:madouebeniguet} has 3 unknowns and $3m$ independent equations: a single RGB image may suffice to ensure that the problem is well-determined. This well-known case, which dates back to the 90's \cite{Kontsevich1994}, has been applied to real-time 3D-reconstruction of a white painted deformable surface~\cite{Hernandez2007}.

	\item When the sources are non-colored \colorme{i.e.,} when ${\Phi}^i(\lambda) \equiv \Phi_0$, $\forall i \in \{1,\dots,m\}$, \eqref{eq:118} gives:
	\begin{equation}
		\rho_\star(\mathbf{p}) \!=\!
		\frac{\displaystyle\int_{\lambda=0}^{+\infty} \!\!\! c_\star(\lambda) \, \rho(\mathbf{p},\!\lambda) \, \mathrm{d}\lambda}
		{\displaystyle\int_{\lambda=0}^{+\infty} c_\star(\lambda) \, \mathrm{d}\lambda},~ \star \in \{R,G,B\}.
	\end{equation}
	Since this expression is independent from $i$, Problem \eqref{eq:119} is rewritten:
	\begin{align}
		& I_\star^i(\mathbf{p}) = \Psi_\star \, \frac{\rho_\star(\mathbf{p})}{\rho_0}
		\!\left[\!  \frac{\mathbf{n}^i_s \!\cdot\! \left(\!\mathbf{x}\!-\!\mathbf{x}_s^i \!\right)} {\|\mathbf{x}\!-\!\mathbf{x}_s^i\|} \! \right]^{\mu^i}
		\frac{\left\{(\mathbf{x}_s^i\!-\!\mathbf{x}) \!\cdot\! \mathbf{n}(\mathbf{p})\right\}_+}{\|\mathbf{x}_s^i\!-\!\mathbf{x}\|^3}, \nonumber \\[-0.5em]
		& \qquad\qquad\qquad i\in \{1,\dots,m\},~\star \in \{R,G,B\}.
	\label{eq:122}
	\end{align}
	In \eqref{eq:122}, the parameter $\Psi_\star$ is independent from $i$, but it really depends on the channel $\star$, although the sources are supposed to be non-colored, since in the definition~\eqref{eq:114} of $\Psi_\star$, the colored intensity $\Phi_\star$ is channel-dependent (cf. Eq.~\eqref{eq:111}). System \eqref{eq:122}, which {has} $3m$ equations and six unknowns, is overdetermined if $m\geqslant3$. If $m=2$, it is well-determined but rank-deficient, since in each point, the $6$ lighting vectors are coplanar. Additional information (e.g., a boundary condition) is required~\cite{Mecca2014b}.
	 \\
\end{itemize}

Another case where the colored albedos are independent from $i$ is when the $m$ LEDs all share the same emission spectrum, up to multiplicative coefficients ($\Phi^i(\lambda) = \kappa^i \, \Phi(\lambda),\,\forall i \in \{1,\dots,m\}$).  Under such an assumption, the colored albedos $\rho_\star(\mathbf{p})$ do not have to be indexed by $i$, according to their definition \eqref{eq:118}. Note however that the parameters $\Psi_\star$ still have to be indexed by $i$, in this case. Using the notation
\begin{equation}
  \overline{\rho}_\star(\mathbf{p}) = \frac{\rho_\star(\mathbf{p})}{\rho_0},\quad \star \in \{R,G,B\},
  \label{eq:rho_starr2}
\end{equation}
we obtain the following result: ~\\

\begin{result}
Under the same hypotheses as in Eq.~\eqref{eq:model}, if the $m$ light sources share the same emission spectrum, up to a multiplicative coefficient, then the $m$ RGB images can be modeled as follows:
  \begin{align}
	&\!\!\!I_\star^i(\mathbf{p}) \!=\! \Psi_\star^i \, \overline{\rho}_\star(\mathbf{p})
	\!\left[ \frac{  \mathbf{n}^i_s \!\cdot\! \left(\mathbf{x}-\mathbf{x}^i_s\right)} {\|\mathbf{x}-\mathbf{x}^i_s\|} \right]^{\mu^i}
	\!\!\frac{ \left\{ (\mathbf{x}^i_s-\mathbf{x}) \cdot \mathbf{n}(\mathbf{p}) \right\}_+}{\|\mathbf{x}^i_s-\mathbf{x}\|^3}, \nonumber \\
	& \qquad\qquad\qquad\quad  \,i\in \{1,\dots,m\}, \,\star \in \{R,G,B\}.
\label{eq:123}
\end{align}
where:
\begin{itemize}
  \item $I^i_\star$ is the (corrected) color level in channel $\star$;
  \item $\Psi_R^i$, $\Psi_G^i$ and $\Psi_B^i$ are the colored intensities of the $i$-th source, multiplied by an unknown factor, which is common to all the sources and depends on several camera parameters and on the albedo $\rho_0$ (cf. Eqs.~\eqref{eq:111} and \eqref{eq:114});
  \item $\overline{\rho}_\star$ is the colored albedo in channel $\star$, relatively to $\rho_0$ (cf. Eq.~\eqref{eq:rho_starr2}). 
\end{itemize}
\end{result}

For the \colorme{setup} of Fig.\ \ref{fig:2}-a, the $m=8$ LEDs probably do not exactly share the same spectrum, although they come from the same batch, yet this assumption seems more realistic than that of ``non-colored sources'', and it allows us to better justify the use of~\eqref{eq:123}, which models both the spectral dependency of the albedo and that of the luminous fluxes. 

The calibration procedure described in Section \ref{sec:5.2} provides us with the values of the parameters $\mathbf{x}^i_s$, $\mathbf{n}^i_s$ and $\Psi_\star^i$, $i \in \{1,\dots,m\}$, and the parameters $\mu^i$, $i\in \{1,\dots,m\}$, are provided by the manufacturer. The unknowns of System \eqref{eq:123} are {thus} the \colorme{depth $z(\mathbf{p})$ of $\mathbf{x}$}, the normal $\mathbf{n}(\mathbf{p})$ and the three colored albedos $\overline{\rho}_\star(\mathbf{p})$, $\star \in \{R,G,B\}$. Resorting to RGB images allows us to replace the system~\eqref{eq:model} of $m$ equations with \colorme{four} unknowns, by the system \eqref{eq:123} of $3m$ equations with \colorme{six} unknowns, which should yield more accurate results.


\subsection{{Solving} Colored Photometric Stereo under Point Light Source Illumination}

The alternating strategy from Section~\ref{sec:3.1} is not straightforward to adapt to the case of RGB-valued images, because the albedo is channel-dependent, while the normal vector is not. Principal component analysis could be employed~\cite{Barsky2003}, but we already know from Section~\ref{sec:3} that a differential approach should be preferred anyway. 

A PDE-based approach similar to that of Section~\ref{sec:3.2} is advocated in~\cite{CVPR2016}: ratios between color levels can be computed in each channel $\star \in \{R,G,B\}$, thus eliminating the colored albedos $\overline{\rho}_\star(\mathbf{p})$ and obtaining a system of PDEs in $z$ similar to \eqref{eq:54}. The PDEs to solve remain quasi-linear, \colorme{unlike in}~\cite{Ikeda2008}. \colorme{Yet, we know that the solution strongly depends on the initialization.}

On the other hand, it is straightforward to adapt the method recommended in Section~\ref{sec:4}, by turning the discrete optimization problem~\eqref{eq:66} into
\begin{equation}
\min_{\substack{ \btrho_R,\btrho_G,\btrho_B,\btz}}
 \sum_{\star \in \{R,G,B\}} \sum_{j=1}^n \sum_{i=1}^m\phi\left(r^i_{\star,j}(\btrho_\star,\btz)\right),
\label{eq:124}
\end{equation}
with the following new definitions, which use straightforward notations for the channel dependencies:
\begin{align}
& r^i_{\star,j}(\btrho_\star,\btz) = \tilde{\rho}_{\star,j} \left\{\zeta^i_{\star,j}(\btz)\right\}_+ -I^i_{\star,j}, \\
& \zeta^i_{\star,j}(\btz) = \left[ \mathbf{Q}_j \mathbf{t}^i_{\star,j}(\tilde{z}_j)\right] \cdot \begin{bmatrix} (\nabla \btz)_j \\ -1 \end{bmatrix}.
\end{align}

The actual \colorme{solution} of~\eqref{eq:124} follows immediately from the algorithm described in Section~\ref{sec:4.2}. The depth update simply uses three times more equations, which improves its robustness, while the estimation of each colored albedo is carried out independently in each channel in exactly the same way as in Section~\ref{sec:4.2}.

\begin{figure}[!ht]
\begin{center}
	\begin{tabular}{cc}
\includegraphics[width = 0.58\linewidth]{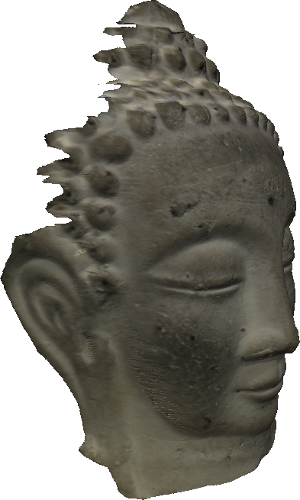} \\
\small{(a)} \\
\def\svgwidth{0.9\linewidth}
		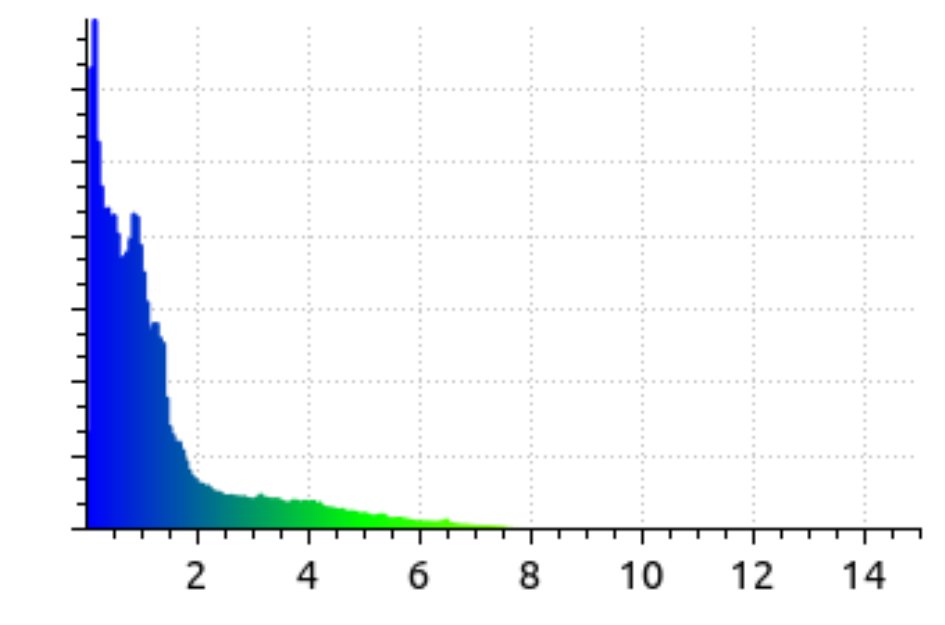 \\
\small{(b)}
	\end{tabular}
\end{center}
\caption{(a) 3D-model estimated from the $m=8$ images {of Fig.\ \ref{fig:2}}, {which are RGB images. (b) Histogram of the distances between this 3D-shape and the ground truth (cf. Fig.~\ref{fig:7}-c). Using RGB images improves the result, \colorme{in comparison with the experiment of Fig.\ \ref{fig:15}}: the median of the \colorme{point-to-point} distances to the ground truth is now equal to \colorme{$0.85~mm$}}.}
\label{fig:17}
\end{figure}

Since the depth estimation now uses more data, {the 3D-model of Fig.\ \ref{fig:17}, which uses RGB images, is improved \colorme{in two ways}, in comparison with that of Fig.\ \ref{fig:15}: it is not only colored, but also more accurate.}

\section{Conclusion and Perspectives}
\label{sec:6}

In this article, we {describe} a photometric stereo-based 3D-reconstruction \colorme{setup} using LEDs as light sources. We {first model} the luminous flux emitted by a LED, then the resulting photometric stereo problem. We pre\-sent a practical procedure for calibrating photometric stereo under point light source illumination, and eventually, we study several numerical solutions. \colorme{Existing methods are based either on alternating estimation of normals and depth, or on direct depth estimation using image ratios. Both these methods have their own advantages, but their convergence is not established. Hence, we introduce a new, provably convergent \colorme{solution based on alternating reweighted least-squares}.} Finally, we {extend} the whole study to RGB images.

\begin{figure}[!htpb]
\begin{center}
	\begin{tabular}{ccc}
		\includegraphics[width = 0.25\linewidth]{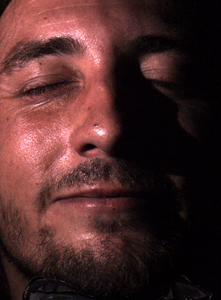} & \quad
		\includegraphics[width = 0.25\linewidth]{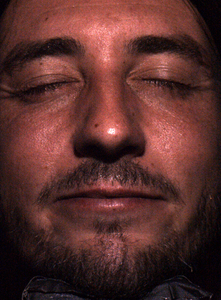} & \quad
		\includegraphics[width = 0.25\linewidth]{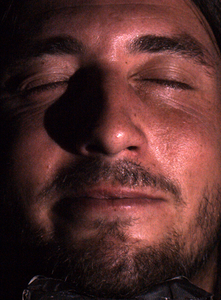} \\
		\small{(a)} & \quad \small{(b)} & \quad \small{(c)} \\
		~ & ~ & ~ \\
		~ & ~ & ~		
	\end{tabular}
	\begin{tabular}{cc}
		\includegraphics[width = 0.45\linewidth]{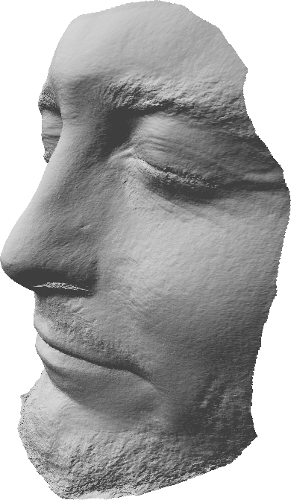} & \quad
		\includegraphics[width = 0.45\linewidth]{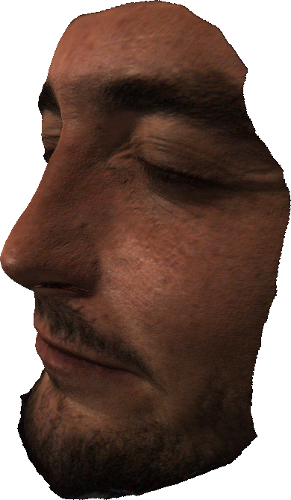} \\
		\small{(d)} & \quad \small{(e)} \\
		~ & ~	
	\end{tabular}
\end{center}
\caption{(a-b-c) Three RGB images (out of $m=8$) of a face {captured} by our \colorme{setup}. (d)~Estimated {3D-shape}. (e) {Colored 3D-model}. {{Since their estimation is relative to the Lambertian planar calibration pattern, the colored albedos of the 3D-model may appear different from the colors of the images}.}}
\label{fig:18}
\end{figure}

The result of Fig.\ \ref{fig:18} suggests that our goal i.e., the estimation of colored 3D-models of faces by photometric stereo, has been reached. Of course, many other types of 3D-scanners exist, but ours relies only on materials which are easy to obtain: a relatively mainstream camera, {eight} LEDs 
and an Arduino controller to synchronize the LEDs with the shutter release. Another significant advantage of our 3D-scanner is that it also estimates the albedo.

\begin{figure}[!htpb]
\begin{center}
	\begin{tabular}{ccc}
		\includegraphics[width = 0.27\linewidth]{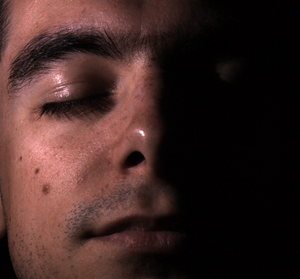} & \,\,
		\includegraphics[width = 0.27\linewidth]{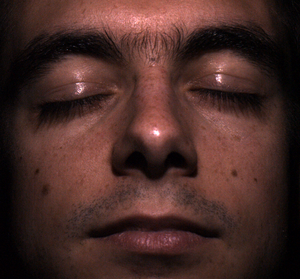} & \,\,
		\includegraphics[width = 0.27\linewidth]{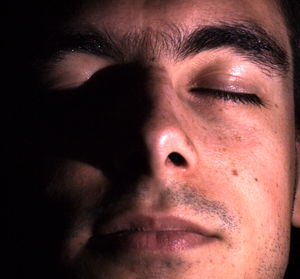} \\
		\small{(a)} & \,\, \small{(b)} & \,\, \small{(c)} \\
		~ & ~ & ~
	\end{tabular}
	\begin{tabular}{cc}
		\includegraphics[width = 0.45\linewidth]{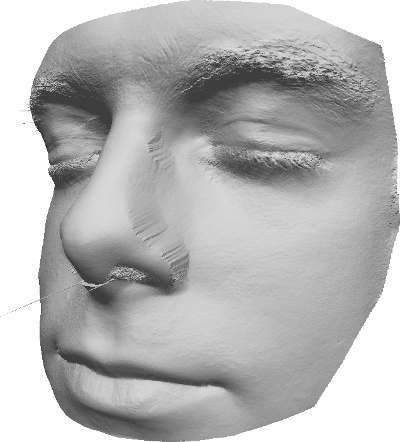} & \quad
		\includegraphics[width = 0.45\linewidth]{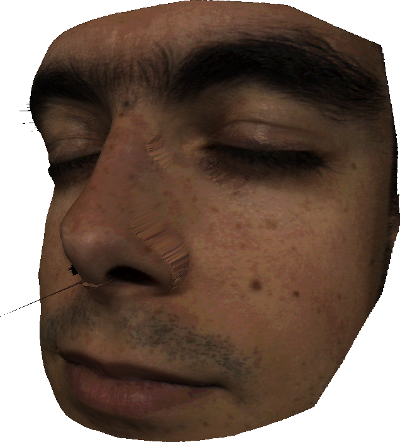} \\
		\small{(d)} & \quad \small{(e)} \\
	\end{tabular}
\end{center}
\caption{(a-b-c) Three images (out of $m=8$) of a face. (d) {Estimated 3D-shape. (e) Colored 3D-model}. {The 3D-reconstruction is not satisfactory under the nose, {which is a dimly lit area}. Robustness of the proposed method to shadows \colorme{could} still be improved.}}
\label{fig:19}
\end{figure}

\colorme{However, there may still be some points where} the shape, and therefore the albedo, are poorly estimated. \colorme{In the example of Fig.~\ref{fig:19},} the area under the nose, which is dimly lit, is poorly reconstructed (this problem does not appear in the example of Fig.\ \ref{fig:18}, because the face is oriented in such a way that it is ``well'' illuminated). \colorme{Although such artifacts remain confined, thanks to robust estimation, future extensions of our work could get rid of them by resorting to an additional regularization term in the variational model.}

\colorme{Besides dealing with these defects, other questions arise. In particular, could we extend our 3D-scanner to full 3D-reconstruction, by coupling the proposed method with multi-view 3D-reconstruction techniques \cite{Hernandez2008}? Aside from obtaining a more complete 3D-reconstruction, this would circumvent the difficult problem of handling possible discontinuities in a depth map, although Fig.~\ref{fig:19} suggests that employing a non-convex estimator already partly allows the recovery of such sharp structures~\cite{Durou2009}.} 

\colorme{Eventually, the proposed numerical framework could be extended in order to automatically refine calibration. Several steps in that direction were already achieved in~\cite{Logothetis2017,Migita2008,Papadhimitri2014b,SSVM2017}, but either without convergence analysis~\cite{Logothetis2017,Migita2008,Papadhimitri2014b} or in the restricted case where only the source intensities are refined~\cite{SSVM2017}. Providing a provably convergent method for uncalibrated photometric stereo under point light source illumination would thus constitute a natural extension of our work.
}

\begin{acknowledgements}

Yvain \textsc{Quéau}, Tao \textsc{Wu} and Daniel \textsc{Cremers} were supported by the ERC Consolidator Grant ``3D Reloaded''.

\end{acknowledgements}

\appendix
{\colormeall 

\section{Proof of Lemma~\ref{lem:1}}

\iprf{
First note that, under the condition \eqref{eq:83}, the function $\mathcal{E}(\cdot,\btz)$ (resp.~$\tilde{\mathcal{E}}_{\btz}(\cdot;\btrho,\btz)$) is twice continuously differentiable at $\btrho$ (resp.~$\btz$), whenever $(\btrho,\btz)$ is sufficiently close to~$(\btrho^*,\btz^*)$. 
The corresponding second-order derivatives are calculated as follows:
\iali{
& \delta\btrho^\top\frac{\partial^2 \mathcal{E}}{\partial \btrho^2}(\btrho,\btz)\delta\btrho  = \sum_{j=1}^n\sum_{i=1}^m \phi''(r^i_j(\btrho,\btz)) \left(\delta\tilde{\rho}_j\{\zeta^i_j(\btz)\}_+\right)^2, \\
& \delta\btz^\top \partial^2 \tilde{\mathcal{E}}_{\btz}(\btz;\btrho,\btz)\delta\btz \nonumber \\ 
& \quad = \sum_{j=1}^n\sum_{i=1}^m \phi''(r^i_j(\btrho,\btz))\left(\tilde{\rho}_j \, 
\chi(\zeta^i_j(\btz)) \, \delta\btz^\top \partial\zeta^i_j(\btz)\right)^2.
}
Comparing the above two formulas with \eqref{eq:78} and \eqref{eq:82}, the conclusion follows from condition~\eqref{eq:dasEq}. 
\qed
}

\section{Proof of Theorem~\ref{thm:1}}

\iprf{
First note that condition \eqref{eq:88} implies that
\iali{
&\dfrac{\partial^2 \mathcal{E}}{\partial\btrho^2}(\btrho^*,\btz^*) \succ {\bm O}, \\
&\frac{\partial^2 \mathcal{E}}{\partial \btz^2}(\btrho^*\!,\btz^*)
-\dfrac{\partial^2 \mathcal{E}}{\partial\btrho\partial\btz}(\btrho^*\!,\btz^*)
\dfrac{\partial^2 \mathcal{E}}{\partial\btrho^2}(\btrho^*\!,\btz^*)^{-1}
\dfrac{\partial^2 \mathcal{E}}{\partial\btrho\partial\btz}(\btrho^*\!,\btz^*) \succ {\bm O}. \label{eq:Bitwo}
}
Utilizing Lemma \ref{lem:1} in conjunction with~\eqref{eq:Bitwo} and~\eqref{eq:89}, we obtain
\iali{
&H_{\btrho}(\btrho^*,\btz^*) \succ {\bm O}, \qquad 
H_{\btz}(\btrho^*,\btz^*) \succ {\bm O}, \\
& \frac{\partial^2 \mathcal{E}}{\partial \btz^2}(\btrho^*\!,\!\btz^*\!)
-\dfrac{\partial^2 \mathcal{E}}{\partial\btrho\partial\btz}(\btrho^*\!,\!\btz^*\!)
H_{\btrho}(\btrho^*,\btz^*)^{-1}
\dfrac{\partial^2 \mathcal{E}}{\partial\btrho\partial\btz}(\btrho^*\!,\!\btz^*) \succ {\bm O}. \label{eq:93}
}
Now consider the iteration
\iali{
& \btz^{(k+1)} = \btz^{(k)}-H_{\btz}\left(\btrho^{(k+1)},\btz^{(k)}\right)^{-1}\frac{\partial \mathcal{E}}{\partial\btz}(\btrho^{(k+1)},\btz^{(k)}) \notag\\[-0.5em]
&= \btz^{(k)} \!-\! H_{\btz}\!\!\left(\!\btrho^{(k)}\!\!-\!H_{\btrho}(\!\btrho^{(k)}\!\!,\!\btz^{(k)}\!)^{-1}\frac{\partial \mathcal{E}}{\partial\btrho}(\btrho^{(k)}\!\!,\!\btz^{(k)}\!),\!\btz^{(k)}\!\!\right)^{\!-1} \notag\\[-0.5em]
&\qquad~ \frac{\partial \mathcal{E}}{\partial\btz}\!\left(\!\btrho^{(k)}\!-\!H_{\btrho}(\btrho^{(k)}\!,\btz^{(k)})^{-1}\frac{\partial \mathcal{E}}{\partial\btrho}(\btrho^{(k)},\btz^{(k)}),\btz^{(k)}\!\right) 
}
as a map $\btz^{(k)}\mapsto\btz^{(k+1)}$. By the Ostrowski theorem \cite[Proposition 10.1.3]{Ortega1970}, 
the local convergence of $\{\btz^{(k)}\}$ to $\btz^*$ follows if the spectral radius of the Jacobian 
\iali{ 
\frac{\partial\btz^{(k+1)}}{\partial\btz^{(k)}}(\btrho^*,\btz^*)
&=\text{id}-H_{\btz}(\btrho^*,\btz^*)^{-1}\frac{\partial^2 \mathcal{E}}{\partial \btz^2}(\btrho^*,\btz^*) \notag\\[-0.5em]
& \!\!\!\!\!\!\!\!\!\!\!\!\!\!\!\!\!\!\!\!\!\!\!\!\!\!\!\!\!\!\!\!\!\!\!\!\!\!\!\!\!\! +H_{\btz}(\btrho^*,\btz^*)^{-1}\frac{\partial^2 \mathcal{E}}{\partial \btrho\partial\btz}(\btrho^*,\btz^*) H_{\btrho}(\btrho^*,\btz^*)^{-1}\notag\\[-0.5em]
& \!\!\!\!\!\!\!\!\!\!\!\!\!\!\!\!\!\!\!\!\!\!\!\!\!\!\!\!\!\!\!\!\!\!\! \frac{\partial^2 \mathcal{E}}{\partial \btrho\partial\btz}(\btrho^*,\btz^*) 
}
is strictly less than 1. 
Using the similarity transform with $H_{\btz}(\btrho^*,\btz^*)^{\frac12}$, we derive:
\iali{ 
&\sr\left(\frac{\partial\btz^{(k+1)}}{\partial\btz^{k}}(\btrho^*,\btz^*)\right) \nonumber \\
&\quad = \sr\left(H_{\btz}(\btrho^*,\btz^*)^{\frac12}\frac{\partial \btz^{(k+1)}}{\partial\btz^{k}}(\btrho^*,\btz^*)H_{\btz}(\btrho^*,\btz^*)^{-\frac12}\right) \\[-0.5em]
&\quad= \sr\bigg( \text{id}-H_{\btz}(\btrho^*,\btz^*)^{-\frac12}\frac{\partial^2  \mathcal{E}}{\partial \btz^2}(\btrho^*,\btz^*)H_{\btz}(\btrho^*,\btz^*)^{-\frac12} \notag\\[-0.5em]
& \qquad\quad 
+ H_{\btz}(\btrho^*,\btz^*)^{-\frac12}\frac{\partial^2  \mathcal{E}}{\partial \btrho\partial\btz}(\btrho^*,\btz^*) H_{\btrho}(\btrho^*,\btz^*)^{-1} \nonumber \\[-0.5em]
& \qquad\qquad\qquad\qquad\quad\frac{\partial^2  \mathcal{E}}{\partial \btrho\partial\btz}(\btrho^*,\btz^*)H_{\btz}(\btrho^*,\btz^*)^{-\frac12} \bigg) \\[-0.5em]
&\quad= \sup_{\|\bv\|=1} \bigg| \|\bv\|^2 \nonumber \\[-0.5em]
& \qquad-\bv^\top H_{\btz}(\btrho^*,\btz^*)^{-\frac12}\frac{\partial^2  \mathcal{E}}{\partial \btz^2}(\btrho^*,\btz^*)H_{\btz}(\btrho^*,\btz^*)^{-\frac12}\bv \notag\\[-0.25em]
& \qquad +\bv^\top H_{\btz}(\btrho^*,\btz^*)^{-\frac12}\frac{\partial^2  \mathcal{E}}{\partial \btrho\partial\btz}(\btrho^*,\btz^*) H_{\btrho}(\btrho^*,\btz^*)^{-1} \notag \\[-0.5em]
& \qquad\qquad\qquad\qquad\frac{\partial^2  \mathcal{E}}{\partial \btrho\partial\btz}(\btrho^*,\btz^*)H_{\btz}(\btrho^*,\btz^*)^{-\frac12}\bv \bigg|.
}
It follows from condition \eqref{eq:90} that
\begin{equation}
\frac{\partial^2 \mathcal{E}}{\partial\btz^2}(\btrho^*,\btz^*)\prec2\partial^2 \tilde{\mathcal{E}}_{\btz}(\btz^*;\btrho^*,\btz^*)\preceq2 H_{\btz}(\btrho^*,\btz^*),
\end{equation}
and hence 
\begin{equation}
\text{id}-H_{\btz}(\btrho^*,\btz^*)^{-\frac12}\frac{\partial^2 \mathcal{E}}{\partial \btz^2}(\btrho^*,\btz^*)H_{\btz}(\btrho^*,\btz^*)^{-\frac12}\succ -\text{id}.
\end{equation}
Consequently, there exists $\epsilon_1\in(0,1)$ such that the following inequality holds for an arbitrary $\bv$:
\iali{
& \|\bv\|^2-\bv^\top H_{\btz}(\btrho^*,\btz^*)^{-\frac12}\frac{\partial^2 \mathcal{E}}{\partial \btz^2}(\btrho^*,\btz^*)H_{\btz}(\btrho^*,\btz^*)^{-\frac12}\bv \notag\\
& \qquad\qquad\qquad\qquad\qquad\qquad\qquad\quad \geq -(1-\epsilon_1)\|\bv\|^2.
}
Meanwhile, condition \eqref{eq:93} implies that, for some $\epsilon_2\in(0,1)$:
\iali{ 
& \bv^\top H_{\btz}(\btrho^*,\btz^*)^{-\frac12}\frac{\partial^2 \mathcal{E}}{\partial \btz^2}(\btrho^*,\btz^*)H_{\btz}(\btrho^*,\btz^*)^{-\frac12}\bv \notag\\
&\qquad
-\bv^\top H_{\btz}(\btrho^*,\btz^*)^{-\frac12}\frac{\partial^2 \mathcal{E}}{\partial \btrho\partial\btz}(\btrho^*,\btz^*) H_{\btrho}(\btrho^*,\btz^*)^{-1} \notag\\ 
& \qquad\qquad\qquad\qquad\frac{\partial^2 \mathcal{E}}{\partial \btrho\partial\btz}(\btrho^*,\btz^*)H_{\btz}(\btrho^*,\btz^*)^{-\frac12}\bv \notag\\
& = (H_{\btz}(\btrho^*,\btz^*)^{-\frac12}\bv)^\top \notag \\
& \quad
\Big( \frac{\partial^2 \mathcal{E}}{\partial \btz^2}(\btrho^*,\btz^*)-\frac{\partial^2 \mathcal{E}}{\partial \btrho\partial\btz}(\btrho^*,\btz^*) H_{\btrho}(\btrho^*,\btz^*)^{-1}\frac{\partial^2 \mathcal{E}}{\partial \btrho\partial\btz}(\btrho^*,\btz^*) \Big) \notag\\
& \quad \left(H_{\btz}(\btrho^*,\btz^*)^{-\frac12}\bv\right) \\ 
& \geq \epsilon_2\|\bv\|^2.
}

Altogether, we conclude
\ieqn{ 
\sr\left(\frac{\partial\btz^{(k+1)}}{\partial\btz^{k}}(\btrho^*,\btz^*)\right)\leq1-\min(\epsilon_1,\epsilon_2),
}
and hence the convergence of $\{\btz^{(k)}\}$.
The convergence of $\{\btrho^{(k)}\}$ to $\btrho^*$ follows from a similar argument. 
\qed
}
} 

\bibliographystyle{spmpsci}
\bibliography{biblio4}

\end{document}